\documentclass[10pt,twocolumn,letterpaper]{article}

\usepackage{cvpr}              

\usepackage{graphicx}
\usepackage{booktabs}
\usepackage{times}
\usepackage{microtype}
\usepackage{soul}
\usepackage{url}
\usepackage[utf8]{inputenc}
\usepackage{amssymb}
\usepackage{booktabs}
\usepackage[accsupp]{axessibility} 
\usepackage{multirow}
\usepackage{mathrsfs}
\usepackage{multirow} 
\usepackage{colortbl} 
\usepackage{xcolor} 
\usepackage{caption}
\usepackage{xspace}
\usepackage[accsupp]{axessibility}
\usepackage{helvet}
\usepackage{graphicx}
\usepackage{booktabs}
\usepackage{times}
\usepackage{microtype}
\usepackage{float}
\usepackage{placeins}
\usepackage{soul}
\usepackage{url}
\usepackage[utf8]{inputenc}
\usepackage{amssymb}
\usepackage{booktabs}
\usepackage[accsupp]{axessibility} 
\usepackage{multirow}
\usepackage{mathrsfs}
\usepackage{multirow} 
\usepackage{colortbl} 
\usepackage{xcolor} 
\usepackage{caption}
\usepackage{xspace}
\usepackage[ruled, vlined, linesnumbered]{algorithm2e}
\usepackage{algpseudocode}
\usepackage{blindtext}
\usepackage{titletoc}

\newcommand{\topsequence}[4]{\{ #1^{#2} \}_{#2=#3}^{#4}}
\newcommand{\botsequence}[4]{\{ #1_{#2} \}_{#2=#3}^{#4}}
\newcommand{\objpair}{$\left(o_{i}^{t}, o_{j}^{t} \right) $\xspace}

\newcommand{\methodname}{\textbf{\textsc{ImparTail}}\xspace}
\newcommand{\methodnamenosp}{\textsc{ImparTail}}

\newcommand{\RNum}[1]{\uppercase\expandafter{\romannumeral #1\relax}}

\definecolor{ice}{RGB}{237,245,255}
\definecolor{oat}{RGB}{232,226,200}
\newcommand{\methodcolor}{oat}
\newcommand{\baselinecolor}{ice}

\usepackage[ruled, vlined, linesnumbered]{algorithm2e}
\usepackage{algpseudocode}
\usepackage{blindtext}
\usepackage{titletoc}

\definecolor{highlightColor}{RGB}{218, 255, 184}

\definecolor{cvprblue}{rgb}{0.21,0.49,0.74}
\usepackage[pagebackref,breaklinks,colorlinks,allcolors=cvprblue]{hyperref}

\def\paperID{16783} 
\def\confName{CVPR}
\def\confYear{2025}

\title{Towards Unbiased and Robust Spatio-Temporal Scene Graph Generation and Anticipation}

\author{Rohith Peddi\\
UT Dallas\\
\and
Saurabh\\
IIT Delhi\\
\and
Ayush Abhay Shrivastava\\
IIT Delhi\\
\and
Parag Singla\\
IIT Delhi\\
\and
Vibhav Gogate\\
UT Dallas\\
}

\begin{document}
\everymath{\displaystyle\textstyle}  
\maketitle
\begin{abstract}

Spatio-Temporal Scene Graphs (STSGs) provide a concise and expressive representation of dynamic scenes by modeling objects and their evolving relationships over time. However, real-world visual relationships often exhibit a long-tailed distribution, causing existing methods for tasks like Video Scene Graph Generation (VidSGG) and Scene Graph Anticipation (SGA) to produce biased scene graphs. To this end, we propose \methodname, a novel training framework that leverages loss masking and curriculum learning to mitigate bias in the generation and anticipation of spatio-temporal scene graphs. Unlike prior methods that add extra architectural components to learn unbiased estimators, we propose an impartial training objective that reduces the dominance of head classes during learning and focuses on underrepresented tail relationships. Our curriculum-driven mask generation strategy further empowers the model to adaptively adjust its bias mitigation strategy over time, enabling more balanced and robust estimations. To thoroughly assess performance under various distribution shifts, we also introduce two new tasks—Robust Spatio-Temporal Scene Graph Generation and Robust Scene Graph Anticipation—offering a challenging benchmark for evaluating the resilience of STSG models. Extensive experiments on the Action Genome dataset demonstrate the superior unbiased performance and robustness of our method compared to existing baselines. 




\end{abstract}    
\vspace{-6.37mm}

\section{Introduction}
\label{sec:intro}

\vspace{-1.37mm}


Spatio-Temporal Scene Graphs (STSGs) are structured graphs where nodes correspond to objects and edges capture the evolving relationships between them over time \cite{Ji_2019, kim_et_al_3dsgg_2019}. Unlike static scene graphs, STSGs offer nuanced video understanding, enabling autonomous systems to interpret interactions in real-time, anticipate changes, and make informed decisions~\cite{Song_2022}. Yet, constructing effective STSGs that mirror the intricacies of the dynamic, ever-changing real world remains an unsolved problem. In particular, real-world relationships follow a highly imbalanced, long-tailed distribution, with a handful of head classes appearing frequently \cite{wang2020longtailedrb, cao_et_al_label_dist_margin_nips_2019}. In contrast, the vast majority of so-called tail classes are rare but essential for a detailed understanding. This imbalance often skews the learned models towards common relationships, yielding an incomplete, incorrect, unsafe, and distorted interpretation of the scene \cite{tan_et_al_eql_2020}. Addressing this challenge is vital: without a nuanced, detailed, and balanced representation, AI systems risk overlooking the subtle yet critical interpretations that define complex scenes.


\begin{figure}[!t]
  \centering
  \includegraphics[width=\linewidth]{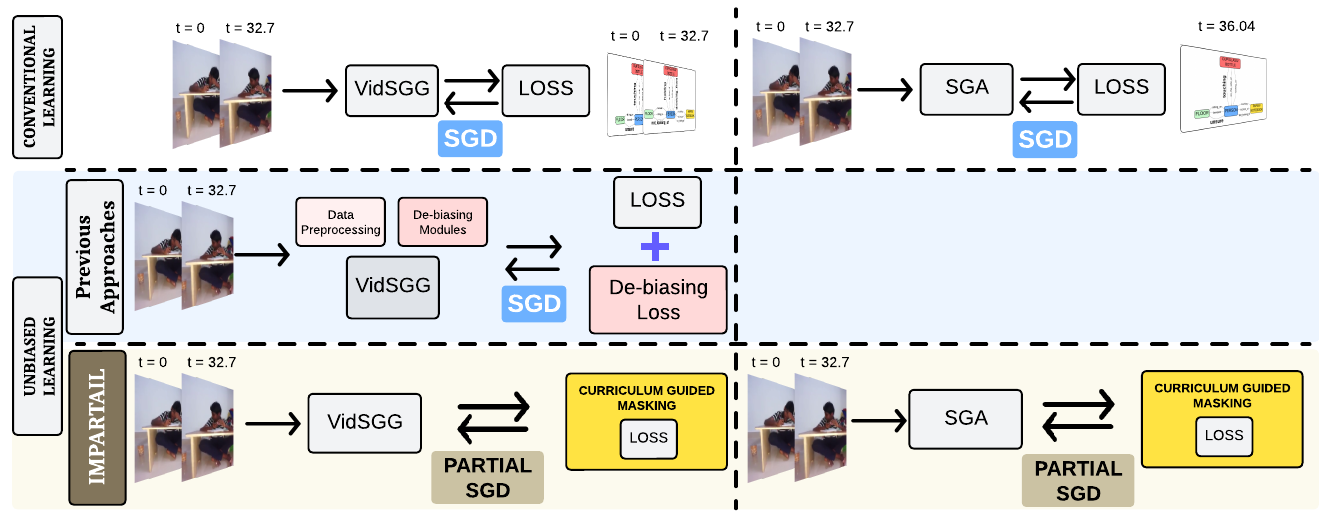}
  \vspace{-6mm}
    \caption{\textbf{Overview.} \underline{Row-1:} Existing training pipelines in the literature for VidSGG/SGA tasks. \underline{Row-2:} Prior unbiased learning work, exemplified by the supplementary architectural modules and loss functions. \underline{Row-3 (\textbf{\methodname}):} A framework through which any prior object-centric, representation learning–based VidSGG/SGA method can be adapted to learn corresponding unbiased estimator.}
    \label{fig:motivation}
    \vspace{-6.37mm}
\end{figure}

To tackle this challenge, we introduce \methodnamenosp, a novel framework designed to overcome long-tail distribution bias in STSG generation. \methodname shifts attention from common head classes to tail classes, promoting a balanced representation that accurately captures all relationship types, not just the dominant ones. It employs a novel \textit{loss masking} technique (similar in spirit to \cite{lin_et_al_iccv_focal_loss_2017, cao_et_al_label_dist_margin_nips_2019}) that amplifies the influence of rare classes during training, ensuring an equitable learning process. This controlled shift towards tail classes allows models to address data distribution disparities without significantly compromising performance on head classes.

\begin{figure*}[!htbp]
  \centering
  \includegraphics[scale=0.32]{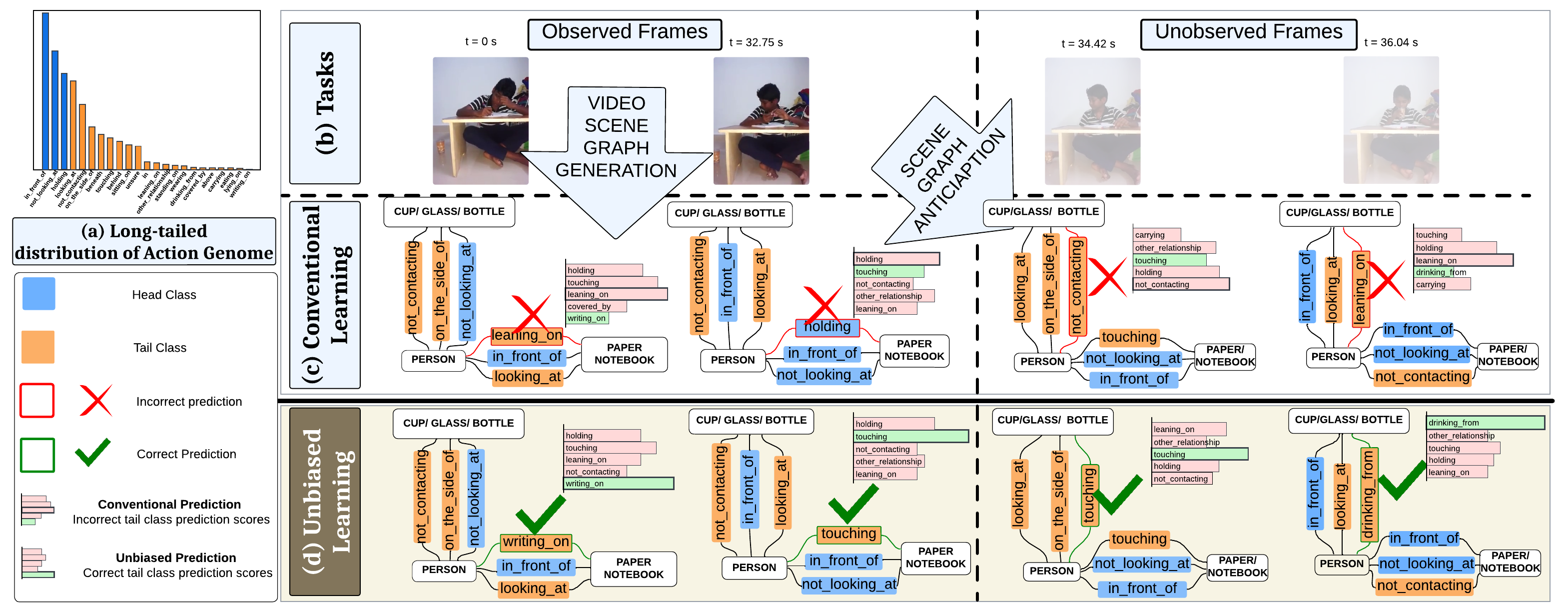}
  \vspace{-3.5mm}
    \caption{\textbf{(a) \underline{Long Tailed Distribution}.} Predicates in Spatio-Temporal Scene Graph (STSG) datasets exhibit a long-tailed distribution; one such example is the Action Genome \cite{Ji_2019} dataset, whose distribution is described at the top left. \textbf{(b) \underline{Tasks}.} We focus on two STSG tasks, including Video Scene Graph Generation (VidSGG) on the left and Scene Graph Anticipation (SGA) on the right. VidSGG entails the identification of fine-grained relationships between the objects observed in the video, such as (\textit{Person}, \textbf{looking\_at}, \textit{Paper Notebook}) and (\textit{Person}, \textbf{not\_looking\_at}, \textit{Paper Notebook}) in respective frames to the left. SGA aims to anticipate the evolution of these relationships to   (\textit{Person}, \textbf{touching}, \textit{Cup}), and eventually, (\textit{Person}, \textbf{drinking\_from}, \textit{Cup}) \cite{peddi_et_al_scene_sayer_2024}. \textbf{(c) \underline{Conventional Learning}.} Due to the inherent long-tailed distribution of these datasets, models learnt using the conventional approaches focus more on the \textit{head} classes and perform poorly on the \textit{tail} classes as illustrated using the prediction scores of STTran \cite{cong_et_al_sttran_2021} on \textit{contacting and attention} relationships (refer middle row). \textbf{(d) \underline{Unbiased Learning}.} To alleviate the dominance of \textit{head} classes during training, in unbiased learning, we focus more on the \textit{tail} classes, ensuring that the learnt models exhibit significantly better performance in predicting both \textit{head and tail} classes (refer bottom row).}
    \label{fig:task_picture}
    \vspace{-5.37mm}
\end{figure*}

Real-world data poses an additional challenge beyond bias: \textit{distributional shifts}. Factors such as lighting, occlusions, and background changes often degrade model performance, affecting the reliability of scene graphs in real-world applications \cite{zhou_et_al_bbn_2019, shrivastava_et_al_hard_example_mining_2016}. Conventional models struggle to generalize across these shifts, limiting their effectiveness in unpredictable conditions. To address this, we introduce two new evaluation tasks\textemdash \textbf{\textit{Robust Spatio-Temporal Scene Graph Generation}} and \textbf{\textit{Robust Scene Graph Anticipation}}\textemdash that simulate realistic distributional shifts to assess model resilience under deployment-like conditions. Thus, this paper makes the following contributions\footnote{\underline{Project:} \href{https://rohithpeddi.github.io/\#/impartail}{https://rohithpeddi.github.io/\#/impartail}}:
\begin{itemize} 
    \item [--] \textbf{\methodnamenosp.} We introduce a novel, effective, unbiased learning framework for generating and anticipating Spatio-Temporal Scene Graphs by leveraging Curriculum-Guided Masked Training with Partial Gradients \cite{hacohen_et_al_cl_tdn_2019, matiisen_et_al_teacher_student_2017}. Our method is especially suited to address the challenges posed by the long-tailed distribution of real-world relationships.
    \item [--] \textbf{Robustness Evaluation.} To assess the resilience of our model against real-world challenges, we systematically induce various corruptions in the input data \cite{tan_et_al_eql_2020, kang_et_al_iclr_2019}. We thoroughly estimate the robustness and generalization capabilities by evaluating the model's performance during the inference phase under these distribution shifts.
    \item [--] \textbf{Empirical Validation.} We conduct extensive experiments on the Action Genome \cite{Ji_2019} benchmark dataset and compare it with existing state-of-the-art methods to empirically validate the efficacy of the proposed approach. 
\end{itemize}

\vspace{-2.5mm}


\section{Related Work} \label{sec:related_work}
\vspace{-1.37mm}
\noindent \textbf{Tasks.} Image Scene Graph Generation (ImgSGG) focuses on representing static visual data—such as 2D and 3D images—as spatial graphs where objects are nodes and their relationships are edges. The field gained significant traction with the foundational Visual Genome project \cite{krishna_et_al_visual_2017}. Building upon this work, Kim et al. \cite{kim_et_al_3dsgg_2019} extended the task to static 3D scene data by incorporating both RGB and depth information. Object interactions over time provide richer contextual information for dynamic visual content like videos. Transforming this content into structured Spatio-Temporal Scene Graphs (STSGs)—with nodes representing objects and edges capturing temporal relationships—is known as Video Scene Graph Generation (VidSGG). Research in VidSGG has concentrated on improving representation learning through advanced object-centric architectures such as STTran \cite{cong_et_al_sttran_2021} and RelFormer \cite{shit_et_al_relformer_2022}. Shifting the focus from the identification and generation of scene graphs, Peddi et al. \cite{peddi_et_al_scene_sayer_2024} recently introduced the Scene Graph Anticipation (SGA) task, which aims to predict STSGs for future frames. 




 \textbf{Unbiased Learning.} TEMPURA \cite{nag_et_al_tempura_2023} and FlCoDe\cite{ khandelwal_correlation_2023} address the challenges posed by long-tailed datasets, such as those found in Action Genome \cite{Ji_2019} and VidVRD \cite{xindi_et_al_vid_vrd_2017} and propose methods for unbiased VidSGG\footnote{To the best of our knowledge, we are the first to investigate biases in the SGA task and assess the robustness of both VidSGG and SGA models.}. Specifically, FloCoDe \cite{khandelwal_correlation_2023} mitigates bias by emphasizing temporal consistency and correcting the imbalanced distribution of visual relationships. Similarly, TEMPURA \cite{nag_et_al_tempura_2023} addresses biases in relationship prediction with memory-guided training to generate balanced relationship representations and applies a Gaussian Mixture Model to reduce predictive uncertainty.

\textbf{Long-Tail Learning.} Long-tailed distributions with a few dominant classes (head classes) often overshadow a more significant number of underrepresented ones (tail classes). This class imbalance typically results in models that perform well on head classes but struggle to generalize to tail classes. To mitigate them, the research community has made significant strides in four directions, which include (a) Cost-Sensitive Learning \cite{huang_et_al_2016, wang_et_al_tail_2017, cui_et_al_2019, ridnik_et_al_asl_loss_2020, wu_et_al_distribution_balanced_2020, ouyang_et_al_factors_finetuning_2016, wang_et_al_tail_2017, you_et_al_exemplar_clustering_2018, zhang_et_al_range_loss_2017}, (b) Mixtures-Of-Experts \cite{Xiang2020LearningFM} , (c) Resampling Techniques \cite{chawla_et_al_smote_2002, drummond_et_al_c45_kdd_2003, zhang_et_al_2021}, and (d) Specialized Architectures \cite{zhang_et_al_2021, zhou_et_al_bbn_2019}.

\textbf{Curriculum Learning (CL)} is a training methodology that structures training by presenting simpler examples first and progressively introducing more complex ones. This approach aims to enhance learning efficiency by aligning the difficulty of training data with the model's learning capacity at each stage \cite{kumar_et_al_self_paced_2010, graves_et_al_auto_cr_2017, hacohen_et_al_cl_tdn_2019, han_et_al_dyn_een_2022}. Despite its potential benefits, implementing CL presents significant challenges. A primary obstacle is the non-trivial task of distinguishing between \textit{easy} and \textit{hard} training samples. Difficulty measures can be predefined based on certain heuristics \cite{ionescu_et_al_diff_vis_search_2016} or learned automatically during the training process \cite{kumar_et_al_self_paced_2010, jiang_et_al_easy_samples_2014, weinshall_et_al_cr_2018, jiang_et_al_mentornet_2017, ren_et_al_ltrw_2018, hacohen_et_al_cl_tdn_2019, matiisen_et_al_teacher_student_2017}. Alongside difficulty assessment, a scheduling strategy is essential to determine when and how to introduce more challenging data \cite{graves_et_al_auto_cr_2017}.





\begin{figure*}[!htbp]
  \centering
  \includegraphics[scale=0.3]{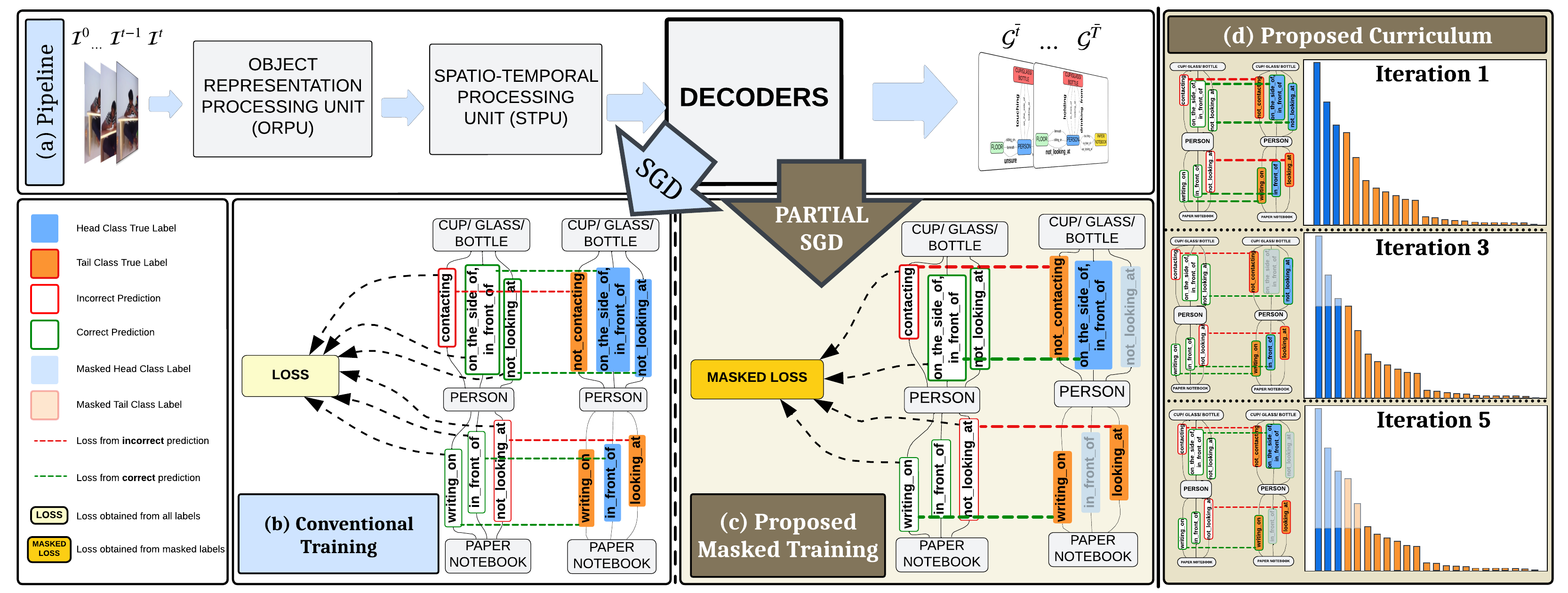}
    \vspace{-3mm}
    \caption{\textbf{Overview of \methodname} 
    \textbf{(a) \underline{Pipeline.}} The forward pass of \methodname begins with an ORPU, where initial object proposals are generated for each observed frame. These object representations are then fed to STPUs designed to construct spatio-temporal context-aware relationship representations of interacting objects. \methodname applied to both tasks VidSGG and SGA remains mostly the same, with an additional LDPU unit added for SGA to anticipate relationship representations for future frames. These observed (for VidSGG)/ anticipated (for SGA) relationship representations are then decoded to construct STSGs. \textbf{(b) \underline{Conventional Training.}} Previous approaches estimated loss for all relationship predicates (head and tail classes).  \textbf{(c) \underline{Masked Training.}} With the inherent long-tailed nature of the STSG datasets, conventional training results in biased VidSGG and SGA models. Thus, to de-bias the training and learn an unbiased model, in \methodname, instead of estimating loss for all relationship predicates, we estimate a \textit{masked loss}, where we selectively mask the labels corresponding to dominant head classes and void their contribution in learning. \textbf{(d) \underline{Curriculum-Guided Mask Generation.}} In \methodname, we introduce a curriculum-based approach for masking relationship predicate labels during training. At each iteration, we adjust the selection of masked predicates to balance the class distribution progressively. As illustrated, initially, the model trains on the original, long-tailed distribution. As training advances, we systematically mask predicate labels from the \textit{head} classes, gradually shifting the distribution toward uniformity.  }
    \label{fig:overview}
    \vspace{-6.37mm}
\end{figure*}

\vspace{-2.37mm}
\section{Notation \& Background} \label{sec:notation}
\vspace{-1.37mm}

We borrow the notation from SceneSayer\cite{peddi_et_al_scene_sayer_2024} and briefly present it below. Let $\mathcal{D}$ correspond to the ground truth annotations of an STSG dataset. Consider a video segment \( V_{1}^{T} \) from a video \( V \), modeled as a sequence of frames at discrete time steps \( t = 1, 2, \ldots, T \): $V_{1}^{T} = \{ I^{1}, I^{2}, \ldots, I^{T} \}$ where \( T \) denotes the total number of frames in the segment. In each frame \( I^{t} \), we represent the scene using a \textbf{scene graph} that captures the visual objects and their pairwise relationships. Let \( O^{t} = \{ o_{1}^{t}, o_{2}^{t}, \ldots, o_{N(t)}^{t} \} \) be the set of observed objects in frame \( I^{t} \), where \( N(t) \) is the number of objects in that frame. Each object \( o_{k}^{t} \) is characterized by: A bounding box \( \mathbf{b}_{k}^{t} \in [0,1]^{4} \), representing its spatial location and A category label \( \mathbf{c}_{k}^{t} \in \mathcal{C} \), where \( \mathcal{C} \) is the set of object categories.

To describe relationships between objects, we define \( \mathcal{P} \) as the set of all predicate classes representing possible spatio-temporal relationships. For any pair of objects \( (o_{i}^{t}, o_{j}^{t}) \), multiple predicates \( \{ p_{ijk}^{t} \} \) may be associated, where each predicate \( p_{ijk}^{t} \in \mathcal{P} \). A \textbf{relationship instance} is then defined as the triplet: $r_{ijk}^{t} = (o_{i}^{t}, p_{ijk}^{t}, o_{j}^{t})$. Thus, the scene graph \( \mathcal{G}^{t} \) for frame \( t \) consists of all such relationship triplets: $\mathcal{G}^{t} = \{ r_{ijk}^{t} \}_{i,j,k}$. For each observed object \( o_{i}^{t} \) and each object pair \( (o_{i}^{t}, o_{j}^{t}) \), we define probability distributions over object categories and predicate classes, respectively: \( \hat{\mathbf{c}}_{i}^{t} \in [0,1]^{|\mathcal{C}|} \) is the probability distribution over object categories for \( o_{i}^{t} \) and \( \hat{\mathbf{p}}_{ij}^{t} \in [0,1]^{|\mathcal{P}|} \) is the probability distribution over predicate classes for the pair \( (o_{i}^{t}, o_{j}^{t}) \). These distributions are normalized, satisfying: $ \sum_{k} \hat{c}_{ik}^{t} = 1, \quad \sum_{k} \hat{p}_{ijk}^{t} = 1.$

We distinguish and clarify the objectives of Video Scene Graph Generation (VidSGG) and Scene Graph Anticipation (SGA) as follows: (a) \textbf{VidSGG.} The primary goal of \textbf{VidSGG} is to construct a sequence of scene graphs \(\{\mathcal{G}^{t}\}_{t=1}^{T}\) corresponding to the observed video segment \( V_{1}^{T} = \{ I^{t} \}_{t=1}^{T} \). This involves: Identifying objects \( \{ o_{k}^{t} \}_{k=1}^{N(t)} \) within each frame \( I^{t} \) and Determining all possible pairwise relationships \( \{ r_{ijk}^{t} \}_{i,j,k} \) among the detected objects. (b) \textbf{SGA.} The main objective of \textbf{SGA} is to generate scene graphs \( \{\mathcal{G}^{t}\}_{t=T+1}^{T+H} \) for future frames \( V_{T+1}^{T+H} = \{ I^{t} \}_{t=T+1}^{T+H} \) of the video, using information from the observed segment \( V_{1}^{T} \). Here, \( H \) denotes the anticipation horizon\footnote{Following the definition and assumptions from SceneSayer \cite{peddi_et_al_scene_sayer_2024}, we assume the continuity of observed objects in future frames.}.



\textbf{Graph Building Strategies.} Following the literature in Video Scene Graph Generation (VidSGG) and Scene Graph Anticipation (SGA), we implement three distinct strategies for constructing scene graphs: \textbf{(a) With Constraint Strategy}: This strategy imposes a unique interaction constraint between every pair of objects within a scene. Specifically, for any two objects \((o_{i}^{t}, o_{j}^{t})\), there exists exactly one relationship predicate \(p_{ij}^{t}\) (for each relationship category) that describes their interaction. We include these relationship triplets \(\{ r_{ij}^{t} \}_{ij}\) in the scene graph \(\mathcal{G}^{t}\). \textbf{(b) No Constraint Strategy}: This approach allows for a more expressive graph structure by permitting multiple relationship predicates between any pair of interacting objects \((o_{i}^{t}, o_{j}^{t})\). We incorporate all predicted relationship triplets \(\{ r_{ijk}^{t} \}_{i,j,k}\) into the scene graph \(\mathcal{G}^{t}\). \textbf{(c) Semi-Constraint Strategy}: This strategy strikes a balance by permitting multiple relationships between object pairs \((o_{i}^{t}, o_{j}^{t})\) only if their confidence scores exceed a predefined threshold. Specifically, we add relationship triplets \(\{ r_{ijk}^{t} \}_{i,j,k}\) to the scene graph \(\mathcal{G}^{t}\) when the confidence score satisfies \(\hat{p}_{ijk}^{t} > \theta_{k}\).



\vspace{-2.37mm}
\section{Technical Approach} \label{sec:technical_approach}
\vspace{-1.37mm}


In \methodnamenosp, we adopt an object-centric relationship representation processing pipeline as illustrated in Fig~\ref{fig:overview}. Thus, it can be easily extended to de-bias any VidSGG/SGA method\footnote{\textbf{\underline{Note:}} \methodname \textbf{does not} train models for both tasks simultaneously. Instead, it adapts existing approaches for VidSGG/SGA \textbf{separately} and learns the corresponding VidSGG/SGA unbiased estimators.} built on an object-centric relationship representation processing pipeline\footnote{Most VidSGG \cite{cong_et_al_sttran_2021, Feng_2021, nag_et_al_tempura_2023, khandelwal_correlation_2023} and SGA \cite{peddi_et_al_scene_sayer_2024} approaches share a similar object-centric relationship representation processing pipeline. We further note that recent SOTA VidSGG methods still rely on the object-centric pipelines introduced by STTran \cite{cong_et_al_sttran_2021} and DSGDetr \cite{Feng_2021}.}. \methodname comprises four key components. (\RNum{1}) \textbf{Object Representation Processing Unit (ORPU):} This module generates representations for objects detected in video frames. (\RNum{2}) \textbf{Spatio-Temporal Context Processing Unit (STPU):} That constructs object-centric relationship representations tailored for two tasks: \textit{observed relationships} for VidSGG and \textit{anticipated relationships} for SGA. (\RNum{3}) \textbf{Relationship Predicate Decoders:} These decoders predict the relationship predicate labels corresponding to the observed or anticipated relationship representations\footnote{\textbf{\underline{Note:}} As \methodname adapts existing VidSGG/SGA methods, we re-use the feature extraction strategy, ORPU, STPU and Predicate decoder blocks employed by corresponding object-centric VidSGG/SGA methods.}. (\RNum{4}) \textbf{Masked Training with Partial Gradients:} We propose an impartial training objective based on loss masking that curbs the influence of dominant head classes. These are the classes that are over-represented in the dataset and often lead to biased predictions. Our approach aims to reduce their influence while focusing on under-represented tail relationships. Furthermore, our curriculum-driven mask generation method enables the model to iteratively refine its bias mitigation strategy over time, resulting in a balanced performance.


The following sections provide a comprehensive overview of the key components and training methodologies of \methodnamenosp. Specifically, In Sec.~\ref{sec:Pipeline}, we outline a detailed explanation of ORPU and STPU, including their respective inputs and outputs within a single training iteration. In Sec.~\ref{sec:MaskedTraining}, we provide a detailed description of the proposed Masked Training with Partial Gradients (refer Appendix Sec.4 for details on the specific changes employed to STSG methods). 

\vspace{-1.37mm}
\subsection{\textbf{\methodname} Pipeline} \label{sec:Pipeline}
\vspace{-0.87mm}

Given that our approach builds upon existing object-centric VidSGG/SGA methods—most of which share similar representation processing pipelines—we detail the essential architectural components necessary for our framework to effectively adapt a method and learn an unbiased estimator\footnote{Current methods in the literature typically incorporate at least one additional component beyond those specified here. \methodname directly integrates the ORPU, STPU units from them ignoring additional components}.\\
\noindent\textbf{(\RNum{1}) Object Representation Processing Unit (ORPU):} extracts visual features ($\botsequence{\mathbf{v}^t}{i}{1}{N(t)}$), bounding boxes ($\botsequence{\mathbf{b}^t}{i}{1}{N(t)}$), and object category distributions ($\botsequence{\hat{\mathbf{c}}^t}{i}{1}{N(t)}$), for object proposals $\botsequence{o^t}{i}{1}{N(t)}$ in the observed frames utilizing a pre-trained object detector proposed by specific VidSGG/SGA method employed for adaptation. It then employs a transformer encoder \cite{vaswani_et_al_2017} to integrate temporal information and generate temporally consistent object representations (see Eq.~\ref{eq:object_representations}):
\vspace{-2.37mm}
\begin{equation}
    \mathbf{V}_{i}^{(n)} = \operatorname{ObjectEncoder}\left(\mathbb{Q} = \mathbb{K} = \mathbb{V} =  \mathbf{V}_{i}^{(n-1)} \right) 
    \label{eq:object_representations}
\end{equation}
\noindent\textbf{(\RNum{2}) Spatio-Temporal Context Processing Unit (STPU):} learns spatio-temporal context-aware relationship representations, it first constructs representations for interacting objects, then refines spatial context with an encoder, and finally use a transformer encoder to integrate spatio-temporal context information. Specifically, let the relationship representation for each observed frame $I^{t}$ as $\mathbf{z}_{ij}^t$, where $\mathbf{Z}^{t}$ is formed by stacking all relationship features $\{\mathbf{z}_{ij}^{t}\}_{ij}$ corresponding to the objects within the frame. $\mathbf{Z}^{t}$ is then passed through the transformer encoder to produce spatial-context-aware representations. Next, it constructs the matrix $\mathbf{Z}_{ij}$ by stacking the relationship representations $\topsequence{\mathbf{z}_{ij}}{t}{1}{T}$ over all observed frames. This matrix $\mathbf{Z}_{ij}$ is then passed through a temporal encoder, which aggregates temporal information across the representations (refer LDPU \cite{peddi_et_al_scene_sayer_2024} for SGA task):
\vspace{-2.37mm}
\begin{multline}
    {[\mathbf{Z}^{t}}]^{(n)} = \operatorname{SpatialEncoder}\left(\mathbb{Q} = \mathbb{K} = \mathbb{V} = [{\mathbf{Z}^{t}}]^{(n-1)} \right) \\
    \mathbf{Z}_{ij}^{(n)} = \operatorname{TemporalEncoder}\left(\mathbf{Q} = \mathbf{K} = \mathbf{V} = \mathbf{Z}_{ij}^{(n)} \right)
\end{multline}

\noindent\textbf{(\RNum{3}) Relationship Predicate Decoders:} Current VidSGG methods utilize a two-layer MLPs to decode the output relationship representations from STPU described as follows:
\vspace{-2.37mm}
\begin{equation}
    \label{eq:PredicateClassificationVidSGG}
    \hat{\mathbf{p}}^t_{ij} = \operatorname{PredClassifier}\left(\mathbf{z}^t_{ij} \right), \forall t \in [1, \bar{T}]
\end{equation}

We note that for SGA task, following SceneSayer \cite{peddi_et_al_scene_sayer_2024}, we employ two relationship predicate decoders. One for the observed relationship representations and the other for the anticipated representations described as follows:
\vspace{-2.37mm}
\begin{multline}
    \label{eq:PredicateClassificationSGA}
    \hat{\mathbf{p}}^t_{ij} = \operatorname{PredClassifier}_{\text{observed}}\left(\mathbf{z}^t_{ij} \right), \forall t \in [1, T] \\
    \hat{\mathbf{p}}^t_{ij} = \operatorname{PredClassifier}_{\text{anticipated}}\left(\mathbf{z}^t_{ij} \right), \forall t \in [T+1, \bar{T}]
\end{multline}


\vspace{-2.37mm}
\subsection{Masked Training with Partial Gradients}\label{sec:MaskedTraining}  

Our approach stems from the observation that instead of re-weighting training data points which, in our case, entails uniformly adjusting the loss contributed by all relationship predicates derived from observed object pairs across all frames of video), selectively masking the loss contributed by head relationship classes helps learn unbiased estimators.

\subsubsection{Masked Loss} During training, for epoch $e$ and mask-ratio $\mathscr{R}_{m}$, Mask Generator generates a set of relationship predicate label masks $\mathcal{M}^{(e)}=\{\{\mathbf{m}^t_{ijk}\}_{ijk}\}_{t}$ where $\mathbf{m}^t_{ijk} \in \{0, 1\}$. Let $\mathcal{L}_{p^t_{ijk}}$ represent any loss defined over a relationship predicate\footnote{(a) Two interacting objects (i,j) at time (t) can have multiple relationship predicates $\{p^t_{ijk}\}_{k}$ that describe the nature of the interaction; (b) We \textbf{re-use} the predicate loss $\mathcal{L}_{p^t_{ijk}}$ from the method employed for adaptation.}. A mask initialized i.e. setting $\mathbf{m}^t_{ijk}=1$ voids the contribution of loss $\mathcal{L}_{p^t_{ijk}}$ and results in masked loss $\mathscr{L}_{p^t_{ij}}$ given by:

\vspace{-1.25em}
\begin{equation}
    \label{eq:RelMaskedLoss}
    \mathscr{L}_{p^t_{ijk}} = (1-\mathbf{m}^t_{ijk}) * \mathcal{L}_{p^t_{ijk}} = 
    \begin{cases}
        \mathcal{L}_{p^t_{ijk}} & \text{if } \mathbf{m}^t_{ijk} = 0 \\
        0 & \text{if } \mathbf{m}^t_{ijk} = 1
    \end{cases}
\end{equation}
\vspace{-1.25em}

\noindent \underline{\textbf{VidSGG.}} models trained using \methodnamenosp, comprise the combination of (1) \textit{Object Classification Loss} on the object representations and (2) \textit{\textbf{Masked} Predicate Classification Loss} on the observed relationship representations as their training objective.The resultant objective is given in Eq~\ref{eq:VidSGGMaskedLoss}: 


\vspace{-2.25em}
\begin{multline} 
    \label{eq:VidSGGMaskedLoss}
    \underbrace{
        \mathcal{L}_{i} = \sum_{t=1}^{\bar{T}} \mathcal{L}_{i}^{t}
    }_\text{(1)};  \quad 
    \underbrace{
        \mathscr{L} = \sum_{t=1}^{\bar{T}} \mathscr{L}^t, 
        \mathscr{L}^t = \sum_{ijk} \mathscr{L}_{p^t_{ijk}}
    }_\text{\textbf{Masked} Predicate Classification Loss (2)}
    \quad \\
    \mathcal{L} = 
        \sum_{t=1}^{\bar{T}} \left(\lambda_{1} \mathscr{L}^{t} + 
        \lambda_{2} \sum_{i} \mathcal{L}_{i}^{t} \right)
\end{multline}
\vspace{-1em}

\noindent \underline{\textbf{SGA.}} models trained using \methodnamenosp, comprise a combination of \textbf{masked losses} over two types of relationship representations: \textbf{(a) Observed Relationship Representations.} and  \textbf{(b) Anticipated Relationship Representations.} and other method specific losses. Thus, the objective when \methodname is employed to SceneSayer \cite{peddi_et_al_scene_sayer_2024} includes (1) Object Classification Loss, (2) Masked Predicate Classification Loss over observed relationship representations, (3) Masked Predicate Classification Loss over anticipated relationship representations, (4) Bounding Box Regression Loss and (5) Reconstruction Loss as shown in Eq~\ref{eq:SGAMaskedLoss}.

\vspace{-2.25em}
\begin{multline} \label{eq:SGAMaskedLoss}
    \underbrace{
        \mathcal{L}_{i} = \sum_{t=1}^{\bar{T}} \mathcal{L}_{i}^{t}
    }_\text{(1)}; \quad
    \underbrace{
        \mathscr{L}_{\text{gen}} = \sum_{t=1}^{\bar{T}} \mathscr{L}_{\text{gen}}^t, 
        \mathscr{L}_{\text{gen}}^t = \sum_{ijk} \mathscr{L}_{p^t_{ijk}}
    }_\text{\textbf{Masked} Observed Predicate Classification Loss (2)} \\
    \underbrace{
        \mathscr{L}_{\text{ant}}^{(1:T)} = \sum_{t=T+1}^{\min(T+H, \bar{T})} \mathscr{L}_{\text{ant}}^t, 
        \quad
        \mathscr{L}_{\text{ant}}^t = \sum_{ijk} \mathscr{L}_{p^t_{ijk}}
    }_\text{\textbf{Masked} Anticipated Predicate Classification Loss (3)}   \\
    \mathcal{L} = 
        \sum_{t=1}^{\bar{T}} \left(\lambda_{1} \mathscr{L}_{\text{gen}}^{t} + 
        \lambda_{2} \sum_{i} \mathcal{L}_{i}^{t} \right)
        + \\
        \sum_{T=3}^{\bar{T}-1} \left( 
            \lambda_{3} \mathscr{L}_{\text{ant}}^{(1:T)} +
            \lambda_{4} \mathcal{L}_{\text{boxes}}^{(1:T)} + 
            \lambda_{5} \mathcal{L}_{\text{recon}}^{(1:T)}
        \right)  
\end{multline}
\vspace{-1em}


\subsubsection{Mask Generation}\label{sec:Curriculum}

We observed that by tuning the \textbf{mask-ratio} ($\mathscr{R}_{m}$) (i.e., the fraction of annotations masked), \methodname can learn estimators along a spectrum, from those exhibiting bias to those that are unbiased. We propose \textbf{curriculum-guided} mask generation to enable models to progressively refine their bias mitigation strategy.  Devising curriculum learning approaches for any problem necessitates precise, problem-specific definitions of three key components \cite{matiisen_et_al_teacher_student_2017}: (a) the ability to rank tasks from easy to hard, (b) the capability to determine mastery over a specific task, and (c) a procedure that periodically integrates easier tasks with more challenging ones. Thus, \methodname employs a simple yet effective curriculum learning method to generate masks for relationship predicate labels to train VidSGG/SGA models.

Banking on the transformers' prowess in sequential processing tasks, we design the curriculum-guided mask generation process for VidSGG and SGA tasks using simple heuristics as follows: (a) \textit{Ranking:} We assume that more data associated with a particular relationship predicate label indicates a better ability to predict that class. Therefore, we rank the hardness of relationship predicate learning based on the quantity of data available for them. (b) \textit{Mastery:} We employ recall as the metric to assess mastery over predicting a specific relationship predicate class, while mean recall evaluates overall performance across all classes. (c) \textit{Progressive Mixing:} We propose a simple distributional strategy that progressively balances the class distribution, gradually moving toward an equilibrium where all classes are uniformly represented. 
Simply put, as training progresses, we stochastically mask the loss of head classes (easier examples), allowing the model to focus on learning about tail classes (harder examples) and the number of masks increases each epoch. The above discussion yields Alg~\ref{alg:cur_guided_mask_generation}.

\vspace{-1em}
\begin{algorithm}[!h]
\SetCustomAlgoRuledWidth{\textwidth}
\DontPrintSemicolon
\footnotesize{
\KwIn{Epoch: $e$, Sampling Ratio: $\mathscr{R}_{s}$, Dataset Annotations: $\mathcal{D}$, Total predicate labels: $N$, Total predicates: $\mathcal{P}$, Videos: $\mathscr{V}$ }
\KwOut{Masks: $\mathcal{M}^{(e)}$}


\textcolor{blue}{*** Determine Target Counts \textbf{[can also be a fixed input]} ***}

$\mathscr{R}_{m} = e \times \mathscr{R}_{s}$ \quad \textcolor{blue}{** Masking Ratio **} \;

$N_{\text{target}} = \operatorname{round}(N \times \mathscr{R}_{m})$\;
    
\textcolor{blue}{*** Curriculum-based sampling probabilities $Prob[rel]$ ***}\;
    
\textcolor{blue}{*** Equally weighted distribution \textbf{[can also be learnt]} *** }\;

Set $Prob[rel] = \frac{1}{|\mathcal{P}|}$ \;
Sample target counts $Tar[rel]$ from Multinomial distribution: $Tar[rel] \sim \operatorname{Multinomial}(N_{\text{target}}, Prob[rel])$ \;



\textcolor{blue}{*** Randomly sample instances of relationships in the dataset based on the target counts and construct filtered dataset ***} \;

Construct filtered dataset $\mathcal{F}$ \quad \textcolor{blue}{** See Appendix for details**} \;

\textcolor{blue}{*** Construct masks such that any relation present in the filtered dataset is unmasked, while all others are masked. ***} \;
Initialize empty mask list $\mathcal{M}^{(e)}$\;
\ForEach{Video $v$ in $\mathscr{V}$}{
    Initialize video mask $\mathcal{M}_v$\;
    \ForEach{Frame $f$ in $v$}{
        Initialize frame mask $\mathcal{M}_f$\;
        \ForEach{Object $o$ in $f$}{
            Initialize object mask $\mathcal{M}_o$\;
            Original relations $\mathcal{R}_{o}$ from $\mathcal{D}[v][f]$\;
            \ForEach{Relation $rel$ in $\mathcal{R}_{o}$}{
                \eIf{$rel \in \mathcal{F}[v][f]$}{
                    Set mask value $\mathcal{M}_o[rel] = 0$\;
                }{
                    Set mask value $\mathcal{M}_o[rel] = 1$\;
                }
            }
            Add $\mathcal{M}_o$ to $\mathcal{M}_f$\;
        }
        Add $\mathcal{M}_f$ to $\mathcal{M}_v$\;
    }
    Add $\mathcal{M}_v$ to $\mathcal{M}^{(e)}$\;
}

}
\caption{Mask Generation}
\label{alg:cur_guided_mask_generation}
\end{algorithm}


\vspace{-6.37mm}
\subsection{Robustness Evaluation} \label{sec:Robustness}
\vspace{-1.37mm}

\begin{figure}[!ht]
  \centering
  \includegraphics[width=\linewidth]{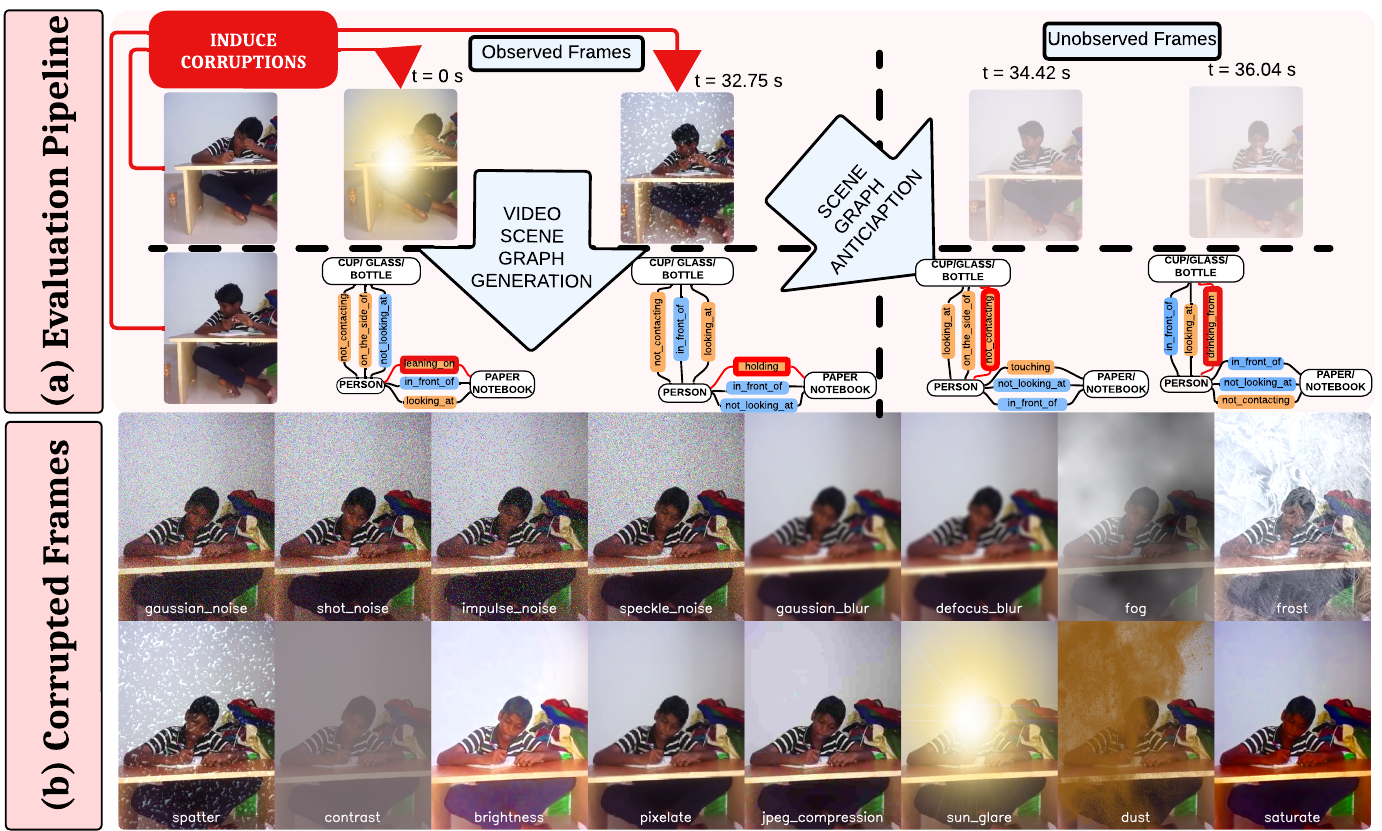}
  \vspace{-7.00mm}
    \caption{\textbf{(a) \underline{Robustness Evaluation Pipeline:}} We present a methodology to assess the robustness of trained VidSGG and SGA models when faced with input distribution shifts. Specifically, we systematically introduce corruptions to the frames of test videos, which are then used as inputs for the trained models. \textbf{(b) \underline{Corrupted Frames:}} We illustrate the frames obtained by inducing various categories of corruptions (see Appendix Sec.4 for details).}
    \label{fig:input_corruptions}
    \vspace{-7.37mm}
\end{figure}

To evaluate the robustness of VidSGG and SGA models against distribution shifts caused by input corruptions, we introduce two tasks: Robust Video Scene Graph Generation and Robust Scene Graph Anticipation. Fig~\ref{fig:input_corruptions} illustrates the evaluation pipeline established to assess the trained models under these distribution shifts induced by input corruptions. As depicted in the figure, we employ a standard set of corruptions commonly used in adversarial robustness research.

\vspace{-3.37mm}
\section{Experiments} \label{sec:experiments}
\vspace{-1.37mm}
We apply \methodname to generate and anticipate STSGs on the Action Genome dataset~\cite{Ji_2019}. We prepare the dataset by following the pre-processing steps described in \cite{cong_et_al_sttran_2021, peddi_et_al_scene_sayer_2024}. 


\textbf{Evaluation Metric.} We evaluate our models using the standard \textit{Recall@K} and \textit{meanRecall@K} metrics, where $K$ takes values within the set \{10, 20, 50\}. The Recall@K metric measures the model's capability to predict the relationships between observed objects in future frames. The long-tailed distribution of relationships in the training set \cite{nag_et_al_tempura_2023} can generate biased scene graphs, causing frequently occurring relationships to dominate the Recall@K metrics. To address this imbalance, we utilize the mean recall metric introduced in \cite{chen2019knowledgeembedded}, a more balanced metric that scores the model's generalisation to all predictive classes. 

In the following sections, we will assess the performance of \methodname on four tasks, namely: (a) \textbf{VidSGG}, (b) \textbf{SGA}, (c) \textbf{RobustVidSGG} and (d) \textbf{RobustSGA}. For each task, we first select previously published methods, train and evaluate them as per the instructions, then adapt them to \methodname framework, re-train and evaluate them. For VidSGG task, we selected STTran \cite{cong_et_al_sttran_2021} and DSGDetr \cite{Feng_2021} as our baselines and for the SGA task, we selected STTran+ \cite{peddi_et_al_scene_sayer_2024}, STTran++ \cite{peddi_et_al_scene_sayer_2024}, DSGDetr+ \cite{peddi_et_al_scene_sayer_2024}, DSGDetr++ \cite{peddi_et_al_scene_sayer_2024}, SceneSayerODE \cite{peddi_et_al_scene_sayer_2024}, SceneSayerSDE \cite{peddi_et_al_scene_sayer_2024} as our baselines. 

\begin{table*}[!h]
    \centering
    \captionsetup{font=small}
    \caption{Mean Recall Results for VidSGG.}
    \vspace{-3.37mm}
    \label{tab:sgg_combined_recall_results}
    \renewcommand{\arraystretch}{1.1}
    \resizebox{\textwidth}{!}{
    \begin{tabular}{l|l|ccc|ccc|ccc}
    \hline
        \multirow{2}{*}{\textbf{Mode}} & \multirow{2}{*}{\textbf{Method}} & \multicolumn{3}{c}{\textbf{With Constraint}} & \multicolumn{3}{c}{\textbf{No Constraint}} & \multicolumn{3}{c}{\textbf{Semi Constraint}} \\ 
         \cmidrule(lr){3-11}         
         &  & \textbf{mR@10} & \textbf{mR@20} & \textbf{mR@50} & \textbf{mR@10} & \textbf{mR@20} & \textbf{mR@50}  & \textbf{mR@10} & \textbf{mR@20} & \textbf{mR@50} \\ 
    \hline
        \multirow{4}{*}{SGDET} & STTran~\cite{cong_et_al_sttran_2021} & 8.0 & 16.6 & 19.3 & 19.3 & 26.9 & 35.6 & 7.7 & 18.2 & 30.4 \\
        & \quad+\textbf{\methodname(Ours)} & \cellcolor{highlightColor}9.4 \textbf{(+17.5\%)} & \cellcolor{highlightColor}21.5 \textbf{(+29.5\%)} & \cellcolor{highlightColor}25.9 \textbf{(+34.2\%)} & \cellcolor{highlightColor}23.5 \textbf{(+21.8\%)} & \cellcolor{highlightColor}33.6 \textbf{(+24.9\%)} & \cellcolor{highlightColor}43.8 \textbf{(+23.0\%)} & \cellcolor{highlightColor}8.6 \textbf{(+11.7\%)} & \cellcolor{highlightColor}21.8 \textbf{(+19.8\%)} & \cellcolor{highlightColor}38.3 \textbf{(+26.0\%)} \\
        & DSGDetr~\cite{Feng_2021} & 6.7 & 14.7 & 19.1 & 23.3 & 29.8 & 36.0 & 6.5 & 16.0 & 30.4 \\
        & \quad+\textbf{\methodname(Ours)} & \cellcolor{highlightColor}7.5 \textbf{(+11.9\%)} & \cellcolor{highlightColor}17.8 \textbf{(+21.1\%)} & \cellcolor{highlightColor}23.7 \textbf{(+24.1\%)} & \cellcolor{highlightColor}27.5 \textbf{(+18.0\%)} & \cellcolor{highlightColor}35.2 \textbf{(+18.1\%)} & \cellcolor{highlightColor}43.3 \textbf{(+20.3\%)} & \cellcolor{highlightColor}7.3 \textbf{(+12.3\%)} & \cellcolor{highlightColor}18.4 \textbf{(+15.0\%)} & \cellcolor{highlightColor}36.6 \textbf{(+20.4\%)} \\
    \hline
        \multirow{4}{*}{SGCLS} & STTran~\cite{cong_et_al_sttran_2021} & 25.0 & 27.5 & 27.6 & 38.8 & 47.1 & 59.9 & 29.5 & 39.9 & 40.9 \\
        & \quad+\textbf{\methodname(Ours)} & \cellcolor{highlightColor}32.3 \textbf{(+29.2\%)} & \cellcolor{highlightColor}36.2 \textbf{(+31.5\%)} & \cellcolor{highlightColor}36.2 \textbf{(+31.2\%)} & \cellcolor{highlightColor}47.4 \textbf{(+22.2\%)} & \cellcolor{highlightColor}57.5 \textbf{(+22.1\%)} & \cellcolor{highlightColor}66.6 \textbf{(+11.2\%)} & \cellcolor{highlightColor}36.2 \textbf{(+22.7\%)} & \cellcolor{highlightColor}50.5 \textbf{(+26.6\%)} & \cellcolor{highlightColor}52.2 \textbf{(+27.6\%)} \\
        & DSGDetr~\cite{Feng_2021} & 25.6 & 28.1 & 28.1 & 39.9 & 49.4 & 64.6 & 30.1 & 40.6 & 41.6 \\
        & \quad+\textbf{\methodname(Ours)} & \cellcolor{highlightColor}32.2 \textbf{(+25.8\%)} & \cellcolor{highlightColor}36.0 \textbf{(+28.1\%)} & \cellcolor{highlightColor}36.0 \textbf{(+28.1\%)} & \cellcolor{highlightColor}48.8 \textbf{(+22.3\%)} & \cellcolor{highlightColor}59.6 \textbf{(+20.6\%)} & \cellcolor{highlightColor}70.1 \textbf{(+8.5\%)} & \cellcolor{highlightColor}36.8 \textbf{(+22.3\%)} & \cellcolor{highlightColor}52.4 \textbf{(+29.1\%)} & \cellcolor{highlightColor}54.9 \textbf{(+32.0\%)} \\
    \hline
        \multirow{4}{*}{PREDCLS} & STTran~\cite{cong_et_al_sttran_2021} & 30.5 & 34.7 & 34.8 & 45.7 & 63.4 & 80.5 & 36.6 & 51.8 & 53.8 \\
        & \quad+\textbf{\methodname(Ours)} & \cellcolor{highlightColor}44.0 \textbf{(+44.3\%)} & \cellcolor{highlightColor}52.7 \textbf{(+51.9\%)} & \cellcolor{highlightColor}52.9 \textbf{(+52.0\%)} & \cellcolor{highlightColor}65.5 \textbf{(+43.3\%)} & \cellcolor{highlightColor}82.0 \textbf{(+29.3\%)} & \cellcolor{highlightColor}93.0 \textbf{(+15.5\%)} & \cellcolor{highlightColor}47.7 \textbf{(+30.3\%)} & \cellcolor{highlightColor}69.7 \textbf{(+34.6\%)} & \cellcolor{highlightColor}73.4 \textbf{(+36.4\%)} \\
        & DSGDetr~\cite{Feng_2021} & 31.5 & 36.1 & 36.2 & 45.6 & 64.4 & 80.5 & 36.5 & 52.5 & 55.2 \\
        & \quad+\textbf{\methodname(Ours)} & \cellcolor{highlightColor}41.0 \textbf{(+30.2\%)} & \cellcolor{highlightColor}48.1 \textbf{(+33.2\%)} & \cellcolor{highlightColor}48.2 \textbf{(+33.1\%)} & \cellcolor{highlightColor}59.4 \textbf{(+30.3\%)} & \cellcolor{highlightColor}76.2 \textbf{(+18.3\%)} & \cellcolor{highlightColor}89.8 \textbf{(+11.6\%)} & \cellcolor{highlightColor}43.9 \textbf{(+20.3\%)} & \cellcolor{highlightColor}65.4 \textbf{(+24.6\%)} & \cellcolor{highlightColor}69.8 \textbf{(+26.4\%)} \\
    \hline
    \end{tabular}
    }
    \vspace{-4.37mm}
\end{table*}

\begin{table*}[!h]
    \centering
    \captionsetup{font=small}
    \caption{Mean Recall Results for SGA.}
    \vspace{-3.37mm}
    \label{tab:sga_mean_recall_results}
    \renewcommand{\arraystretch}{1.1} 
    \resizebox{\textwidth}{!}{
    \begin{tabular}{l|l|cccccc|cccccc|cccccc}
    \hline
         \multirow{3}{*}{$\mathcal{F}$} & \multirow{3}{*}{Method} & \multicolumn{6}{c}{\textbf{AGS}} & \multicolumn{6}{c}{\textbf{PGAGS}} & \multicolumn{6}{c}{\textbf{GAGS}} \\ 
        \cmidrule(lr){3-8} \cmidrule(lr){9-14} \cmidrule(lr){15-20} 
          & & \multicolumn{3}{c}{\textbf{With Constraint}} & \multicolumn{3}{c}{\textbf{No Constraint}}  & \multicolumn{3}{c}{\textbf{With Constraint}} & \multicolumn{3}{c}{\textbf{No Constraint}} & \multicolumn{3}{c}{\textbf{With Constraint}} & \multicolumn{3}{c}{\textbf{No Constraint}}  \\ 
        \cmidrule(lr){3-5} \cmidrule(lr){6-8} \cmidrule(lr){9-11} \cmidrule(lr){12-14} \cmidrule(lr){15-17} \cmidrule(lr){18-20} 
       &   & \textbf{@10} & \textbf{@20} & \textbf{@50} & \textbf{@10} & \textbf{@20} & \textbf{@50}  & \textbf{@10} & \textbf{@20} & \textbf{@50}  & \textbf{@10} & \textbf{@20} & \textbf{@50}  & \textbf{@10} & \textbf{@20} & \textbf{@50}  & \textbf{@10} & \textbf{@20} & \textbf{@50}  \\ \hline
        \multirow{8}{*}{0.5} &        STTran++~\cite{peddi_et_al_scene_sayer_2024} & 7.9 & 16.4 & 18.4 & \cellcolor{highlightColor} \textbf{13.9} & \cellcolor{highlightColor} \textbf{21.3} & 38.5 & 14.3 & 15.8 & 15.8 & 20.9 & 32.5 & 50.1 & 17.8 & 20.9 & 21.0 & \cellcolor{highlightColor} \textbf{25.2} & \cellcolor{highlightColor} \textbf{39.4} & 63.5  \\ 
        &        \quad+\textbf{\methodname(Ours)} & \cellcolor{highlightColor} \textbf{9.3} & \cellcolor{highlightColor} \textbf{18.7} & \cellcolor{highlightColor} \textbf{20.9} & 12.1 & 20.0 & \cellcolor{highlightColor} \textbf{39.6} & \cellcolor{highlightColor} \textbf{20.2} & \cellcolor{highlightColor} \textbf{22.0} & \cellcolor{highlightColor} \textbf{22.0} & \cellcolor{highlightColor} \textbf{24.0} & \cellcolor{highlightColor} \textbf{34.9} & \cellcolor{highlightColor} \textbf{50.5} & \cellcolor{highlightColor} \textbf{19.9} & \cellcolor{highlightColor} \textbf{22.7} & \cellcolor{highlightColor} \textbf{22.8} & 23.7 & 39.2 & \cellcolor{highlightColor} \textbf{64.1}  \\ 
          \cmidrule(lr){2-11} \cmidrule(lr){12-20} 
        &        DSGDetr++~\cite{peddi_et_al_scene_sayer_2024} & 7.4 & 13.4 & 14.6 & 11.8 & 18.2 & 36.1 & 15.0 & 16.3 & 16.3 & 19.9 & 32.3 & 50.6 & 17.1 & 20.0 & 20.0 & 23.2 & 37.3 & 62.9  \\ 
        &        \quad+\textbf{\methodname(Ours)} & \cellcolor{highlightColor} \textbf{8.9} & \cellcolor{highlightColor} \textbf{17.0} & \cellcolor{highlightColor} \textbf{18.6} & \cellcolor{highlightColor} \textbf{13.1} & \cellcolor{highlightColor} \textbf{21.6} & \cellcolor{highlightColor} \textbf{39.6} & \cellcolor{highlightColor} \textbf{18.6} & \cellcolor{highlightColor} \textbf{20.1} & \cellcolor{highlightColor} \textbf{20.1} & \cellcolor{highlightColor} \textbf{22.6} & \cellcolor{highlightColor} \textbf{35.2} & \cellcolor{highlightColor} \textbf{52.5} & \cellcolor{highlightColor} \textbf{21.2} & \cellcolor{highlightColor} \textbf{24.5} & \cellcolor{highlightColor} \textbf{24.6} & \cellcolor{highlightColor} \textbf{28.2} & \cellcolor{highlightColor} \textbf{42.2} & \cellcolor{highlightColor} \textbf{64.9}  \\ 
          \cmidrule(lr){2-11} \cmidrule(lr){12-20} 
        &        SceneSayerODE~\cite{peddi_et_al_scene_sayer_2024} & 5.8 & 12.6 & 16.9 & 14.0 & 22.3 & 36.5 & 11.2 & 12.8 & 12.8 & 16.9 & 26.3 & 45.7 & 17.5 & 20.7 & 20.9 & 24.9 & 38.0 & 61.8  \\ 
        &        \quad+\textbf{\methodname(Ours)} & \cellcolor{highlightColor} \textbf{6.8} & \cellcolor{highlightColor} \textbf{16.1} & \cellcolor{highlightColor} \textbf{22.0} & \cellcolor{highlightColor} \textbf{15.6} & \cellcolor{highlightColor} \textbf{24.8} & \cellcolor{highlightColor} \textbf{39.7} & \cellcolor{highlightColor} \textbf{14.5} & \cellcolor{highlightColor} \textbf{16.4} & \cellcolor{highlightColor} \textbf{16.4} & \cellcolor{highlightColor} \textbf{22.7} & \cellcolor{highlightColor} \textbf{33.6} & \cellcolor{highlightColor} \textbf{49.7} & \cellcolor{highlightColor} \textbf{19.3} & \cellcolor{highlightColor} \textbf{23.2} & \cellcolor{highlightColor} \textbf{23.5} & \cellcolor{highlightColor} \textbf{26.5} & \cellcolor{highlightColor} \textbf{40.9} & \cellcolor{highlightColor} \textbf{63.2}  \\ 
          \cmidrule(lr){2-11} \cmidrule(lr){12-20} 
        &        SceneSayerSDE~\cite{peddi_et_al_scene_sayer_2024} & 6.4 & 13.7 & 18.3 & 15.4 & 23.7 & 38.7 & 15.2 & 17.5 & 17.5 & 22.9 & 34.3 & \cellcolor{highlightColor} \textbf{51.0} & \cellcolor{highlightColor} \textbf{18.2} & \cellcolor{highlightColor} \textbf{21.7} & \cellcolor{highlightColor} \textbf{21.8} & 25.0 & 39.0 & 62.7  \\ 
        &        \quad+\textbf{\methodname(Ours)} & \cellcolor{highlightColor} \textbf{7.4} & \cellcolor{highlightColor} \textbf{19.1} & \cellcolor{highlightColor} \textbf{27.7} & \cellcolor{highlightColor} \textbf{21.8} & \cellcolor{highlightColor} \textbf{31.4} & \cellcolor{highlightColor} \textbf{45.4} & \cellcolor{highlightColor} \textbf{15.7} & \cellcolor{highlightColor} \textbf{17.9} & \cellcolor{highlightColor} \textbf{17.9} & \cellcolor{highlightColor} \textbf{23.6} & \cellcolor{highlightColor} \textbf{34.3} & 50.6 & 17.8 & 21.2 & 21.4 & \cellcolor{highlightColor} \textbf{27.0} & \cellcolor{highlightColor} \textbf{40.7} & \cellcolor{highlightColor} \textbf{63.6}  \\ 
          \hline 
        \multirow{8}{*}{0.7} &        STTran++~\cite{peddi_et_al_scene_sayer_2024} & 9.1 & 18.2 & 20.2 & \cellcolor{highlightColor} \textbf{15.7} & \cellcolor{highlightColor} \textbf{23.7} & 41.9 & 17.2 & 18.6 & 18.6 & 25.3 & 38.3 & 56.1 & 21.9 & 25.0 & 25.0 & \cellcolor{highlightColor} \textbf{31.2} & 47.0 & 75.4  \\ 
        &        \quad+\textbf{\methodname(Ours)} & \cellcolor{highlightColor} \textbf{10.9} & \cellcolor{highlightColor} \textbf{21.9} & \cellcolor{highlightColor} \textbf{24.1} & 14.0 & 23.2 & \cellcolor{highlightColor} \textbf{43.7} & \cellcolor{highlightColor} \textbf{21.0} & \cellcolor{highlightColor} \textbf{22.7} & \cellcolor{highlightColor} \textbf{22.7} & \cellcolor{highlightColor} \textbf{28.0} & \cellcolor{highlightColor} \textbf{41.7} & \cellcolor{highlightColor} \textbf{57.1} & \cellcolor{highlightColor} \textbf{25.8} & \cellcolor{highlightColor} \textbf{29.1} & \cellcolor{highlightColor} \textbf{29.1} & 31.1 & \cellcolor{highlightColor} \textbf{49.2} & \cellcolor{highlightColor} \textbf{76.5}  \\ 
          \cmidrule(lr){2-11} \cmidrule(lr){12-20} 
        &        DSGDetr++~\cite{peddi_et_al_scene_sayer_2024} & 8.4 & 14.8 & 16.0 & 13.2 & 20.0 & 38.8 & 18.1 & 19.4 & 19.4 & 24.8 & 39.5 & 57.3 & 20.8 & 23.8 & 23.8 & 28.6 & 46.1 & 73.8  \\ 
        &        \quad+\textbf{\methodname(Ours)} & \cellcolor{highlightColor} \textbf{10.5} & \cellcolor{highlightColor} \textbf{19.5} & \cellcolor{highlightColor} \textbf{21.2} & \cellcolor{highlightColor} \textbf{14.9} & \cellcolor{highlightColor} \textbf{24.8} & \cellcolor{highlightColor} \textbf{43.9} & \cellcolor{highlightColor} \textbf{20.6} & \cellcolor{highlightColor} \textbf{21.8} & \cellcolor{highlightColor} \textbf{21.8} & \cellcolor{highlightColor} \textbf{26.3} & \cellcolor{highlightColor} \textbf{41.0} & \cellcolor{highlightColor} \textbf{58.1} & \cellcolor{highlightColor} \textbf{28.3} & \cellcolor{highlightColor} \textbf{32.5} & \cellcolor{highlightColor} \textbf{32.5} & \cellcolor{highlightColor} \textbf{31.4} & \cellcolor{highlightColor} \textbf{49.7} & \cellcolor{highlightColor} \textbf{75.7}  \\ 
          \cmidrule(lr){2-11} \cmidrule(lr){12-20} 
        &        SceneSayerODE~\cite{peddi_et_al_scene_sayer_2024} & 6.7 & \cellcolor{highlightColor} \textbf{14.0} & \cellcolor{highlightColor} \textbf{18.5} & 16.4 & 24.9 & 40.5 & 13.6 & 15.1 & 15.1 & 20.5 & 32.4 & 52.8 & 20.7 & 24.0 & 24.0 & 29.8 & 45.2 & 72.0  \\ 
        &        \quad+\textbf{\methodname(Ours)} & \cellcolor{highlightColor} \textbf{6.8} & 13.9 & 18.2 & \cellcolor{highlightColor} \textbf{17.5} & \cellcolor{highlightColor} \textbf{25.8} & \cellcolor{highlightColor} \textbf{41.1} & \cellcolor{highlightColor} \textbf{22.2} & \cellcolor{highlightColor} \textbf{25.6} & \cellcolor{highlightColor} \textbf{25.7} & \cellcolor{highlightColor} \textbf{30.7} & \cellcolor{highlightColor} \textbf{43.9} & \cellcolor{highlightColor} \textbf{55.9} & \cellcolor{highlightColor} \textbf{23.2} & \cellcolor{highlightColor} \textbf{27.5} & \cellcolor{highlightColor} \textbf{27.5} & \cellcolor{highlightColor} \textbf{31.7} & \cellcolor{highlightColor} \textbf{49.9} & \cellcolor{highlightColor} \textbf{73.8}  \\ 
          \cmidrule(lr){2-11} \cmidrule(lr){12-20} 
        &        SceneSayerSDE~\cite{peddi_et_al_scene_sayer_2024} & 7.1 & 14.6 & 19.3 & 17.3 & 26.1 & 42.5 & 17.9 & 19.9 & 19.9 & 27.0 & 40.2 & 57.2 & \cellcolor{highlightColor} \textbf{21.0} & \cellcolor{highlightColor} \textbf{24.6} & \cellcolor{highlightColor} \textbf{24.6} & 30.2 & 45.4 & 72.8  \\ 
        &        \quad+\textbf{\methodname(Ours)} & \cellcolor{highlightColor} \textbf{8.6} & \cellcolor{highlightColor} \textbf{21.3} & \cellcolor{highlightColor} \textbf{29.3} & \cellcolor{highlightColor} \textbf{25.6} & \cellcolor{highlightColor} \textbf{35.1} & \cellcolor{highlightColor} \textbf{50.0} & \cellcolor{highlightColor} \textbf{25.9} & \cellcolor{highlightColor} \textbf{30.0} & \cellcolor{highlightColor} \textbf{30.1} & \cellcolor{highlightColor} \textbf{35.5} & \cellcolor{highlightColor} \textbf{48.2} & \cellcolor{highlightColor} \textbf{58.5} & 20.9 & 24.4 & 24.4 & \cellcolor{highlightColor} \textbf{31.6} & \cellcolor{highlightColor} \textbf{47.9} & \cellcolor{highlightColor} \textbf{73.4}  \\ 
          \hline 
    \end{tabular}
    }
    \vspace{-5.37mm}
\end{table*}

\textbf{Settings.} We follow \cite{cong_et_al_sttran_2021} and use three different settings to evaluate models for the VidSGG task. Each setting has a different level of information provided as input to the models. (a) \textbf{Scene Graph Detection (SGDET):} We only provide frames of the video as input, (b) \textbf{Scene Graph Classification (SGCLS):} Along with frames, bounding box information is provided as input to the model and (c) \textbf{Predicate Classification (PREDCLS):} We provide frames, bounding boxes and object labels as input to the model. Similarly, for the SGA task, we follow SceneSayer~\cite{peddi_et_al_scene_sayer_2024} and use three different settings to evaluate models. (a)  \textbf{Action Genome Scenes (AGS):} In AGS, the model's input is limited to raw frames of the video. (b)  \textbf{Partially Grounded Action Genome Scenes (PGAGS):} In this intermediate setting, along with raw video frames, we additionally input the model with precise bounding box information of active interacting objects observed in the scene. (c) \textbf{Grounded Action Genome Scenes (GAGS):} In this model takes precise bounding box information and the categories of the observed interacting objects as input.

\textbf{Remark.} Although \methodname is designed to learn estimators ranging from biased to unbiased through tuning the mask ratio in mask generation algorithm Alg~\ref{alg:cur_guided_mask_generation}, our experiments on ActionGenome revealed that a fixed Stochastic Masking Ratio of 0.9 delivered \textbf{similar performance} in mean recall metrics to a sequential strategy that started at a masking ratio of 0.6 and gradually increased to 0.9. Moreover, the fixed ratio also showed competitive results within fewer training epochs. Therefore, in the paper to report results, we opted for a fixed Masking Ratio of 0.9\footnote{Due to the skewed distribution of the dataset, using a 0.9 masking ratio creates a scenario where, in each training epoch, the model primarily samples head (easier) classes while leaving tail (harder) classes unaltered.}. For completeness, we provide a thorough ablation in the Appendix.


\vspace{-1.37mm}
\subsection{Video Scene Graph Generation.}
\vspace{-1.37mm}

We first trained and evaluated the chosen baseline models. Next, we integrated \methodnamenosp into these baselines and conducted retraining and evaluation. Our experiments encompassed three VidSGG modes SGDET, SGCLS, and PREDCLS and each included three distinct settings: (1) With Constraint, (2) No Constraint, and (3) Semi Constraint. As demonstrated in Table~\ref{tab:sgg_combined_recall_results}, models trained using \methodname consistently outperformed the chosen VidSGG baselines.

\noindent \underline{\textbf{Insights:}} Our results indicate a significant improvement of approximately $\sim$12\% in mR@10 across all configurations for SGDET, a $\sim$22\% increase in mR@10 for the SGCLS mode, and at least $\sim$20\% improvement in mR@10 for the PREDCLS mode. Moreover, our results are on par and occasionally surpass SOTA unbiased learning methods \cite{nag_et_al_tempura_2023}\footnote{\underline{\textbf{Note:}} (a) SOTA at the time of writing this paper, (b) These improvements were achieved without introducing any new architectural components}.

\begin{table*}[!h]
    \centering
    \captionsetup{font=small}
    \caption{Robustness Evaluation Results for VidSGG.}
    \vspace{-2.37mm}
    \label{tab:sgg_corruptions_results}
    \renewcommand{\arraystretch}{1.2} 
    \resizebox{\textwidth}{!}{
    \begin{tabular}{l|l|l|l|ccc|cccccc|ccc}
    \hline
      \multirow{2}{*}{$\mathcal{S}$} & \multirow{2}{*}{Mode} & \multirow{2}{*}{Corruption} & \multirow{2}{*}{Method} & \multicolumn{3}{c}{\textbf{With Constraint}} & \multicolumn{6}{c}{\textbf{No Constraint}} & \multicolumn{3}{c}{\textbf{Semi Constraint}} \\ 
        \cmidrule(lr){5-7} \cmidrule(lr){8-13} \cmidrule(lr){14-16} 
  & & & & \textbf{mR@10} & \textbf{mR@20} & \textbf{mR@50} & \textbf{R@10} & \textbf{R@20} & \textbf{R@50}  & \textbf{mR@10} & \textbf{mR@20} & \textbf{mR@50}  & \textbf{mR@10} & \textbf{mR@20} & \textbf{mR@50}  \\ \hline
   \multirow{10}{*}{3} &      \multirow{10}{*}{SGCLS} & \multirow{2}{*}{Gaussian Noise} &         DSGDetr~\cite{Feng_2021} & 9.6 & 10.3 & 10.3 & 20.9 & 25.4 & 26.8 & 15.7 & 19.4 & 23.4 & 11.4 & 15.3 & 15.7  \\ 
    &    & &         \quad+\textbf{\methodname(Ours)} & \cellcolor{highlightColor} \textbf{13.7} (+42.7\%) & \cellcolor{highlightColor} \textbf{14.9} (+44.7\%) & \cellcolor{highlightColor} \textbf{15.0} (+45.6\%) & \cellcolor{highlightColor} \textbf{21.0} (+0.5\%) & \cellcolor{highlightColor} \textbf{27.3} (+7.5\%) & \cellcolor{highlightColor} \textbf{30.1} (+12.3\%) & \cellcolor{highlightColor} \textbf{20.8} (+32.5\%) & \cellcolor{highlightColor} \textbf{25.4} (+30.9\%) & \cellcolor{highlightColor} \textbf{29.1} (+24.4\%) & \cellcolor{highlightColor} \textbf{15.6} (+36.8\%) & \cellcolor{highlightColor} \textbf{21.5} (+40.5\%) & \cellcolor{highlightColor} \textbf{22.2} (+41.4\%)  \\ 
 \cmidrule(lr){4-16}  
     &    &\multirow{2}{*}{Fog} &         DSGDetr~\cite{Feng_2021} & 22.6 & 24.9 & 24.9 & \cellcolor{highlightColor} \textbf{47.8} & \cellcolor{highlightColor} \textbf{58.4} & 62.0 & 35.5 & 43.4 & 54.6 & 26.6 & 36.1 & 37.2  \\ 
    &    & &         \quad+\textbf{\methodname(Ours)} & \cellcolor{highlightColor} \textbf{28.3} (+25.2\%) & \cellcolor{highlightColor} \textbf{31.8} (+27.7\%) & \cellcolor{highlightColor} \textbf{31.9} (+28.1\%) & 42.6 (-10.9\%) & 56.0 (-4.1\%) & \cellcolor{highlightColor} \textbf{62.1} (+0.2\%) & \cellcolor{highlightColor} \textbf{43.8} (+23.4\%) & \cellcolor{highlightColor} \textbf{53.0} (+22.1\%) & \cellcolor{highlightColor} \textbf{61.7} (+13.0\%) & \cellcolor{highlightColor} \textbf{31.8} (+19.5\%) & \cellcolor{highlightColor} \textbf{45.7} (+26.6\%) & \cellcolor{highlightColor} \textbf{48.2} (+29.6\%)  \\ 
 \cmidrule(lr){4-16}  
     &    &\multirow{2}{*}{Frost} &         DSGDetr~\cite{Feng_2021} & 16.7 & 18.5 & 18.5 & \cellcolor{highlightColor} \textbf{34.5} & \cellcolor{highlightColor} \textbf{42.3} & 45.1 & 26.8 & 33.0 & 40.1 & 19.6 & 26.8 & 27.7  \\ 
    &    & &         \quad+\textbf{\methodname(Ours)} & \cellcolor{highlightColor} \textbf{22.4} (+34.1\%) & \cellcolor{highlightColor} \textbf{25.0} (+35.1\%) & \cellcolor{highlightColor} \textbf{25.1} (+35.7\%) & 31.9 (-7.5\%) & 42.0 (-0.7\%) & \cellcolor{highlightColor} \textbf{47.1} (+4.4\%) & \cellcolor{highlightColor} \textbf{34.9} (+30.2\%) & \cellcolor{highlightColor} \textbf{42.3} (+28.2\%) & \cellcolor{highlightColor} \textbf{48.2} (+20.2\%) & \cellcolor{highlightColor} \textbf{25.9} (+32.1\%) & \cellcolor{highlightColor} \textbf{36.5} (+36.2\%) & \cellcolor{highlightColor} \textbf{38.4} (+38.6\%)  \\ 
 \cmidrule(lr){4-16}  
     &    &\multirow{2}{*}{Brightness} &         DSGDetr~\cite{Feng_2021} & 23.6 & 25.7 & 25.7 & \cellcolor{highlightColor} \textbf{50.8} & \cellcolor{highlightColor} \textbf{61.9} & 65.5 & 36.8 & 45.4 & 57.5 & 27.6 & 37.5 & 38.6  \\ 
    &    & &         \quad+\textbf{\methodname(Ours)} & \cellcolor{highlightColor} \textbf{29.8} (+26.3\%) & \cellcolor{highlightColor} \textbf{33.2} (+29.2\%) & \cellcolor{highlightColor} \textbf{33.2} (+29.2\%) & 45.5 (-10.4\%) & 59.9 (-3.2\%) & \cellcolor{highlightColor} \textbf{66.1} (+0.9\%) & \cellcolor{highlightColor} \textbf{45.0} (+22.3\%) & \cellcolor{highlightColor} \textbf{55.4} (+22.0\%) & \cellcolor{highlightColor} \textbf{65.3} (+13.6\%) & \cellcolor{highlightColor} \textbf{33.7} (+22.1\%) & \cellcolor{highlightColor} \textbf{48.1} (+28.3\%) & \cellcolor{highlightColor} \textbf{50.6} (+31.1\%)  \\ 
 \cmidrule(lr){4-16}  
     &    &\multirow{2}{*}{Sun Glare} &         DSGDetr~\cite{Feng_2021} & 12.1 & 13.2 & 13.2 & \cellcolor{highlightColor} \textbf{26.3} & 32.5 & 34.7 & 19.3 & 24.4 & 30.2 & 14.2 & 19.2 & 19.6  \\ 
    &    & &         \quad+\textbf{\methodname(Ours)} & \cellcolor{highlightColor} \textbf{17.3} (+43.0\%) & \cellcolor{highlightColor} \textbf{19.4} (+47.0\%) & \cellcolor{highlightColor} \textbf{19.4} (+47.0\%) & 25.8 (-1.9\%) & \cellcolor{highlightColor} \textbf{34.3} (+5.5\%) & \cellcolor{highlightColor} \textbf{38.5} (+11.0\%) & \cellcolor{highlightColor} \textbf{26.6} (+37.8\%) & \cellcolor{highlightColor} \textbf{32.2} (+32.0\%) & \cellcolor{highlightColor} \textbf{37.3} (+23.5\%) & \cellcolor{highlightColor} \textbf{19.4} (+36.6\%) & \cellcolor{highlightColor} \textbf{27.6} (+43.7\%) & \cellcolor{highlightColor} \textbf{29.0} (+48.0\%)   \\ \hline
   \multirow{10}{*}{5} &      \multirow{10}{*}{PREDCLS} & \multirow{2}{*}{Gaussian Noise} &         STTran~\cite{cong_et_al_sttran_2021} & 20.0 & 22.3 & 22.4 & \cellcolor{highlightColor} \textbf{64.2} & \cellcolor{highlightColor} \textbf{87.6} & 99.0 & 31.4 & 52.5 & 79.7 & 26.0 & 36.6 & 38.5  \\ 
    &    & &         \quad+\textbf{\methodname(Ours)} & \cellcolor{highlightColor} \textbf{37.6} (+88.0\%) & \cellcolor{highlightColor} \textbf{43.8} (+96.4\%) & \cellcolor{highlightColor} \textbf{43.9} (+96.0\%) & 62.5 (-2.6\%) & 84.6 (-3.4\%) & \cellcolor{highlightColor} \textbf{99.0} (0.0\%) & \cellcolor{highlightColor} \textbf{57.5} (+83.1\%) & \cellcolor{highlightColor} \textbf{77.7} (+48.0\%) & \cellcolor{highlightColor} \textbf{92.7} (+16.3\%) & \cellcolor{highlightColor} \textbf{42.2} (+62.3\%) & \cellcolor{highlightColor} \textbf{60.0} (+63.9\%) & \cellcolor{highlightColor} \textbf{62.9} (+63.4\%)  \\ 
 \cmidrule(lr){4-16}  
     &    &\multirow{2}{*}{Fog} &         STTran~\cite{cong_et_al_sttran_2021} & 26.5 & 30.2 & 30.3 & \cellcolor{highlightColor} \textbf{70.2} & \cellcolor{highlightColor} \textbf{91.1} & \cellcolor{highlightColor} \textbf{99.1} & 41.6 & 61.0 & 80.5 & 33.2 & 46.8 & 48.7  \\ 
    &    & &         \quad+\textbf{\methodname(Ours)} & \cellcolor{highlightColor} \textbf{42.6} (+60.8\%) & \cellcolor{highlightColor} \textbf{50.9} (+68.5\%) & \cellcolor{highlightColor} \textbf{51.1} (+68.6\%) & 64.8 (-7.7\%) & 86.3 (-5.3\%) & 98.8 (-0.3\%) & \cellcolor{highlightColor} \textbf{63.8} (+53.4\%) & \cellcolor{highlightColor} \textbf{80.2} (+31.5\%) & \cellcolor{highlightColor} \textbf{92.7} (+15.2\%) & \cellcolor{highlightColor} \textbf{46.2} (+39.2\%) & \cellcolor{highlightColor} \textbf{65.5} (+40.0\%) & \cellcolor{highlightColor} \textbf{68.2} (+40.0\%)  \\ 
 \cmidrule(lr){4-16}  
     &    &\multirow{2}{*}{Frost} &         STTran~\cite{cong_et_al_sttran_2021} & 25.6 & 29.2 & 29.2 & \cellcolor{highlightColor} \textbf{69.4} & \cellcolor{highlightColor} \textbf{90.7} & \cellcolor{highlightColor} \textbf{99.1} & 41.0 & 60.9 & 80.5 & 32.7 & 46.1 & 48.0  \\ 
    &    & &         \quad+\textbf{\methodname(Ours)} & \cellcolor{highlightColor} \textbf{41.0} (+60.2\%) & \cellcolor{highlightColor} \textbf{49.0} (+67.8\%) & \cellcolor{highlightColor} \textbf{49.2} (+68.5\%) & 62.2 (-10.4\%) & 84.3 (-7.1\%) & 98.5 (-0.6\%) & \cellcolor{highlightColor} \textbf{62.5} (+52.4\%) & \cellcolor{highlightColor} \textbf{78.6} (+29.1\%) & \cellcolor{highlightColor} \textbf{92.7} (+15.2\%) & \cellcolor{highlightColor} \textbf{45.1} (+37.9\%) & \cellcolor{highlightColor} \textbf{62.9} (+36.4\%) & \cellcolor{highlightColor} \textbf{65.1} (+35.6\%)  \\ 
 \cmidrule(lr){4-16}  
     &    &\multirow{2}{*}{Brightness} &         STTran~\cite{cong_et_al_sttran_2021} & 28.2 & 32.0 & 32.1 & \cellcolor{highlightColor} \textbf{71.3} & \cellcolor{highlightColor} \textbf{91.6} & \cellcolor{highlightColor} \textbf{99.2} & 42.8 & 62.0 & 80.4 & 34.5 & 49.0 & 51.2  \\ 
    &    & &         \quad+\textbf{\methodname(Ours)} & \cellcolor{highlightColor} \textbf{42.3} (+50.0\%) & \cellcolor{highlightColor} \textbf{50.4} (+57.5\%) & \cellcolor{highlightColor} \textbf{50.5} (+57.3\%) & 65.9 (-7.6\%) & 87.2 (-4.8\%) & 98.9 (-0.3\%) & \cellcolor{highlightColor} \textbf{64.0} (+49.5\%) & \cellcolor{highlightColor} \textbf{80.8} (+30.3\%) & \cellcolor{highlightColor} \textbf{92.8} (+15.4\%) & \cellcolor{highlightColor} \textbf{46.9} (+35.9\%) & \cellcolor{highlightColor} \textbf{67.8} (+38.4\%) & \cellcolor{highlightColor} \textbf{71.0} (+38.7\%)  \\ 
 \cmidrule(lr){4-16}  
     &    &\multirow{2}{*}{Sun Glare} &         STTran~\cite{cong_et_al_sttran_2021} & 22.5 & 25.1 & 25.2 & \cellcolor{highlightColor} \textbf{66.7} & \cellcolor{highlightColor} \textbf{89.9} & \cellcolor{highlightColor} \textbf{99.1} & 36.7 & 56.7 & 80.0 & 28.9 & 40.5 & 42.3  \\ 
    &    & &         \quad+\textbf{\methodname(Ours)} & \cellcolor{highlightColor} \textbf{40.2} (+78.7\%) & \cellcolor{highlightColor} \textbf{47.5} (+89.2\%) & \cellcolor{highlightColor} \textbf{47.7} (+89.3\%) & 57.9 (-13.2\%) & 81.7 (-9.1\%) & 98.0 (-1.1\%) & \cellcolor{highlightColor} \textbf{60.3} (+64.3\%) & \cellcolor{highlightColor} \textbf{77.5} (+36.7\%) & \cellcolor{highlightColor} \textbf{92.7} (+15.9\%) & \cellcolor{highlightColor} \textbf{43.3} (+49.8\%) & \cellcolor{highlightColor} \textbf{59.3} (+46.4\%) & \cellcolor{highlightColor} \textbf{61.0} (+44.2\%)  \\ 
          \hline 
    \end{tabular}
    }
    \vspace{-5.37mm}
\end{table*}

\begin{table}[!h]
    \centering
    \captionsetup{font=small}
    \caption{Robustness Evaluation Results for SGA.}
    \vspace{-3.37mm}
    \label{tab:sga_corruptions_results}
    \renewcommand{\arraystretch}{1.2} 
    \resizebox{\linewidth}{!}{
    \begin{tabular}{l|l|l|l|ccc}
    \hline
      \multirow{2}{*}{$\mathcal{F}$} & \multirow{2}{*}{Mode} & \multirow{2}{*}{Corruption} & \multirow{2}{*}{Method} & \multicolumn{3}{c}{\textbf{With Constraint}}  \\ 
        \cmidrule(lr){5-7} 
  & & & & \textbf{mR@10} & \textbf{mR@20} & \textbf{mR@50}  \\ \hline
   \multirow{32}{*}{0.5} &      \multirow{32}{*}{PGAGS} & \multirow{8}{*}{Gaussian Noise} &         STTran+~\cite{peddi_et_al_scene_sayer_2024} & \cellcolor{highlightColor} \textbf{7.9} & \cellcolor{highlightColor} \textbf{8.4} & \cellcolor{highlightColor} \textbf{8.4}  \\ 
    &    & &         \quad+\textbf{\methodname(Ours)} & 5.1 (-35.4\%) & 5.4 (-35.7\%) & 5.4 (-35.7\%)  \\ 
    &    & &         DSGDetr+~\cite{peddi_et_al_scene_sayer_2024} & 5.5 & 5.8 & 5.8  \\ 
    &    & &         \quad+\textbf{\methodname(Ours)} & \cellcolor{highlightColor} \textbf{7.5} (+36.4\%) & \cellcolor{highlightColor} \textbf{8.0} (+37.9\%) & \cellcolor{highlightColor} \textbf{8.0} (+37.9\%)  \\ 
    &    & &         STTran++~\cite{peddi_et_al_scene_sayer_2024} & 5.9 & 6.4 & 6.4  \\ 
    &    & &         \quad+\textbf{\methodname(Ours)} & \cellcolor{highlightColor} \textbf{9.4} (+59.3\%) & \cellcolor{highlightColor} \textbf{10.2} (+59.4\%) & \cellcolor{highlightColor} \textbf{10.2} (+59.4\%)  \\ 
    &    & &         DSGDetr++~\cite{peddi_et_al_scene_sayer_2024} & 5.7 & 6.1 & 6.1  \\ 
    &    & &         \quad+\textbf{\methodname(Ours)} & \cellcolor{highlightColor} \textbf{8.3} (+45.6\%) & \cellcolor{highlightColor} \textbf{8.8} (+44.3\%) & \cellcolor{highlightColor} \textbf{8.8} (+44.3\%)  \\ 
 \cmidrule(lr){4-7}  
     &    &\multirow{8}{*}{Frost} &         STTran+~\cite{peddi_et_al_scene_sayer_2024} & 8.2 & \cellcolor{highlightColor} \textbf{9.0} & \cellcolor{highlightColor} \textbf{9.0}  \\ 
    &    & &         \quad+\textbf{\methodname(Ours)} & \cellcolor{highlightColor} \textbf{8.3} (+1.2\%) & 8.7 (-3.3\%) & 8.7 (-3.3\%)  \\ 
    &    & &         DSGDetr+~\cite{peddi_et_al_scene_sayer_2024} & 9.0 & 9.9 & 9.9  \\ 
    &    & &         \quad+\textbf{\methodname(Ours)} & \cellcolor{highlightColor} \textbf{12.4} (+37.8\%) & \cellcolor{highlightColor} \textbf{13.3} (+34.3\%) & \cellcolor{highlightColor} \textbf{13.3} (+34.3\%)  \\ 
    &    & &         STTran++~\cite{peddi_et_al_scene_sayer_2024} & 9.6 & 10.5 & 10.5  \\ 
    &    & &         \quad+\textbf{\methodname(Ours)} & \cellcolor{highlightColor} \textbf{13.8} (+43.7\%) & \cellcolor{highlightColor} \textbf{15.3} (+45.7\%) & \cellcolor{highlightColor} \textbf{15.3} (+45.7\%)  \\ 
    &    & &         DSGDetr++~\cite{peddi_et_al_scene_sayer_2024} & 9.9 & 10.8 & 10.8  \\ 
    &    & &         \quad+\textbf{\methodname(Ours)} & \cellcolor{highlightColor} \textbf{13.2} (+33.3\%) & \cellcolor{highlightColor} \textbf{14.4} (+33.3\%) & \cellcolor{highlightColor} \textbf{14.4} (+33.3\%)  \\ 
 \cmidrule(lr){4-7}  
     &    &\multirow{8}{*}{Brightness} &         STTran+~\cite{peddi_et_al_scene_sayer_2024} & 11.0 & 12.1 & 12.1  \\ 
    &    & &         \quad+\textbf{\methodname(Ours)} & \cellcolor{highlightColor} \textbf{11.7} (+6.4\%) & \cellcolor{highlightColor} \textbf{12.5} (+3.3\%) & \cellcolor{highlightColor} \textbf{12.5} (+3.3\%)  \\ 
    &    & &         DSGDetr+~\cite{peddi_et_al_scene_sayer_2024} & 12.2 & 13.5 & 13.5  \\ 
    &    & &         \quad+\textbf{\methodname(Ours)} & \cellcolor{highlightColor} \textbf{15.9} (+30.3\%) & \cellcolor{highlightColor} \textbf{17.2} (+27.4\%) & \cellcolor{highlightColor} \textbf{17.2} (+27.4\%)  \\ 
    &    & &         STTran++~\cite{peddi_et_al_scene_sayer_2024} & 12.8 & 14.1 & 14.1  \\ 
    &    & &         \quad+\textbf{\methodname(Ours)} & \cellcolor{highlightColor} \textbf{17.7} (+38.3\%) & \cellcolor{highlightColor} \textbf{19.3} (+36.9\%) & \cellcolor{highlightColor} \textbf{19.3} (+36.9\%)  \\ 
    &    & &         DSGDetr++~\cite{peddi_et_al_scene_sayer_2024} & 13.6 & 14.7 & 14.7  \\ 
    &    & &         \quad+\textbf{\methodname(Ours)} & \cellcolor{highlightColor} \textbf{16.6} (+22.1\%) & \cellcolor{highlightColor} \textbf{18.1} (+23.1\%) & \cellcolor{highlightColor} \textbf{18.1} (+23.1\%)  \\ 
          \hline 
    \end{tabular}
    }
    \vspace{-6.37mm}
\end{table}

\vspace{-2.37mm}
\subsection{Scene Graph Anticipation.}
\vspace{-1.37mm}


We employed a procedure for training and evaluating SGA models that closely mirrors our approach for VidSGG. First, we trained and evaluated the chosen baseline models. We then integrated these models into the \methodname framework, retrained them, and re-evaluated their performance. As demonstrated in Table~\ref{tab:sga_mean_recall_results}, Models trained using \methodname consistently outperformed their baseline variants.

\noindent \underline{\textbf{Insights:}} (1) The results demonstrate an improvement of $\sim$12\% in mR@10 metric across all modes in STTran++ and DSGDetr++. Although the performance is sometimes inferior, it closely aligns with the non-\methodname variants. (2) We also note that SceneSayer ODE/SDE models trained using \methodname demonstrated an improvement of $\sim$20\%.




\vspace{-2.37mm}
\subsection{Robustness Evaluation.}
\vspace{-1.37mm}



We evaluated VidSGG and SGA models trained with \methodname and those trained using traditional approaches across 16 input corruption scenarios.  Our experiments revealed that introducing noise to the inputs generally led to a decline in performance. However, models trained with \methodname exhibited a less severe performance drop than baseline models. For example, in the PREDCLS mode under the No Constraint setting with Gaussian Noise for STTran, the baseline model’s mR@10 decreased from 45.7 to 31.4. In contrast, the \methodname-trained model dropped from 65.5 to 57.5. 

\begin{figure}[!htbp]
  \centering
  \includegraphics[width=\linewidth]{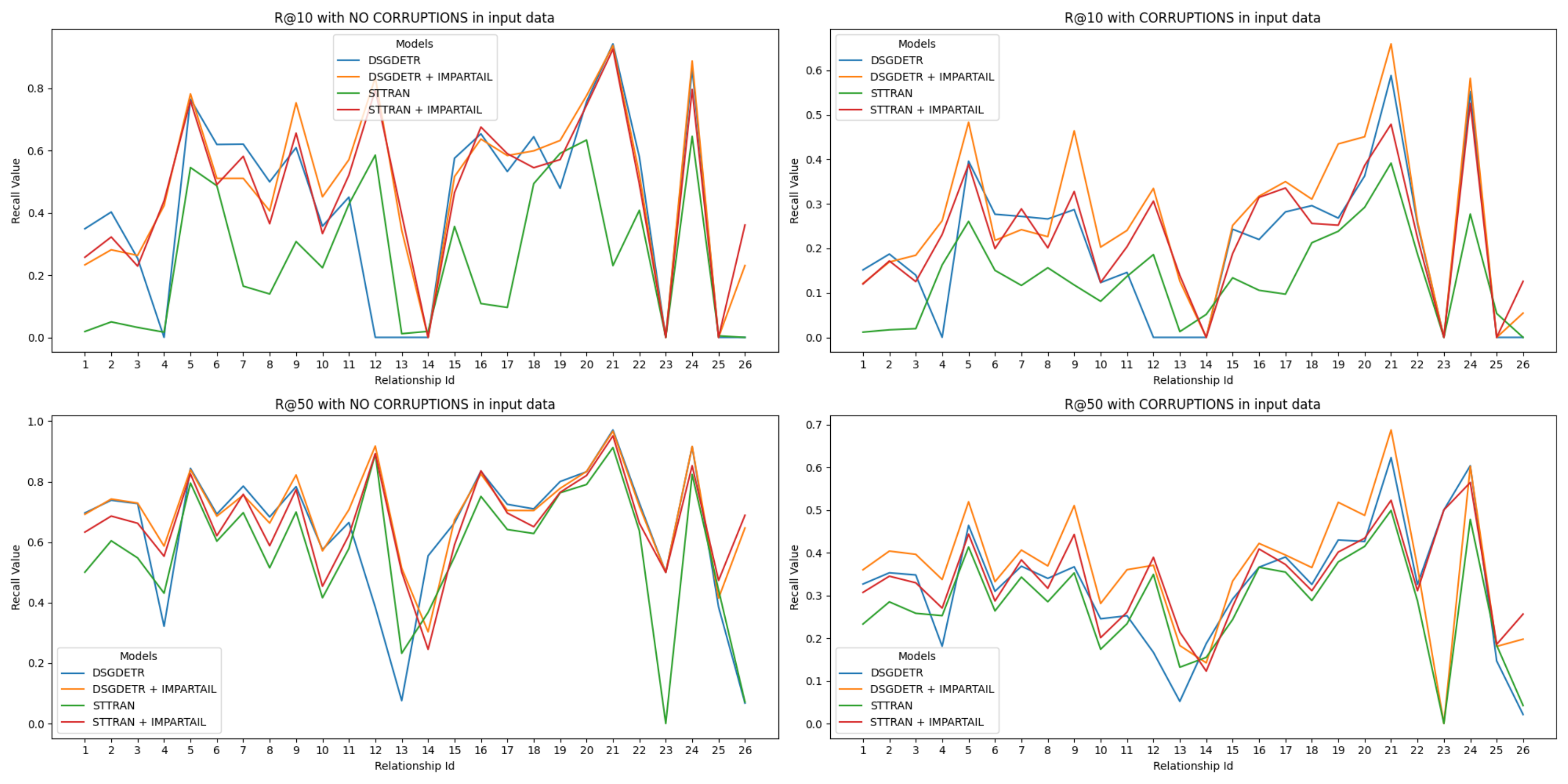}
  \vspace{-6.37mm}
    \caption{\textbf{Predicate Classification recall performance.} of models trained using existing VidSGG methods STTran, DSGDetr, and models trained using their \methodname adaptations. In each row, we compare the R@10/50 performance of each relationship category without corruptions(left) and with corruptions(right) in input data).}
    \label{fig:RecallComparison}
    \vspace{-6.37mm}
\end{figure}




\noindent \underline{\textbf{Insights:}} (1) We noticed that the drop in performance of both VidSGG and SGA models trained using \methodname is lower than the models trained with traditional approaches (see Tables~\ref{tab:sgg_corruptions_results}, \ref{tab:sga_corruptions_results}). (2) We also observed that models trained using \methodname not only exhibited consistent strong mean recall scores but also outperformed baselines in recall metrics sometimes (refer Tab~\ref{tab:sgg_corruptions_results}). (3) \methodname yields models that exhibit: (a) \textbf{Comprehensive Relational Knowledge:} \methodname captures when certain relationships rarely co-occur (e.g., \textit{(sitting\_on, in\_front\_of)}) and when they often appear together (e.g., \textit{(sitting\_on, beneath), (sitting\_on, behind)}). Thus, \methodname trained models focus on a broader spatio-temporal context and other discovered associations to infer missing connections when frames are corrupted. (b) \textbf{Better Generalization:} By balancing out the data distribution, \methodname avoids overfitting to the most frequent object-relationship patterns (refer to Figure~\ref{fig:RecallComparison}). This \textit{regularizing effect} (similar in spirit to dropout) pushes the model to rely on shared patterns across categories, not just the apparent cues in dominant categories. Consequently, when inputs are noisy, the model is more likely to recover the underlying spatio-temporal relationships using generalized features.
\vspace{-2.37mm}
\section{Conclusion} \label{sec:discussion}
\vspace{-2.37mm}

We introduced \methodnamenosp, a framework that employs masked training with partial gradients to shift the focus of learning toward tail classes without sacrificing performance on head classes. To further evaluate the practical applicability of \methodname, we proposed two new tasks: Robust VidSGG and Robust SGA. Future work includes developing robust learning techniques \cite{ilyas2019adversarialexamplesbugsfeatures, peddi_et_al_uai_2022, carlini2019evaluatingadversarialrobustness} for both VidSGG and SGA tasks and exploring the application of unbiased VidSGG and SGA in long-tailed regimes such as error recognition \cite{peddi2024captaincook4ddatasetunderstandingerrors, sener2022assembly101largescalemultiviewvideo}, action anticipation \cite{nawhal2022rethinkinglearningapproacheslongterm}, grasping \cite{xiang2024grasping} etc. 


\section*{Acknowledgements}
This work was supported by IBM AI Horizons Network (AIHN) grant and IBM SUR Award. Parag Singla was supported by Shanthi and K Ananth Krishnan Young Faculty Chair Professorship in AI. 
Rohith Peddi and Vibhav Gogate were supported in part by the DARPA
Perceptually-Enabled Task Guidance (PTG) Program under contract number HR00112220005, by the DARPA Assured Neuro Symbolic Learning and Reasoning (ANSR) Program under contract number HR001122S0039, by the National Science Foundation grant IIS-1652835 and by the AFOSR award
FA9550-23-1-0239. Further, this work was partially supported by the CSE Research Acceleration Fund of IIT Delhi. We would like to thank IIT Delhi HPC facility\footnote{https://supercomputing.iitd.ac.in/}
for computational resources. We would also like to thank Saksham Singh for useful discussions during the initial phase of the project. We would like to express our gratitude to all the reviewers for their insightful comments and queries, which have substantially enhanced the quality of our manuscript.

{
    \small
    \bibliographystyle{ieeenat_fullname}
    \bibliography{main}
}

\newpage

\onecolumn

\addtocontents{toc}{\protect\setcounter{tocdepth}{2}}
\startcontents[sections]
\printcontents[sections]{l}{0.5}{\section*{{Appendices}}}

\newpage
\section{Overview} \label{sec:overview}

\subsection{Motivation}

\begin{enumerate}
    \item \textbf{Long-Tailed Distributions}: Real-world visual relationships are dominated by a few frequent \textit{head classes}, while many rare but critical \textit{tail classes} are underrepresented. This imbalance leads to biased models that fail to generalize effectively across all relationship types, compromising nuanced and accurate scene understanding.
    \item \textbf{Robustness to Distribution Shifts}: Models often struggle with real-world factors such as lighting variations, occlusions, or environmental changes. These distributional shifts degrade performance, limiting the applicability of scene graph models in dynamic and unpredictable environments.
\end{enumerate}

\subsection{Motivational Drivers}
\begin{itemize}
    \item \textbf{Improving Unbiasedness in Scene Understanding}: A long-tailed distribution skews model learning towards frequent classes, leading to biased predictions. Correctly predicting rare relationships is vital for tasks such as autonomous driving, human-robot interaction, and security surveillance, where underrepresented classes can carry critical contextual information.
    \item \textbf{Enhancing Real-World Applicability}: Distributional shifts are unavoidable in real-world scenarios. Ensuring robustness allows STSG models to remain reliable under practical deployment conditions, bolstering trust and usability.
\end{itemize}

\subsection{Contributions}

Thus, concisely, we re-iterate the contributions of the proposed work: 

\begin{itemize}
    \item[--] \textbf{Unbiased Learning with Curriculum-Guided Masking}: The proposed \methodname framework leverages curriculum learning and loss masking to prioritize tail classes progressively during training.
    \item[--] \textbf{Introduction of Robustness Metrics}: Two new tasks—\textit{Robust Spatio-Temporal Scene Graph Generation} and \textit{Robust Scene Graph Anticipation}—evaluate model resilience to input corruptions (a step towards analyzing the performance of STSG models under realistic conditions).
\end{itemize}

\section{Extended Related Work}\label{sec:extended_related_work}

\subsection{Structured Visual Representation}
\noindent \paragraph{Tasks.} Learning to represent static visual data like 2D and 3D images as spatial graphs, with objects as nodes and relationships as edges, is called Image Scene Graph Generation (ImgSGG). This field gained momentum with the foundational Visual Genome project~\cite{krishna_et_al_visual_2017}, advancing 2D ImgSGG research. Building on this foundation, \cite{kim_et_al_3dsgg_2019} expanded the task to encompass static 3D scene data, including RGB and depth information. Object interactions over time provide richer context when dealing with dynamic visual content like videos. Converting such content into structured Spatio-Temporal Scene Graphs (STSGs), where nodes represent objects and edges capture their temporal relationships, is called Video Scene Graph Generation (VidSGG). The research community has concentrated on improving representation learning through sophisticated object-centric architectures like STTran \cite{cong_et_al_sttran_2021} and RelFormer \cite{shit_et_al_relformer_2022}. These include Open-Vocabulary ImgSGG \cite{chen_et_al_expanding_scene_graph_2023}, which expands the range of recognizable objects and relationships. Weakly Supervised ImgSGG \cite{kim_et_al_llm4sgg_2023} to reduce the dependency on extensive labelled data by leveraging weak supervision techniques.
 Panoptic ImgSGG \cite{zhou_et_al_vlprompt_2023} where panoptic segmentation has been integrated to enhance scene graph representations. Zero-Shot ImgSGG \cite{zhao_et_al_less_2023, li2023zeroshotvisualrelationdetection} to enable the detection of unseen visual relationships without explicit labels. Shifting gears from identification and generating scene graphs, recently, Peddi et al.\cite{peddi_et_al_scene_sayer_2024} introduced the Scene Graph Anticipation (SGA) task to anticipate STSGs for future frames. Alongside these developments, foundation models have advanced various ImgSGG task variants \cite{chen_et_al_expanding_scene_graph_2023, kim_et_al_llm4sgg_2023, zhou_et_al_vlprompt_2023, zhao_et_al_less_2023, li2023zeroshotvisualrelationdetection}.

\noindent \paragraph{Unbiased Learning.} TEMPURA \cite{nag_et_al_tempura_2023} and FlCoDe\cite{ khandelwal_correlation_2023} address the challenges posed by long-tailed datasets, such as those found in Action Genome \cite{Ji_2019} and VidVRD \cite{xindi_et_al_vid_vrd_2017} and propose methods for unbiased VidSGG. Specifically, FloCoDe \cite{khandelwal_correlation_2023} mitigates bias by emphasizing temporal consistency and correcting the imbalanced distribution of visual relationships. Similarly, TEMPURA \cite{nag_et_al_tempura_2023} addresses biases in relationship prediction with memory-guided training to generate balanced relationship representations and applies a Gaussian Mixture Model to reduce predictive uncertainty. \underline{\textbf{Note.}} To the best of our knowledge, we are the first to systematically investigate model biases in the SGA task and assess the robustness of both VidSGG and SGA models. In contrast to the above methods, although \methodname shares the goal of training unbiased VidSGG models, it does so without additional architectural components. By modifying the training procedure, \methodname achieves comparable performance and occasionally exceeds the results of these existing methods.

\subsection{Learning Paradigms}

\subsubsection{Long-Tail Learning.} Long-tailed distributions with a few dominant classes (head classes) often overshadow a more significant number of underrepresented ones (tail classes). This class imbalance typically results in models that perform well on head classes but struggle to generalize to tail classes. To mitigate them, the research community has made significant strides in four directions, which include (a) Cost-Sensitive Learning \cite{huang_et_al_2016, wang_et_al_tail_2017, cui_et_al_2019, ridnik_et_al_asl_loss_2020, wu_et_al_distribution_balanced_2020, ouyang_et_al_factors_finetuning_2016, wang_et_al_tail_2017, you_et_al_exemplar_clustering_2018, zhang_et_al_range_loss_2017}, (b) Mixtures-Of-Experts \cite{Xiang2020LearningFM} , (c) Resampling Techniques \cite{chawla_et_al_smote_2002, drummond_et_al_c45_kdd_2003, zhang_et_al_2021}, and (d) Specialized Architectures \cite{zhang_et_al_2021, zhou_et_al_bbn_2019}.

\paragraph{(a) Cost-Sensitive Learning} addresses class imbalance by adjusting the loss function to assign different costs to classes during training. Early approaches involved re-weighting samples inversely proportional to class frequency \cite{huang_et_al_2016, wang_et_al_tail_2017}, but this often led to suboptimal performance on real-world data \cite{cui_et_al_2019}. To improve upon this, advanced methods like label-distribution-aware margin loss with Deferred Re-Weighting (DRW) were proposed \cite{cao_et_al_label_dist_margin_nips_2019}. Equalization Loss (EQL) \cite{tan_et_al_eql_2020} showed that ignoring discouraging gradients for tail classes can prevent adverse effects on model learning. The Class-Balanced (CB) loss \cite{cui_et_al_2019} re-weights the loss based on the effective number of samples per class, achieving notable performance in single-label classification. Asymmetric Loss (ASL) \cite{ridnik_et_al_asl_loss_2020} and Distribution-Balanced (DB) loss \cite{wu_et_al_distribution_balanced_2020} focus on balancing positive and negative labels in multi-label classification. Other approaches include transferring knowledge from head to tail classes \cite{ouyang_et_al_factors_finetuning_2016, wang_et_al_tail_2017} and designing better training objectives through metric learning \cite{you_et_al_exemplar_clustering_2018, zhang_et_al_range_loss_2017}.

\paragraph{(b) Mixture-of-Experts (MoE)} methods tackle class imbalance by distributing samples among specialized expert models. LFME \cite{Xiang2020LearningFM} merges multiple experts using self-paced knowledge distillation, while RIDE employs diversity loss and dynamic routing for sample assignment.

\paragraph{ (c) Resampling methods} adjust the training data distribution by over-sampling tail classes or under-sampling head classes. Techniques like SMOTE \cite{chawla_et_al_smote_2002} create synthetic samples for minority classes, while under-sampling methods reduce samples from majority classes \cite{drummond_et_al_c45_kdd_2003}. Approaches, such as concatenating frames from different video clips \cite{zhang_et_al_2021}, offer a different way to balance data, particularly spatio-temporal data. 

\paragraph{(d) Specialized architectures}  aim to enhance feature representation and aggregation for tail classes. FrameStack \cite{zhang_et_al_2021} performs frame-level sampling guided by running average precision to improve tail class representation without explicitly differentiating classes at the feature level. The Bilateral-Branch Network (BBN) \cite{zhou_et_al_bbn_2019} uses cumulative learning to balance representation learning and classifier discrimination. 
Kang et al. \cite{kang_et_al_iclr_2019} demonstrated that decoupling representation learning from classifier training prevents head classes from overshadowing tail classes. 
While Kang et al. \cite{kang_et_al_iclr_2019} argued that resampling might not always be necessary if classifier training is focused correctly, Zhou et al. \cite{zhou_et_al_bbn_2019} showed that standard resampling could harm representation learning. 
Li et al. \cite{li_et_al_gaussian_clouded_cvpr_2022} proposed Gaussian Clouded Logit Adjustment to perturb class logits, adjusting decision boundaries for better generalization across classes.

\subsubsection{Curriculum Learning (CL)} is a training methodology that structures training by presenting simpler examples first and progressively introducing more complex ones. This approach aims to enhance learning efficiency by aligning the difficulty of training data with the model's learning capacity at each stage \footnote{It's important to differentiate CL from other dynamic sampling techniques such as self-paced learning \cite{kumar_et_al_self_paced_2010}, boosting \cite{freund_boosting_1997}, hard example mining \cite{shrivastava_et_al_hard_example_mining_2016}, and active learning \cite{schein_et_al_al_2005}. While these methods also adjust the training data based on certain criteria, they typically rely on the model's current performance or hypotheses to select samples rather than following a predefined difficulty progression as in CL.}\cite{kumar_et_al_self_paced_2010, graves_et_al_auto_cr_2017, hacohen_et_al_cl_tdn_2019, han_et_al_dyn_een_2022}. Despite its potential benefits, implementing CL presents significant challenges. A primary obstacle is distinguishing between \textit{easy} and \textit{hard} training samples. This differentiation often requires additional mechanisms, such as auxiliary neural networks acting as \textit{teachers} or specialized algorithms. Difficulty measures can be predefined based on certain heuristics \cite{ionescu_et_al_diff_vis_search_2016} or learned automatically during the training process \cite{kumar_et_al_self_paced_2010, jiang_et_al_easy_samples_2014, weinshall_et_al_cr_2018, jiang_et_al_mentornet_2017, ren_et_al_ltrw_2018, hacohen_et_al_cl_tdn_2019, matiisen_et_al_teacher_student_2017}. Alongside difficulty assessment, a scheduling strategy is essential to determine when and how to introduce more challenging data \cite{graves_et_al_auto_cr_2017}. The \textit{starting small} concept influences our methodology \cite{elman_starting_small_1993}, which recommends initiating learning with easier tasks. Unlike conventional Curriculum Learning (CL) methods that introduce data progressively, our approach utilizes the entire training dataset from the start. We adopt label selection to mask the loss function, offering a unique strategy that impacts the learning process while keeping all training examples in play. This approach not only streamlines the implementation of CL but also tackles the difficulties of determining and scheduling the complexity of training data.


\section{Limitations} \label{sec:limitations}

\begin{enumerate}
    \item \textbf{Limited Scope of Datasets:} Experiments are primarily conducted on the Action Genome dataset. 
    \begin{itemize}
        \item This is a primary concern of the field as the Action Genome is the only large-scale dataset available as a testbed for the Spatio-Temporal Scene Graph tasks.
    \end{itemize}
    \item \textbf{Model Robustness to Distribution Shifts:}  
    \begin{itemize}
        \item Although robustness is considered, the specific test corruptions may not cover all real-world scenarios. Instead, our work can be considered a starting point for further developing robust learning techniques. 
    \end{itemize}
    \item \textbf{Bias Mitigation vs. Performance:} 
    \begin{itemize}
        \item In \methodname, balancing unbiased learning with high performance on head classes, although small, might result in a trade-off between performance over head and tail classes. We conjecture that adding an external memory block to our framework can help mitigate this issue. 
    \end{itemize}
    \item \textbf{Limited Evaluation Metrics:} 
    \begin{itemize}
        \item Although metrics such as recall and mean recall provide us insights about the performance of the trained models. These might fail to capture the performance over higher-order spatial and temporal relationships. 
    \end{itemize}
\end{enumerate}

\newpage
\section{Approach} \label{sec:approach}

\subsection{\methodname}

Here, we present the complete algorithm for the proposed unbiased learning framework. Our framework has four key components. 

\begin{enumerate}
    \item (\RNum{1}) \textbf{Object Representation Processing Unit (ORPU):} This module extracts object representations for detected objects within video frames using a pre-trained object detector.
    \item (\RNum{2}) \textbf{Spatio-Temporal Context Processing Unit (STPU):} This unit creates object-centric relationship representations, tailored for two tasks: (i) \textit{observed relationships} for VidSGG and (ii) \textit{anticipated relationships} for SGA.
    \item (\RNum{3}) \textbf{Relationship Predicate Decoders:} These decoders assign predicate labels to the observed or anticipated relationship representations. \underline{\textbf{Note:}} ORPU, STPU, and the predicate decoders can be adapted from any VidSGG or SGA method following an object-centric framework.
    \item (\RNum{4}) \textbf{Curriculum-Guided Masked Loss:} This loss mechanism employs a curriculum-based masking strategy to exclude certain relationship predicate labels during training progressively. Focusing on underrepresented classes helps the model achieve a balanced class distribution.
\end{enumerate}

\paragraph{(\RNum{4}) Curriculum-Guided Masked Loss.} This has two components as explained in the main paper: (a) \textbf{Curriculum-Guided Mask Generation} and (b) \textbf{Masked Predicate Classification Loss}. First, we provide the complete algorithm for \textit{Mask Generation} and give an overview of the loss function employed. In the subsequent sections, we clearly explain and contrast the loss functions for the original and the proposed \methodname variants.

\subsubsection{Curriculum Guided Mask Generation.}
\noindent
\begin{minipage}[t]{0.46\textwidth}

\begin{algorithm}[H]
\DontPrintSemicolon
\footnotesize{
\KwIn{Epoch: $e$, Sampling Ratio: $\mathscr{R}_{s}$, Dataset Annotations: $\mathcal{D}$, Total predicate labels: $N$, Total predicates: $\mathcal{P}$, Videos: $\mathscr{V}$}
\KwOut{Filtered Dataset: $\mathcal{F}$}

\textcolor{blue}{*** Determine Target Counts \textbf{[can also be a fixed input]} ***}

$\mathscr{R}_{m} = e \times \mathscr{R}_{s}$ \quad \textcolor{blue}{** Masking Ratio **} \;

$N_{\text{target}} = \operatorname{round}(N \times \mathscr{R}_{m})$\;
    
\textcolor{blue}{*** Curriculum-based sampling probabilities $Prob[rel]$ ***}\;
    
\textcolor{blue}{*** Equally weighted distribution \textbf{[can also be learnt]} *** }\;

Set $Prob[rel] = \frac{1}{|\mathcal{P}|}$ \;
Sample target counts $Tar[rel]$ from Multinomial distribution: $Tar[rel] \sim \operatorname{Multinomial}(N_{\text{target}}, Prob[rel])$ \;

\textcolor{blue}{*** Randomly sample instances of relationships in the dataset based on the target counts and construct filtered dataset ***} \;

Initialize positions $\mathscr{P}[rel]$ to collect occurrences of $rel$\;
\ForEach{$v$, $f$ in $\mathcal{D}$}{
    \ForEach{Relation $rel$ in $\mathcal{D}[v][f]$}{
        Append position $(v, f, \text{index})$ to $\mathscr{P}[rel]$\;
    }
}
Initialize empty set $\mathcal{K}$\;

\textcolor{blue}{*** These relations are ignored and are omitted from the filtered dataset constructed below ***} \;

\ForEach{Relation $rel$}{
    Randomly select $Tar[rel]$ positions which should be masked from $\mathscr{P}[rel]$ and add the remaining to $\mathcal{K}$\;
}

\textcolor{blue}{*** Filter Data ***} \;
    Initialize filtered data $\mathcal{F}$\;
    \ForEach{$v$, $f$ in $\mathcal{D}$}{
        Initialize filtered frame $\mathcal{F}[v][f]$\;
        \ForEach{Relation at position $(v, f, i)$}{
            \If{$(v, f, i) \in \mathcal{K}$}{
                Add relation to $\mathcal{F}[v][f]$\;
            }
        }
    }

}
\caption{Filtered Dataset Construction}
\label{alg:cur_guided_mask_generation}
\end{algorithm}
\end{minipage}%
\hfill
\begin{minipage}[t]{0.46\textwidth}
\begin{algorithm}[H]
\DontPrintSemicolon
\footnotesize{
\KwIn{Epoch: $e$, Dataset Annotations: $\mathcal{D}$, Videos: $\mathscr{V}$, Filtered Dataset: $\mathcal{F}$}
\KwOut{Masks: $\mathcal{M}$}
Initialize empty mask list $\mathcal{M}$\;
\ForEach{Video $v$ in $\mathscr{V}$}{
    Initialize video mask $\mathcal{M}_v$\;
    \ForEach{Frame $f$ in $v$}{
        Initialize frame mask $\mathcal{M}_f$\;
        \ForEach{Object $o$ in $f$}{
            Initialize object mask $\mathcal{M}_o$\;
            Original relations $\mathcal{R}_{o}$ from $\mathcal{D}[v][f]$\;
            \ForEach{Relation $rel$ in $\mathcal{R}_{o}$}{
                \eIf{$rel \in \mathcal{F}[v][f]$}{
                    Set mask value $\mathcal{M}_o[rel] = 0$\;
                }{
                    Set mask value $\mathcal{M}_o[rel] = 1$\;
                }
            }
            Add $\mathcal{M}_o$ to $\mathcal{M}_f$\;
        }
        Add $\mathcal{M}_f$ to $\mathcal{M}_v$\;
    }
    Add $\mathcal{M}_v$ to $\mathcal{M}^{(e)}$\;
}

}
\caption{Mask Generation}
\label{alg:cur_guided_mask_generation}
\end{algorithm}
\end{minipage}


\clearpage

\subsection{Video Scene Graph Generation}

\subsubsection{\methodname + STTran}

\begin{figure*}[!htbp]
    \centering
    \includegraphics[width=\textwidth]{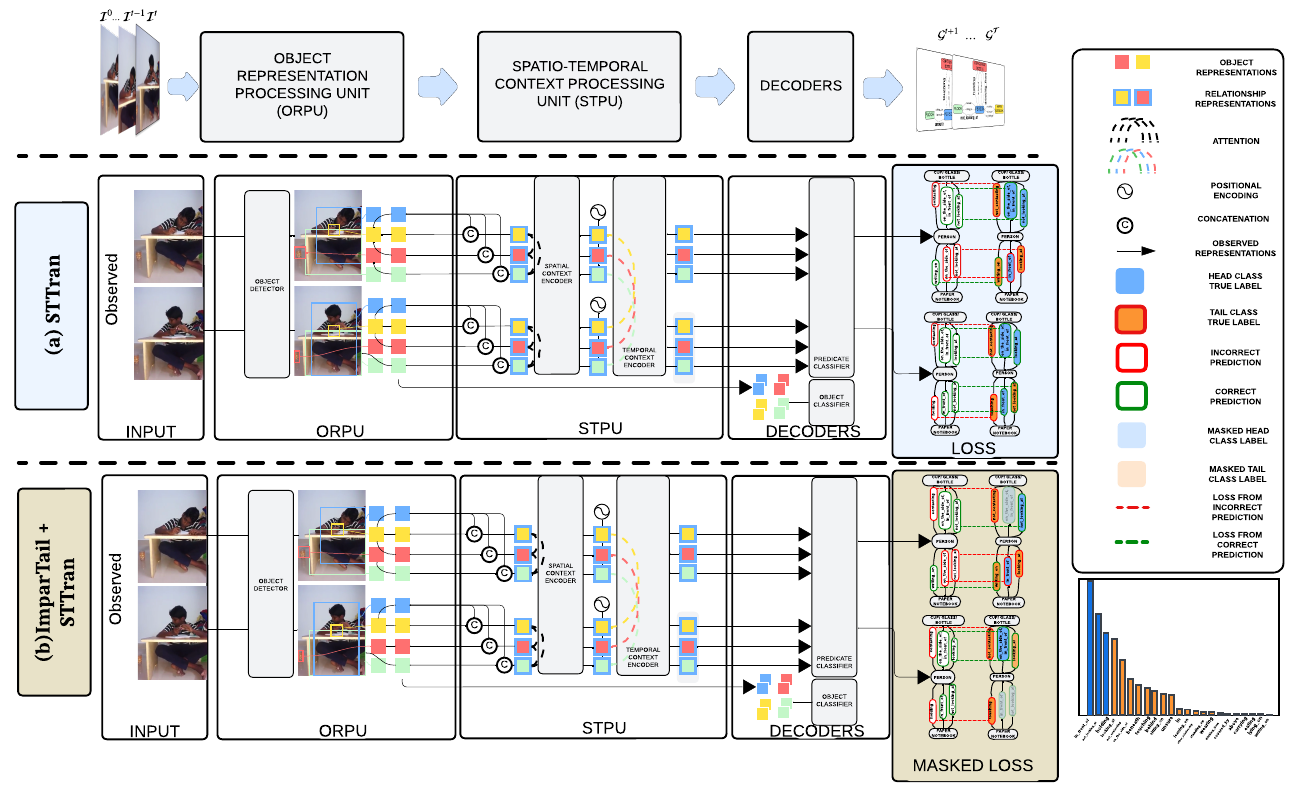}
   \caption{\underline{(a) \textbf{Architectural Components.}} In \textbf{STTran}, the Object Representations Processing Unit (ORPU) primarily consists of an object detector and the visual features output by the object detector. Then, the Spatio-Temporal Context Processing Unit (STPU )takes these visual features as input and first constructs relationship representations utilizing the features of interacting objects; then, these relationship representations are fed to a spatial encoder and a transformer encoder. Thus, the spatio-temporal context-aware representations output by the temporal encoder are fed into the predicate classifier for final predictions. Finally, these representations are decoded for predicate classification. \underline{(b) \textbf{Loss Function.}} The primary difference between \colorbox{\baselinecolor}{STTran} loss and the proposed \colorbox{\methodcolor}{\methodname + STTran} loss is illustrated using highlighting the employed losses. We do not mask any predicate label in \colorbox{\baselinecolor}{STTran} loss. In contrast, in the proposed \colorbox{\methodcolor}{\methodname + STTran} loss, we mask the losses corresponding to the \textit{head} classes as generated by following the curriculum-based strategy.}
    \label{fig:sup_sttran}
\end{figure*}

\paragraph{Loss Function \colorbox{\baselinecolor}{STTran}}

\begin{gather*} 
    \underbrace{
        \mathcal{L}_{i} = \sum_{t=1}^{\bar{T}} \mathcal{L}_{i}^{t}
    }_\text{(1)}; \quad
    \underbrace{
        \textcolor{red}{\mathcal{L}_{\text{gen}} = \sum_{t=1}^{\bar{T}} \mathcal{L}_{\text{gen}}^t, 
        \mathcal{L}_{\text{gen}}^t = \sum_{ij} \mathcal{L}_{p^t_{ij}}}
    }_\text{Predicate Classification Loss (2)}
    \quad \\
    \mathcal{L} = 
        \sum_{t=1}^{\bar{T}} \left(\lambda_{1} \textcolor{red}{\mathcal{L}_{\text{gen}}^{t}} + 
        \lambda_{2} \sum_{i} \mathcal{L}_{i}^{t} \right)    
\end{gather*}

\paragraph{Loss Function \colorbox{\methodcolor}{\methodname + STTran}}

\begin{gather*} 
    \underbrace{
        \mathcal{L}_{i} = \sum_{t=1}^{\bar{T}} \mathcal{L}_{i}^{t}
    }_\text{(1)}; \quad
    \underbrace{
        \textcolor{olive}{\mathscr{L}_{\text{gen}} = \sum_{t=1}^{\bar{T}} \mathscr{L}_{\text{gen}}^t, 
        \mathscr{L}_{\text{gen}}^t = \sum_{ij} \mathbf{m}^t_{ij} * \mathcal{L}_{p^t_{ij}}}
    }_\text{\textbf{Masked} Predicate Classification Loss (2)}
    \quad \\
    \mathcal{L} = 
        \sum_{t=1}^{\bar{T}} \left(\lambda_{1} \textcolor{olive}{\mathscr{L}_{\text{gen}}^{t}} + 
        \lambda_{2} \sum_{i} \mathcal{L}_{i}^{t} \right)    
\end{gather*}

Here, 

\begin{equation}
    \label{eq:PredicateClassification}
    \hat{\mathbf{p}}^t_{ij} = \operatorname{PredClassifier}_{\text{observed}}\left(\mathbf{z}^t_{ij} \right), \forall t \in [1, T] 
\end{equation}

\noindent\textbf{Predicate Classification Loss ($\mathcal{L}_{\text{gen}}$).} focuses on classifying the relationship representations between pairs of objects \objpair across all frames (\(t \in [1, \bar{T}]\)) as detailed above. Here, $\mathcal{L}_{p^t_{ij}}$ represents multi-label margin loss and is computed as follows:

\begin{equation} \label{eq:predicate_classification_loss}
    \mathcal{L}_{p^t_{ij}}=\sum_{u \in \mathcal{P}^{+}} \sum_{v \in \mathcal{P}^{-}} \max (0,1-\hat{\mathbf{p}}^t_{ij}[v]+\hat{\mathbf{p}}^t_{ij}[u])
\end{equation}

\paragraph{Implementation Details.}

\begin{itemize}
    \item[--] \textbf{Training Epochs.} We have capped the number of training epochs for both models where one uses conventional loss and the other uses the proposed \methodname framework to \textbf{5} epochs.  
    \item[--] \textbf{Loss Function.} Results reported in the literature for the method \colorbox{\baselinecolor}{STTran} were not reproducible using the Multi-Label Margin Loss. However, we noticed we could reach closer numbers (still lower than reported) by employing BCE Loss and training to 10 epochs. 
    \item[--] \textbf{Hyperparameters.} We use the same hyperparameter settings described in the paper. 
\end{itemize}

\paragraph{\colorbox{highlightColor}{\textbf{Insight.}}} Our reported mean recall numbers closely match the numbers reported by the SOTA model TEMPURA \cite{nag_et_al_tempura_2023} without any additional architectural changes just by changing how the model is learnt. We also emphasize that although our recall performance was hurt slightly, it is marginally lower than recall values compared to the original model and TEMPURA.

\clearpage

\subsubsection{\methodname + DSGDetr}

\begin{figure*}[!htbp]
    \centering
    \includegraphics[width=\textwidth]{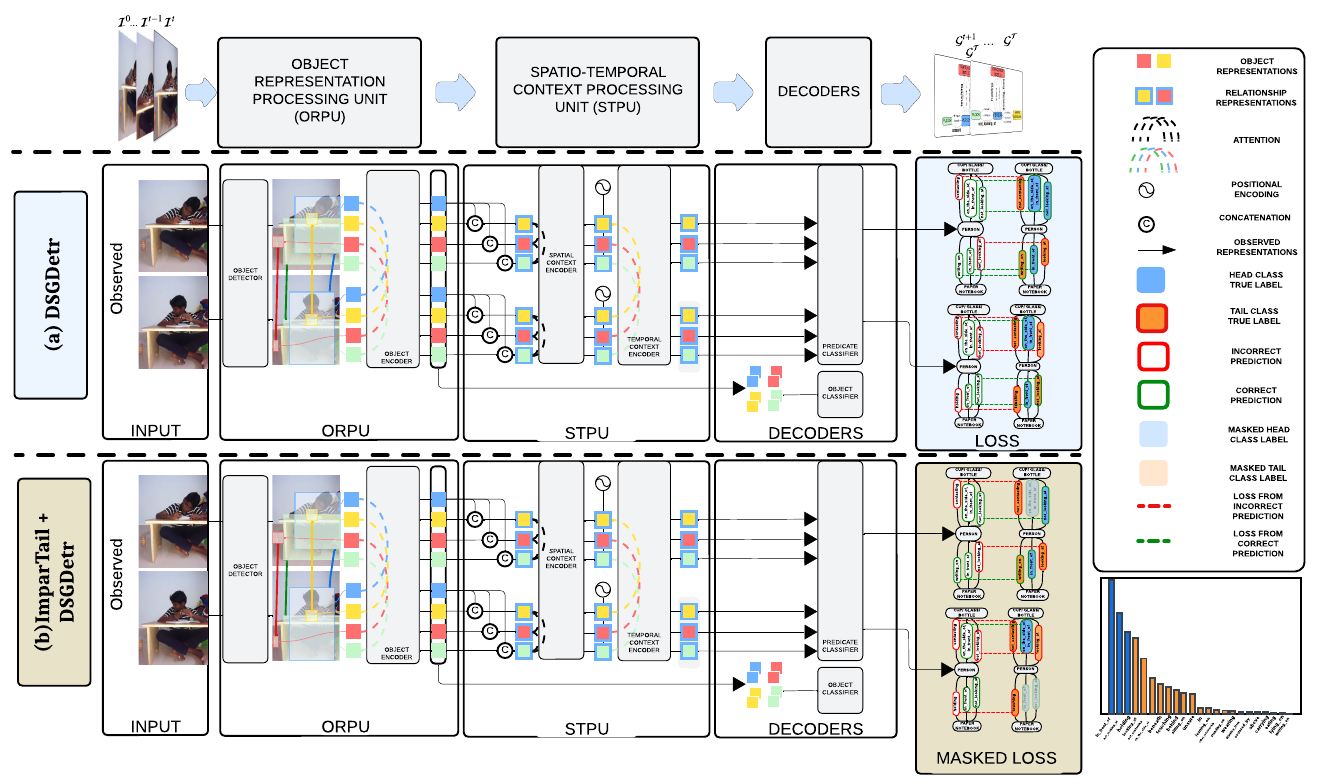}
    \caption{\underline{(a) \textbf{Architectural Components.}} In \textbf{DSGDetr}, the Object Representations Processing Unit (ORPU) primarily consists of an object detector, \textcolor{teal}{an object tracker and an object encoder. The visual features output by the object detector are used to construct tracklets corresponding to each object, and these representations are further enhanced by passing them through an object encoder}. Then, the Spatio-Temporal Context Processing Unit (STPU )takes these visual features as input and first constructs relationship representations utilizing the features of interacting objects; then, these relationship representations are fed to a spatial encoder and a transformer encoder. Thus, the spatio-temporal context-aware representations output by the temporal encoder are fed into the predicate classifier for final predictions. Finally, these representations are decoded for predicate classification. \underline{(b) \textbf{Loss Function.}} The primary difference between \colorbox{\baselinecolor}{DSGDetr} loss and the proposed \colorbox{\methodcolor}{\methodname + DSGDetr} loss is illustrated using highlighting the employed losses. We do not mask any predicate label in \colorbox{\baselinecolor}{DSGDetr} loss. In contrast, in the proposed \colorbox{\methodcolor}{\methodname + DSGDetr} loss, we mask the losses corresponding to the \textit{head} classes as generated by following the curriculum-based strategy.}
    \label{fig:enter-label}
\end{figure*}

\paragraph{Loss Function \colorbox{\baselinecolor}{DSGDetr}}

\begin{gather*} 
    \underbrace{
        \mathcal{L}_{i} = \sum_{t=1}^{\bar{T}} \mathcal{L}_{i}^{t}
    }_\text{(1)}; \quad
    \underbrace{
        \textcolor{red}{\mathcal{L}_{\text{gen}} = \sum_{t=1}^{\bar{T}} \mathcal{L}_{\text{gen}}^t, 
        \mathcal{L}_{\text{gen}}^t = \sum_{ij} \mathcal{L}_{p^t_{ij}}}
    }_\text{Predicate Classification Loss (2)}
    \quad \\
    \mathcal{L} = 
        \sum_{t=1}^{\bar{T}} \left(\lambda_{1} \textcolor{red}{\mathcal{L}_{\text{gen}}^{t}} + 
        \lambda_{2} \sum_{i} \mathcal{L}_{i}^{t} \right)    
\end{gather*}

\paragraph{Loss Function \colorbox{\methodcolor}{\methodname + DSGDetr}}

\begin{gather*} 
    \underbrace{
        \mathcal{L}_{i} = \sum_{t=1}^{\bar{T}} \mathcal{L}_{i}^{t}
    }_\text{(1)}; \quad
    \underbrace{
        \textcolor{olive}{\mathscr{L}_{\text{gen}} = \sum_{t=1}^{\bar{T}} \mathscr{L}_{\text{gen}}^t, 
        \mathscr{L}_{\text{gen}}^t = \sum_{ij} \mathbf{m}^t_{ij} * \mathcal{L}_{p^t_{ij}}}
    }_\text{\textbf{Masked} Predicate Classification Loss (2)}
    \quad \\
    \mathcal{L} = 
        \sum_{t=1}^{\bar{T}} \left(\lambda_{1} \textcolor{olive}{\mathscr{L}_{\text{gen}}^{t}} + 
        \lambda_{2} \sum_{i} \mathcal{L}_{i}^{t} \right)    
\end{gather*}

Here, 

\begin{equation}
    \label{eq:PredicateClassification}
    \hat{\mathbf{p}}^t_{ij} = \operatorname{PredClassifier}_{\text{observed}}\left(\mathbf{z}^t_{ij} \right), \forall t \in [1, T] 
\end{equation}

\noindent\textbf{Predicate Classification Loss ($\mathcal{L}_{\text{gen}}$).} focuses on classifying the relationship representations between pairs of objects \objpair across all frames (\(t \in [1, \bar{T}]\)) as detailed above. Here, $\mathcal{L}_{p^t_{ij}}$ represents multi-label margin loss and is computed as follows:

\begin{equation} \label{eq:predicate_classification_loss}
    \mathcal{L}_{p^t_{ij}}=\sum_{u \in \mathcal{P}^{+}} \sum_{v \in \mathcal{P}^{-}} \max (0,1-\hat{\mathbf{p}}^t_{ij}[v]+\hat{\mathbf{p}}^t_{ij}[u])
\end{equation}

\paragraph{Implementation Details.}

\begin{itemize}
    \item[--] \textbf{Training Epochs.} We have capped the number of training epochs for both models where one uses conventional loss and the other uses the proposed \methodname framework to \textbf{5} epochs.  
    \item[--] \textbf{Loss Function.} Results reported in the literature for the method \colorbox{\baselinecolor}{DSGDetr} were not reproducible using the Multi-Label Margin Loss. However, we noticed we could reach closer numbers (still lower than reported) by employing BCE Loss and training to 10 epochs. 
    \item[--] \textbf{Hyperparameters.} We use the same hyperparameter settings described in the paper. 
\end{itemize}

\paragraph{\colorbox{highlightColor}{\textbf{Insight.}}} Our reported mean recall numbers closely match the numbers reported by the SOTA model TEMPURA \cite{nag_et_al_tempura_2023} without any additional architectural changes just by changing how the model is learnt. We also emphasize that although our recall performance was hurt slightly, it is marginally lower than recall values compared to the original model and TEMPURA.

\clearpage

\subsection{Scene Graph Anticipation}

\subsubsection{\methodname + STTran++}

\begin{figure*}[!htbp]
    \centering
    \includegraphics[width=\textwidth]{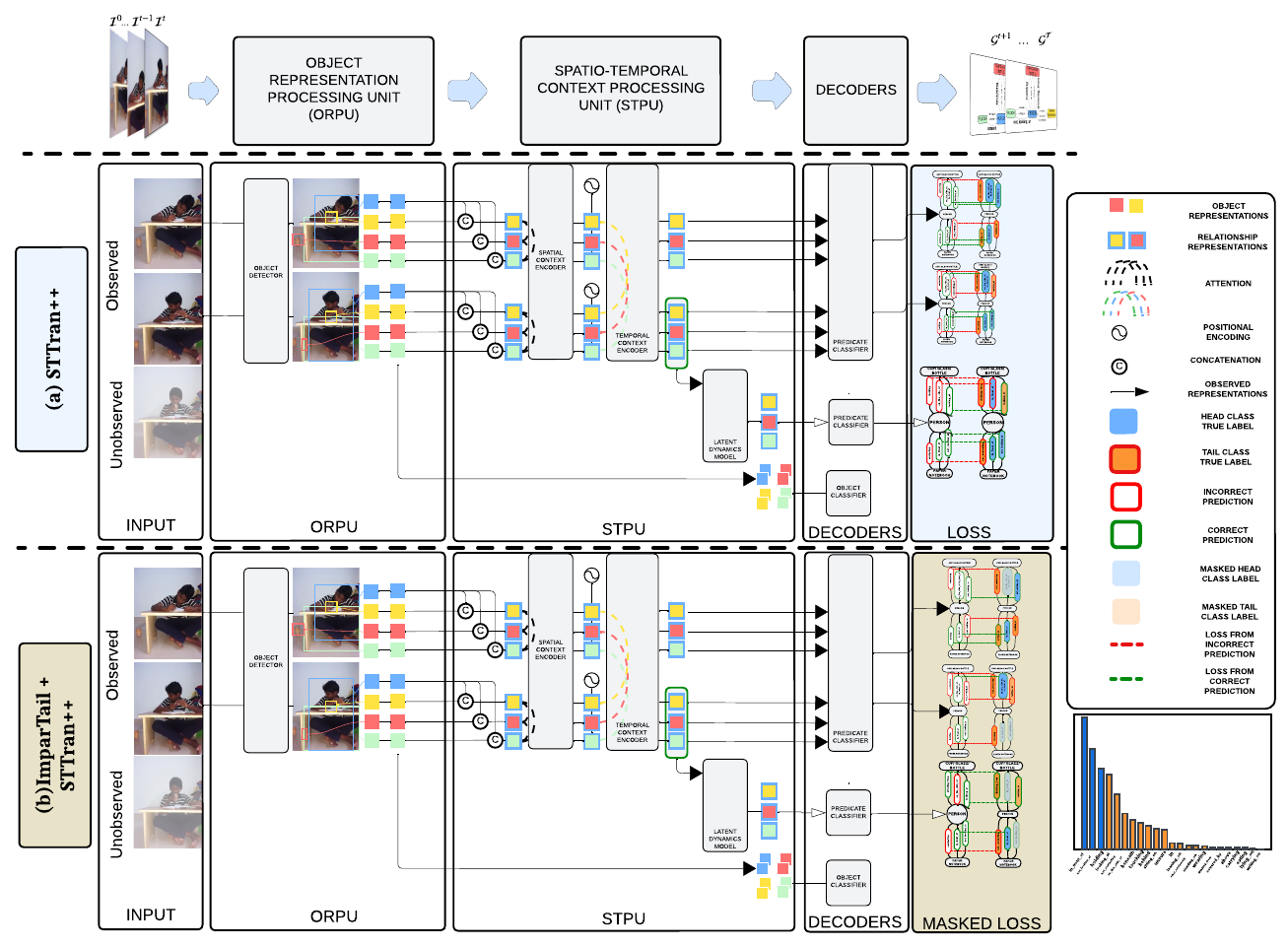}
    \caption{\underline{(a) \textbf{Architectural Components.}} In \textbf{STTran++}, the Object Representations Processing Unit (ORPU) primarily consists of an object detector and the visual features output by the object detector. Then, the Spatio-Temporal Context Processing Unit (STPU) takes these visual features as input and first constructs relationship representations utilizing the features of interacting objects; then, these relationship representations are fed to a spatial encoder and a transformer encoder. Thus, the spatio-temporal context-aware representations output by the temporal encoder are fed as input to another transformer encoder to anticipate the future relationship representations corresponding to interacting objects. Thus, relationship representations from the temporal encoder and future relationship representations from the anticipatory transformer encoder are input to two predicate classifiers for final predictions. \underline{(b) \textbf{Loss Function.}} The primary difference between \colorbox{\baselinecolor}{STTran++} loss and the proposed \colorbox{\methodcolor}{\methodname + STTran++} loss is illustrated using highlighting the employed losses. We do not mask any predicate label in \colorbox{\baselinecolor}{STTran++} loss. In contrast, in the proposed \colorbox{\methodcolor}{\methodname + STTran++} loss, we mask the losses corresponding to the \textit{head} classes output by predicate classification heads corresponding to both observed and anticipated relationship representations.}
    \label{fig:enter-label}
\end{figure*}

\paragraph{Loss Function \colorbox{\baselinecolor}{STTran++}}

\begin{equation}
    \overbrace{
        \mathcal{L}_{i} = \sum_{t=1}^{\bar{T}} \mathcal{L}_{i}^{t},
        \quad
        \mathcal{L}_{i}^{t} = -\sum_{n=1}^{|\mathcal{C}|} y_{i,n}^{t} \log(\hat{\mathbf{c}}_{i,n}^{t})
    }^\text{Object Classification Loss (\RNum{1})}
\end{equation}

\begin{equation}
    \overbrace{
        \textcolor{red}{\mathcal{L}_{\text{gen}} = \sum_{t=1}^{\bar{T}} \mathcal{L}_{\text{gen}}^t, 
        \quad
        \mathcal{L}_{\text{gen}}^t = \sum_{ij} \mathcal{L}_{p^t_{ij}}}
    }^\text{Predicate Classification Loss (\RNum{2})}
\end{equation}

\begin{equation}
     \underbrace{
        \textcolor{red}{\mathcal{L}_{\text{ant}}^{(1:T)} = \sum_{t=T+1}^{\min(T+H, \bar{T})} \mathcal{L}_{\text{ant}}^t, 
        \quad
        \mathcal{L}_{\text{ant}}^t = \sum_{ij} \mathcal{L}_{p^t_{ij}}}
    }_\text{Predicate Classification Loss (\RNum{3})}
\end{equation}

\begin{equation}
    \underbrace{
        \mathcal{L}_{\text{recon}}^{(1:T)} = \sum_{t = T+1}^{\min(T+H, \bar{T})} \mathcal{L}_{\text{recon}}^{t},
        \quad
        \mathcal{L}_{\text{recon}}^t = \frac{1}{N(t) \times N(t)} \sum_{ij}^ {(N(t) \times N(t))}\text{L}_{\text{smooth}}(\mathbf{z}_{ij}^t - \hat{\mathbf{z}}_{ij}^t)
    }_\text{Reconstruction Loss (\RNum{4})}
\end{equation}

Thus, the total objective for training the proposed method can be written as:

\begin{equation}
    \mathcal{L} = 
    \underbrace{
        \sum_{t=1}^{\bar{T}} \left(\lambda_{1} \textcolor{red}{\mathcal{L}_{\text{gen}}^{t}} + 
        \lambda_{2} \sum_{i} \mathcal{L}_{i}^{t} \right)
    }_\text{Loss Over Observed Representations} 
    + 
    \underbrace{
        \sum_{T=3}^{\bar{T}-1} \left( 
            \lambda_{3} \textcolor{red}{\mathcal{L}_{\text{ant}}^{(1:T)}} +
            \lambda_{4} \mathcal{L}_{\text{recon}}^{(1:T)}
        \right)
    }_\text{Loss Over Anticipated Representations}
\end{equation}

\paragraph{Loss Function \colorbox{\methodcolor}{\methodname + STTran++}}

\begin{equation}
    \overbrace{
        \mathcal{L}_{i} = \sum_{t=1}^{\bar{T}} \mathcal{L}_{i}^{t},
        \quad
        \mathcal{L}_{i}^{t} = -\sum_{n=1}^{|\mathcal{C}|} y_{i,n}^{t} \log(\hat{\mathbf{c}}_{i,n}^{t})
    }^\text{Object Classification Loss (\RNum{1})}
\end{equation}

\begin{equation}
    \overbrace{
        \textcolor{olive}{\mathscr{L}_{\text{gen}} = \sum_{t=1}^{\bar{T}} \mathscr{L}_{\text{gen}}^t, 
        \quad
        \mathscr{L}_{\text{gen}}^t = \sum_{ij} \mathbf{m}^t_{ij} * \mathcal{L}_{p^t_{ij}}}
    }^\text{Predicate Classification Loss (\RNum{2})}
\end{equation}

\begin{equation}
     \underbrace{
        \textcolor{olive}{\mathscr{L}_{\text{ant}}^{(1:T)} = \sum_{t=T+1}^{\min(T+H, \bar{T})} \mathscr{L}_{\text{ant}}^t, 
        \quad
        \mathscr{L}_{\text{ant}}^t = \sum_{ij} \mathbf{m}^t_{ij} * \mathcal{L}_{p^t_{ij}}}
    }_\text{Predicate Classification Loss (\RNum{3})}
\end{equation}

\begin{equation}
    \underbrace{
        \mathcal{L}_{\text{recon}}^{(1:T)} = \sum_{t = T+1}^{\min(T+H, \bar{T})} \mathcal{L}_{\text{recon}}^{t},
        \quad
        \mathcal{L}_{\text{recon}}^t = \frac{1}{N(t) \times N(t)} \sum_{ij}^ {(N(t) \times N(t))}\text{L}_{\text{smooth}}(\mathbf{z}_{ij}^t - \hat{\mathbf{z}}_{ij}^t)
    }_\text{Reconstruction Loss (\RNum{4})}
\end{equation}

Thus, the total objective for training the proposed method can be written as:

\begin{equation}
    \mathcal{L} = 
    \underbrace{
        \sum_{t=1}^{\bar{T}} \left(\lambda_{1} \textcolor{olive}{\mathscr{L}_{\text{gen}}^{t}} + 
        \lambda_{2} \sum_{i} \mathcal{L}_{i}^{t} \right)
    }_\text{Loss Over Observed Representations} 
    + 
    \underbrace{
        \sum_{T=3}^{\bar{T}-1} \left( 
            \lambda_{3} \textcolor{olive}{\mathscr{L}_{\text{ant}}^{(1:T)}} +
            \lambda_{4} \mathcal{L}_{\text{recon}}^{(1:T)}
        \right)
    }_\text{Loss Over Anticipated Representations}
\end{equation}

\paragraph{Implementation Details.}

\begin{itemize}
    \item[--] \textbf{Training Epochs.} We have capped the number of training epochs for both models, where one uses conventional loss, and the other uses the proposed \methodname framework to \textbf{5} epochs.  
    \item[--] \textbf{Loss Function.} Results reported in the literature for the method \colorbox{\baselinecolor}{STTran++} were reproducible using the Multi-Label Margin Loss. We sometimes achieved higher numbers than those reported in the original paper. 
    \item[--] \textbf{Hyperparameters.} We use the same hyperparameter settings described in the paper\cite{peddi_et_al_scene_sayer_2024}. 
\end{itemize}

\subsubsection{\methodname + DSGDetr++}

\begin{figure*}[!htbp]
    \centering
    \includegraphics[width=\textwidth]{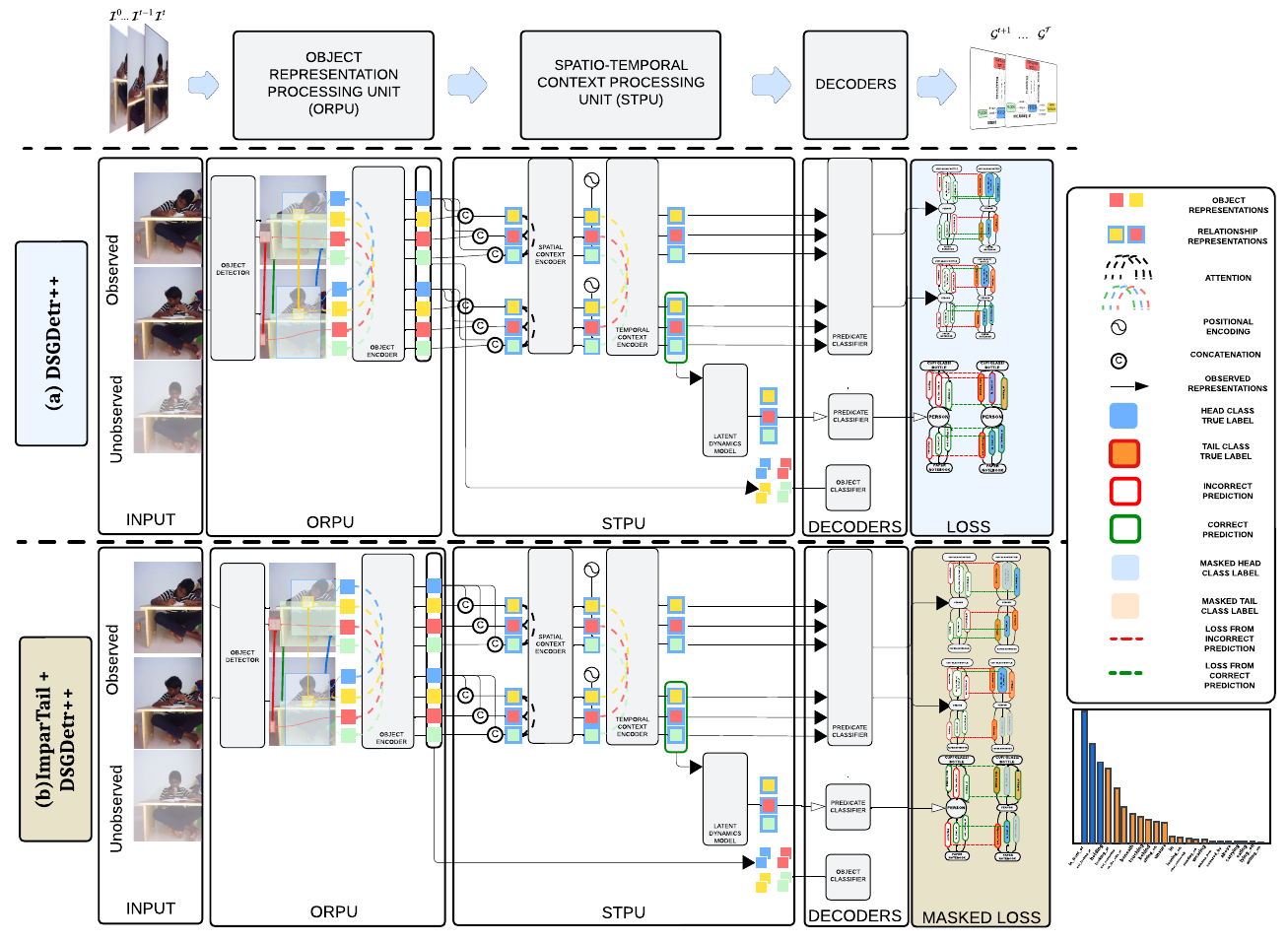}
    \caption{\underline{(a) \textbf{Architectural Components.}} In \textbf{DSGDetr++}, the Object Representations Processing Unit (ORPU) primarily consists of an object detector \textcolor{teal}{an object tracker and an object encoder. The visual features output by the object detector are used to construct tracklets corresponding to each object, and these representations are further enhanced by passing them through an object encoder}. Then, the Spatio-Temporal Context Processing Unit (STPU) takes these visual features as input and first constructs relationship representations utilizing the features of interacting objects; then, these relationship representations are fed to a spatial encoder and a transformer encoder. Thus, the spatio-temporal context-aware representations output by the temporal encoder are fed as input to another transformer encoder to anticipate the future relationship representations corresponding to interacting objects. Thus, relationship representations from the temporal encoder and future relationship representations from the anticipatory transformer encoder are input to two predicate classifiers for final predictions. \underline{(b) \textbf{Loss Function.}} The primary difference between \colorbox{\baselinecolor}{DSGDetr++} loss and the proposed \colorbox{\methodcolor}{\methodname + DSGDetr++} loss is illustrated using highlighting the employed losses. We do not mask any predicate label in \colorbox{\baselinecolor}{DSGDetr++} loss. In contrast, in the proposed \colorbox{\methodcolor}{\methodname + DSGDetr++} loss, we mask the losses corresponding to the \textit{head} classes output by predicate classification heads corresponding to both observed and anticipated relationship representations.}
    \label{fig:enter-label}
\end{figure*}

\paragraph{Loss Function \colorbox{\baselinecolor}{DSGDetr++}}

\begin{equation}
    \overbrace{
        \mathcal{L}_{i} = \sum_{t=1}^{\bar{T}} \mathcal{L}_{i}^{t},
        \quad
        \mathcal{L}_{i}^{t} = -\sum_{n=1}^{|\mathcal{C}|} y_{i,n}^{t} \log(\hat{\mathbf{c}}_{i,n}^{t})
    }^\text{Object Classification Loss (\RNum{1})}
\end{equation}

\begin{equation}
    \overbrace{
        \textcolor{red}{\mathcal{L}_{\text{gen}} = \sum_{t=1}^{\bar{T}} \mathcal{L}_{\text{gen}}^t, 
        \quad
        \mathcal{L}_{\text{gen}}^t = \sum_{ij} \mathcal{L}_{p^t_{ij}}}
    }^\text{Predicate Classification Loss (\RNum{2})}
\end{equation}

\begin{equation}
     \underbrace{
        \textcolor{red}{\mathcal{L}_{\text{ant}}^{(1:T)} = \sum_{t=T+1}^{\min(T+H, \bar{T})} \mathcal{L}_{\text{ant}}^t, 
        \quad
        \mathcal{L}_{\text{ant}}^t = \sum_{ij} \mathcal{L}_{p^t_{ij}}}
    }_\text{Predicate Classification Loss (\RNum{3})}
\end{equation}

\begin{equation}
    \underbrace{
        \mathcal{L}_{\text{recon}}^{(1:T)} = \sum_{t = T+1}^{\min(T+H, \bar{T})} \mathcal{L}_{\text{recon}}^{t},
        \quad
        \mathcal{L}_{\text{recon}}^t = \frac{1}{N(t) \times N(t)} \sum_{ij}^ {(N(t) \times N(t))}\text{L}_{\text{smooth}}(\mathbf{z}_{ij}^t - \hat{\mathbf{z}}_{ij}^t)
    }_\text{Reconstruction Loss (\RNum{4})}
\end{equation}

Thus, the total objective for training the proposed method can be written as:

\begin{equation}
    \mathcal{L} = 
    \underbrace{
        \sum_{t=1}^{\bar{T}} \left(\lambda_{1} \textcolor{red}{\mathcal{L}_{\text{gen}}^{t}} + 
        \lambda_{2} \sum_{i} \mathcal{L}_{i}^{t} \right)
    }_\text{Loss Over Observed Representations} 
    + 
    \underbrace{
        \sum_{T=3}^{\bar{T}-1} \left( 
            \lambda_{3} \textcolor{red}{\mathcal{L}_{\text{ant}}^{(1:T)}} +
            \lambda_{4} \mathcal{L}_{\text{recon}}^{(1:T)}
        \right)
    }_\text{Loss Over Anticipated Representations}
\end{equation}

\paragraph{Loss Function \colorbox{\methodcolor}{\methodname + DSGDetr++}}

\begin{equation}
    \overbrace{
        \mathcal{L}_{i} = \sum_{t=1}^{\bar{T}} \mathcal{L}_{i}^{t},
        \quad
        \mathcal{L}_{i}^{t} = -\sum_{n=1}^{|\mathcal{C}|} y_{i,n}^{t} \log(\hat{\mathbf{c}}_{i,n}^{t})
    }^\text{Object Classification Loss (\RNum{1})}
\end{equation}

\begin{equation}
    \overbrace{
        \textcolor{olive}{\mathscr{L}_{\text{gen}} = \sum_{t=1}^{\bar{T}} \mathscr{L}_{\text{gen}}^t, 
        \quad
        \mathscr{L}_{\text{gen}}^t = \sum_{ij} \mathbf{m}^t_{ij} * \mathcal{L}_{p^t_{ij}}}
    }^\text{Predicate Classification Loss (\RNum{2})}
\end{equation}

\begin{equation}
     \underbrace{
        \textcolor{olive}{\mathscr{L}_{\text{ant}}^{(1:T)} = \sum_{t=T+1}^{\min(T+H, \bar{T})} \mathscr{L}_{\text{ant}}^t, 
        \quad
        \mathscr{L}_{\text{ant}}^t = \sum_{ij} \mathbf{m}^t_{ij} * \mathcal{L}_{p^t_{ij}}}
    }_\text{Predicate Classification Loss (\RNum{3})}
\end{equation}

\begin{equation}
    \underbrace{
        \mathcal{L}_{\text{recon}}^{(1:T)} = \sum_{t = T+1}^{\min(T+H, \bar{T})} \mathcal{L}_{\text{recon}}^{t},
        \quad
        \mathcal{L}_{\text{recon}}^t = \frac{1}{N(t) \times N(t)} \sum_{ij}^ {(N(t) \times N(t))}\text{L}_{\text{smooth}}(\mathbf{z}_{ij}^t - \hat{\mathbf{z}}_{ij}^t)
    }_\text{Reconstruction Loss (\RNum{4})}
\end{equation}

Thus, the total objective for training the proposed method can be written as:

\begin{equation}
    \mathcal{L} = 
    \underbrace{
        \sum_{t=1}^{\bar{T}} \left(\lambda_{1} \textcolor{olive}{\mathscr{L}_{\text{gen}}^{t}} + 
        \lambda_{2} \sum_{i} \mathcal{L}_{i}^{t} \right)
    }_\text{Loss Over Observed Representations} 
    + 
    \underbrace{
        \sum_{T=3}^{\bar{T}-1} \left( 
            \lambda_{3} \textcolor{olive}{\mathscr{L}_{\text{ant}}^{(1:T)}} +
            \lambda_{4} \mathcal{L}_{\text{recon}}^{(1:T)}
        \right)
    }_\text{Loss Over Anticipated Representations}
\end{equation}

\paragraph{Implementation Details.}

\begin{itemize}
    \item[--] \textbf{Training Epochs.} We have capped the number of training epochs for both models, where one uses conventional loss, and the other uses the proposed \methodname framework to \textbf{5} epochs.  
    \item[--] \textbf{Loss Function.} Results reported in the literature for the method \colorbox{\baselinecolor}{DSGDetr++} were reproducible using the Multi-Label Margin Loss. We sometimes achieved higher numbers than those reported in the original paper. 
    \item[--] \textbf{Hyperparameters.} We use the same hyperparameter settings described in the paper\cite{peddi_et_al_scene_sayer_2024}. 
\end{itemize}

\subsubsection{\methodname + SceneSayer}

\begin{figure*}[!htbp]
    \centering
    \includegraphics[width=\textwidth]{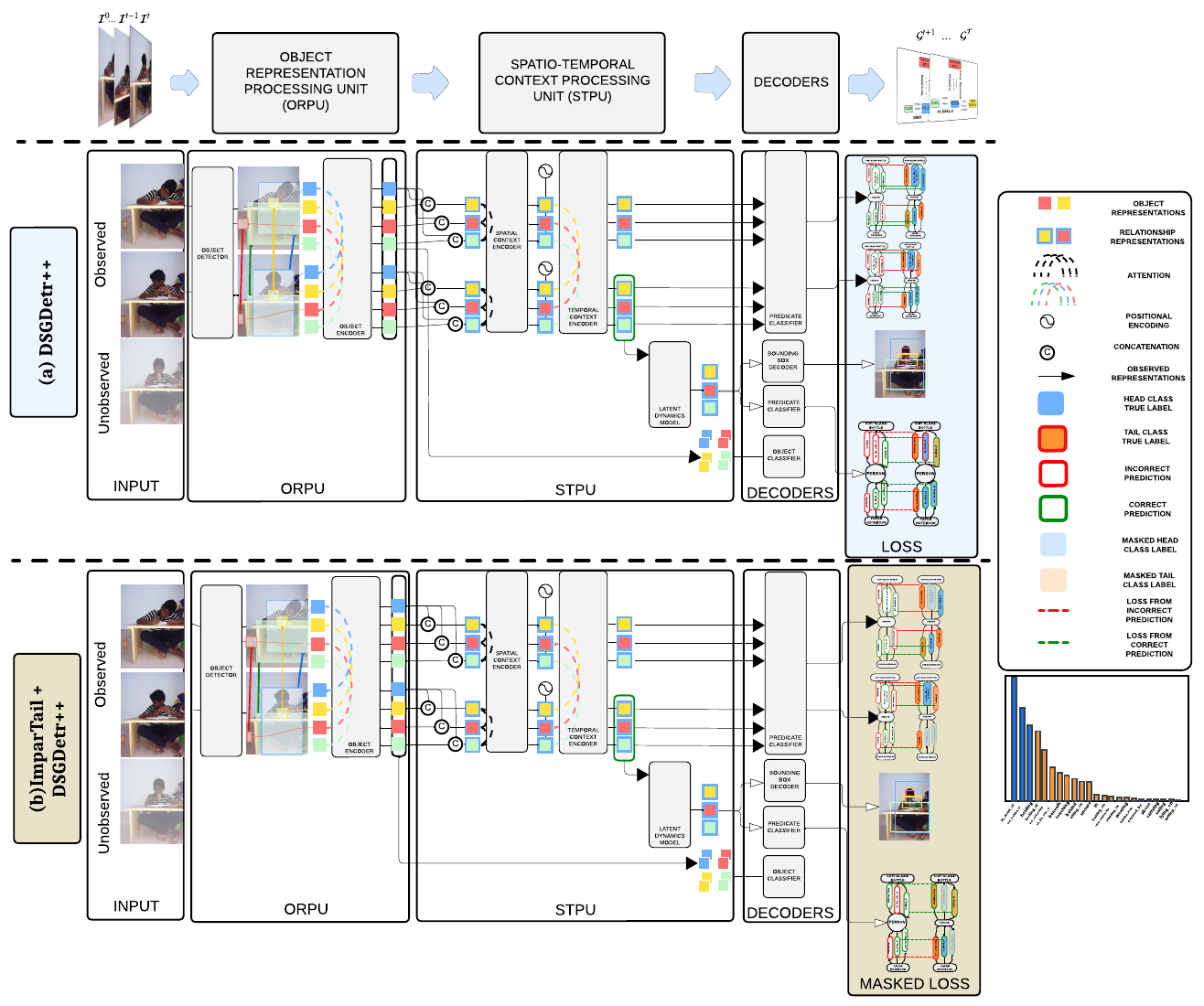}
    \caption{\underline{(a) \textbf{Architectural Components.}} In \textbf{SceneSayer}, the Object Representations Processing Unit (ORPU) primarily consists of an object detector \textcolor{teal}{an object tracker and an object encoder. The visual features output by the object detector are used to construct tracklets corresponding to each object, and these representations are further enhanced by passing them through an object encoder}. Then, the Spatio-Temporal Context Processing Unit (STPU) takes these visual features as input and first constructs relationship representations utilizing the features of interacting objects; then, these relationship representations are fed to a spatial encoder and a transformer encoder. Thus, the spatio-temporal context-aware representations output by the temporal encoder are \textcolor{teal}{used as initial values and an Ordinary Differential Equation/ Stochastic Differential Equation is solved to estimate the anticipated future relationship representations corresponding to the interacting objects}. Thus, relationship representations from the temporal encoder and future relationship representations from the anticipatory transformer encoder are input to two predicate classifiers for final predictions. \underline{(b) \textbf{Loss Function.}} The primary difference between \colorbox{\baselinecolor}{SceneSayer} loss and the proposed \colorbox{\methodcolor}{\methodname + SceneSayer} loss is illustrated using highlighting the employed losses. We do not mask any predicate label in \colorbox{\baselinecolor}{SceneSayer} loss. In contrast, in the proposed \colorbox{\methodcolor}{\methodname + SceneSayer} loss, we mask the losses corresponding to the \textit{head} classes output by predicate classification heads corresponding to both observed and anticipated relationship representations.}
    \label{fig:sup_scene_sayer}
\end{figure*}

\paragraph{Loss Function \colorbox{\baselinecolor}{SceneSayer}}

\begin{equation*} 
    \underbrace{
        \mathcal{L}_{i} = \sum_{t=1}^{\bar{T}} \mathcal{L}_{i}^{t},
        \quad
        \mathcal{L}_{i}^{t} = -\sum_{n=1}^{|\mathcal{C}|} y_{i,n}^{t} \log(\hat{\mathbf{c}}_{i,n}^{t})
    }_\text{Object Classification Loss (\RNum{1})}; 
    \quad
    \underbrace{
        \textcolor{red}{\mathcal{L}_{\text{gen}} = \sum_{t=1}^{\bar{T}} \mathcal{L}_{\text{gen}}^t, 
        \quad
        \mathcal{L}_{\text{gen}}^t = \sum_{ij} \mathcal{L}_{p^t_{ij}}}
    }_\text{Predicate Classification Loss (\RNum{2})}
\end{equation*}

\begin{equation*}
    \textcolor{red}{\mathcal{L}_{\text{ant}}^{(1:T)} = \sum_{t=T+1}^{\min(T+H, \bar{T})} \mathcal{L}_{\text{ant}}^t, 
    \quad
    \mathcal{L}_{\text{ant}}^t = \sum_{ij} \mathcal{L}_{p^t_{ij}}}
\end{equation*}

\begin{equation*} \label{eq:bounding_box_regression_loss}
    \mathcal{L}_{\text{boxes}}^{(1:T)} = \sum_{t = T+1}^{\min(T+H, \bar{T})} \mathcal{L}_{\text{boxes}}^{t},
    \quad \mathcal{L}_{\text{boxes}}^{t} = \sum_{k \in \text{boxes}} \text{L}_{\text{smooth}}(b_k^t - \hat{b}_k^t)
\end{equation*}

\begin{equation*}
    \mathcal{L}_{\text{recon}}^{(1:T)} = \sum_{t = T+1}^{\min(T+H, \bar{T})} \mathcal{L}_{\text{recon}}^{t},
    \quad
    \mathcal{L}_{\text{recon}}^t = \frac{1}{N(t) \times N(t)} \sum_{ij}^ {(N(t) \times N(t))}\text{L}_{\text{smooth}}(\mathbf{z}_{ij}^t - \hat{\mathbf{z}}_{ij}^t)
\end{equation*}

\begin{equation*}
    \mathcal{L} = 
    \underbrace{
        \sum_{t=1}^{\bar{T}} \left(\lambda_{1} \textcolor{red}{\mathcal{L}_{\text{gen}}^{t}} + 
        \lambda_{2} \sum_{i} \mathcal{L}_{i}^{t} \right)
    }_\text{Loss Over Observed Representations} 
    + 
    \underbrace{
        \sum_{T=3}^{\bar{T}-1} \left( 
            \lambda_{3} \textcolor{red}{\mathcal{L}_{\text{ant}}^{(1:T)}} +
            \lambda_{4} \mathcal{L}_{\text{boxes}}^{(1:T)} + 
            \lambda_{5} \mathcal{L}_{\text{recon}}^{(1:T)}
        \right)
    }_\text{Loss Over Anticipated Representations}
\end{equation*}

\paragraph{Loss Function \colorbox{\methodcolor}{\methodname + SceneSayer}}

\begin{equation*} 
    \underbrace{
        \mathcal{L}_{i} = \sum_{t=1}^{\bar{T}} \mathcal{L}_{i}^{t},
        \quad
        \mathcal{L}_{i}^{t} = -\sum_{n=1}^{|\mathcal{C}|} y_{i,n}^{t} \log(\hat{\mathbf{c}}_{i,n}^{t})
    }_\text{Object Classification Loss (\RNum{1})}; 
    \quad
    \underbrace{
        \textcolor{olive}{\mathscr{L}_{\text{gen}} = \sum_{t=1}^{\bar{T}} \mathscr{L}_{\text{gen}}^t, 
        \quad
        \mathscr{L}_{\text{gen}}^t = \sum_{ij} \mathbf{m}^t_{ij} * \mathcal{L}_{p^t_{ij}}}
    }_\text{Predicate Classification Loss (\RNum{2})}
\end{equation*}

\begin{equation*}
    \textcolor{olive}{\mathscr{L}_{\text{ant}}^{(1:T)} = \sum_{t=T+1}^{\min(T+H, \bar{T})} \mathscr{L}_{\text{ant}}^t, 
    \quad
    \mathscr{L}_{\text{ant}}^t = \sum_{ij} \mathbf{m}^t_{ij} * \mathcal{L}_{p^t_{ij}}}
\end{equation*}

\begin{equation*} \label{eq:bounding_box_regression_loss}
    \mathcal{L}_{\text{boxes}}^{(1:T)} = \sum_{t = T+1}^{\min(T+H, \bar{T})} \mathcal{L}_{\text{boxes}}^{t},
    \quad \mathcal{L}_{\text{boxes}}^{t} = \sum_{k \in \text{boxes}} \text{L}_{\text{smooth}}(b_k^t - \hat{b}_k^t)
\end{equation*}

\begin{equation*}
    \mathcal{L}_{\text{recon}}^{(1:T)} = \sum_{t = T+1}^{\min(T+H, \bar{T})} \mathcal{L}_{\text{recon}}^{t},
    \quad
    \mathcal{L}_{\text{recon}}^t = \frac{1}{N(t) \times N(t)} \sum_{ij}^ {(N(t) \times N(t))}\text{L}_{\text{smooth}}(\mathbf{z}_{ij}^t - \hat{\mathbf{z}}_{ij}^t)
\end{equation*}

\begin{equation*}
    \mathcal{L} = 
    \underbrace{
        \sum_{t=1}^{\bar{T}} \left(\lambda_{1} \textcolor{olive}{\mathscr{L}_{\text{gen}}^{t}} + 
        \lambda_{2} \sum_{i} \mathcal{L}_{i}^{t} \right)
    }_\text{Loss Over Observed Representations} 
    + 
    \underbrace{
        \sum_{T=3}^{\bar{T}-1} \left( 
            \lambda_{3} \textcolor{olive}{\mathscr{L}_{\text{ant}}^{(1:T)}} +
            \lambda_{4} \mathcal{L}_{\text{boxes}}^{(1:T)} + 
            \lambda_{5} \mathcal{L}_{\text{recon}}^{(1:T)}
        \right)
    }_\text{Loss Over Anticipated Representations}
\end{equation*}

\newpage
\section{Ablation-Overview} \label{sec:ablation}

\subsection{Video Scene Graph Generation}

\subsubsection{Modes}

We evaluate the trained models corresponding to baseline variants \colorbox{\baselinecolor}{STTran}, \colorbox{\baselinecolor}{DSGDetr} and the proposed method \\ \colorbox{\methodcolor}{\methodname + STTran}, \colorbox{\methodcolor}{\methodname + DSGDetr} using three standard modes described in the literature. \textbf{(1) Scene Graph Detection (SGDET), (2) Scene Graph Classification (SGCLS) and (3) Predicate Classification (PREDCLS)}.

\begin{itemize}
    \item[--]\textbf{Scene Graph Detection (SGDET):} In this mode, the model is input with frames corresponding to videos. It is tasked to detect objects and predict the relationship predicates between the detected objects.  
    \item[--]\textbf{Scene Graph Classification (SGCLS):} In this mode, the model is input with frames corresponding to videos along with bounding boxes of the objects. It is tasked to predict the relationship predicates between the objects.
    \item[--]\textbf{Predicate Classification (PREDCLS):} In this mode, the model is input with frames corresponding to videos along with bounding boxes of the objects and the object labels. It is tasked to predict the relationship predicates between the objects.
\end{itemize}

\subsubsection{Implementation Details.}

\begin{itemize}
    \item[--] \textbf{Min Threshold.} As \methodname proposes a curriculum-guided mask generation strategy, where the number of labels masked in each epoch increases monotonically. 
    \begin{enumerate}
        \item Thus, based on the maximum amount of masking applied, we train three variants of models - \{70, 40, 10\}. 
        \item These models correspond to the following scenarios: (1) \textbf{70:} Start from the complete data and reach a \{70\%, 40\%, 10\%\} masked settings in the last epochs, respectively.
    \end{enumerate}
    \item[--] In section~\ref{sec:ablation_results}, we provide findings corresponding to the proposed training scenarios. 
\end{itemize}

\subsection{Scene Graph Anticipation}

\subsubsection{Modes}

We evaluate the trained models corresponding to baseline variants \colorbox{\baselinecolor}{STTran++}, \colorbox{\baselinecolor}{DSGDetr++}, \\ \colorbox{\baselinecolor}{SceneSayerODE},  \colorbox{\baselinecolor}{SceneSayerSDE} and the proposed methods \colorbox{\methodcolor}{\methodname + STTran++}, \colorbox{\methodcolor}{\methodname + DSGDetr++}, \colorbox{\methodcolor}{\methodname + SceneSayerODE}, \\ \colorbox{\methodcolor}{\methodname + SceneSayerSDE}  using three standard modes described in the literature. \textbf{(1) Action Genome Scenes (AGS), (2) Partially Grounded Action Genome Scenes (PGAGS) and (3) Grounded Action Genome Scenes (GAGS)}.

\begin{itemize}
    \item[--] \textbf{Action Genome Scenes (AGS):} In this mode, the model receives only the video frames and is tasked to detect objects and infer future relationships between them. 
    \item[--] \textbf{Partially Grounded Action Genome Scenes (PGAGS):} In this mode, the model, along with frames, also receives the bounding boxes corresponding to the objects. It is tasked to use this information to infer relationships corresponding to future frames
    \item[--] \textbf{Grounded Action Genome Scenes (GAGS):} In this mode, the model, along with frames, also receives the bounding boxes corresponding to the objects and their labels. It is tasked to use this information to infer future relationships corresponding to interacting objects. 
\end{itemize}

\subsubsection{Implementation Details.}

\begin{itemize}
    \item[--] \textbf{Min Threshold.} As \methodname proposes a curriculum-guided mask generation strategy, where the number of labels masked in each epoch increases monotonically. 
    \begin{enumerate}
        \item Thus, based on the maximum amount of masking applied, we train three variants of models - \{70, 40, 10\}. 
        \item These models correspond to the following scenarios: (1) \textbf{70:} Start from the complete data and reach a \{70\%, 40\%, 10\%\} masked settings in the last epochs, respectively.
    \end{enumerate}
    \item[--] In section~\ref{sec:ablation_results}, we provide findings corresponding to the proposed training scenarios. 
\end{itemize}

\subsection{Robustness Evaluation.}

In section~\ref{sec:ablation_results}, we evaluate the robustness of trained models corresponding to input corruptions and present the results for each mode described above.

\newpage
\section{Ablation Results} \label{sec:ablation_results}

\subsection{Video Scene Graph Generation}

\subsubsection{Findings}

\begin{enumerate}
    \item \textbf{Table~\ref{tab:sup_sgg_no_constraint}} provides a comparative analysis under \textbf{NO CONSTRAINT} graph building strategy for different modes and methods for VidSGG, presenting results under various recall metrics (R@10, R@20, R@50, R@100) and mean recall metrics (mR@10, mR@20, mR@50, mR@100). 
        \begin{enumerate}
            \item We observe that employing the proposed method, the mean recall metrics improved across all modes with only a marginal decrease in recall scores; for example, in the SGDET mode with the STTran method, R@10 slightly decreases from 20.30 to 20.20, conversely, mR@10 increases from 19.30 to 23.50. 
            \item We also observe that the mean recall scores follow a monotonic trend as we increase the masking ratio (avoiding more head classes). We note that the reduction in the recall values is very low. 
        \end{enumerate}
    \item \textbf{Table~\ref{tab:sup_sgg_with_constraint}, Table~\ref{tab:sup_sgg_semi_constraint}} provides a comparative analysis under \textbf{WITH/SEMI CONSTRAINT} graph building strategy
        \begin{enumerate}
            \item We observe that the proposed method improves the mean recall significantly across most setups, though its effect on standard recall metrics is mixed, slightly resulting in a decrease.
        \end{enumerate} 
    \item  In {Table~\ref{tab:sup_sgg_with_constraint}} for STTran under the SGCLS mode, while our method slightly reduced standard recall metrics @100 by 11.6\%, there is a notable improvement in mean recall, mR@100 by 36.2\%. PREDCLS mode shows less variability in recall changes using our method but substantially increases mean recalls @50 for STTran, jumping from 34.80 to 52.90.  While \textbf{\methodname(Ours)} augmentation in constrained and semi-constrained results in decreases in R@50 by 15\%, it boosts mR@50 by over 25\%.         
\end{enumerate}

\subsubsection{Results} 
\begin{table*}[!h]
    \centering
    \captionsetup{font=small}
    \caption{No Constraint Results for VidSGG.}
    \label{tab:sup_sgg_no_constraint}
    \renewcommand{\arraystretch}{1.2} 
    \resizebox{\textwidth}{!}{
    \begin{tabular}{l|l|l|cccc|cccc}
    \hline
        \multirow{2}{*}{Mode} & \multirow{2}{*}{Method} & \multirow{2}{*}{$\mathcal{S}$} &  \multicolumn{8}{c}{\textbf{NO CONSTRAINT}} \\ 
        \cmidrule(lr){4-7} \cmidrule(lr){8-11} 
         & & & \textbf{R@10} & \textbf{R@20} & \textbf{R@50} & \textbf{R@100} &\textbf{mR@10} & \textbf{mR@20} & \textbf{mR@50}  & \textbf{mR@100} \\ \hline
   \multirow{8}{*}{\textbf{SGCLS}} & STTran~\cite{cong_et_al_sttran_2021}& -  & \cellcolor{highlightColor}\textbf{51.60}  & \cellcolor{highlightColor}\textbf{62.80}  & 66.30  & 66.60  & 38.80  & 47.10  & 59.90  & 66.70  \\ 
  &  \quad+\textbf{\methodname(Ours)}& 70  & 49.80  (-3.49\%) & 62.40  (-0.64\%) & \cellcolor{highlightColor}\textbf{66.40}  (+0.15\%) & \cellcolor{highlightColor}\textbf{66.70}  (+0.15\%) & 43.10  (+11.08\%) & 53.10  (+12.74\%) & 61.00  (+1.84\%) & 65.00  (-2.55\%) \\ 
  &  \quad+\textbf{\methodname(Ours)}& 40  & 49.90  (-3.29\%) & 62.10  (-1.11\%) & 66.40  (+0.15\%) & 66.70  (+0.15\%) & 45.10  (+16.24\%) & 55.10  (+16.99\%) & 64.10  (+7.01\%) & 66.60  (-0.15\%) \\ 
  &  \quad+\textbf{\methodname(Ours)}& 10  & 48.50  (-6.01\%) & 61.30  (-2.39\%) & 66.20  (-0.15\%) & 66.50  (-0.15\%) & \cellcolor{highlightColor}\textbf{47.40}  (+22.16\%) & \cellcolor{highlightColor}\textbf{57.50}  (+22.08\%) & \cellcolor{highlightColor}\textbf{66.60}  (+11.19\%) & \cellcolor{highlightColor}\textbf{68.10}  (+2.10\%) \\ 
    \cmidrule(lr){2-11}
  &  DSGDetr~\cite{Feng_2021}& -  & \cellcolor{highlightColor}\textbf{55.50}  & \cellcolor{highlightColor}\textbf{68.00}  & \cellcolor{highlightColor}\textbf{72.40}  & \cellcolor{highlightColor}\textbf{72.80}  & 39.90  & 49.40  & 64.60  & \cellcolor{highlightColor}\textbf{72.70}  \\ 
  &  \quad+\textbf{\methodname(Ours)}& 70  & 53.30  (-3.96\%) & 67.30  (-1.03\%) & 72.10  (-0.41\%) & 72.60  (-0.27\%) & 42.90  (+7.52\%) & 53.10  (+7.49\%) & 66.10  (+2.32\%) & 72.20  (-0.69\%) \\ 
  &  \quad+\textbf{\methodname(Ours)}& 40  & 50.90  (-8.29\%) & 66.10  (-2.79\%) & 72.10  (-0.41\%) & 72.60  (-0.27\%) & 45.00  (+12.78\%) & 55.50  (+12.35\%) & 67.20  (+4.02\%) & 71.40  (-1.79\%) \\ 
  &  \quad+\textbf{\methodname(Ours)}& 10  & 50.20  (-9.55\%) & 65.30  (-3.97\%) & 71.90  (-0.69\%) & 72.70  (-0.14\%) & \cellcolor{highlightColor}\textbf{48.80}  (+22.31\%) & \cellcolor{highlightColor}\textbf{59.60}  (+20.65\%) & \cellcolor{highlightColor}\textbf{70.10}  (+8.51\%) & 72.20  (-0.69\%) \\ 
    \cmidrule(lr){2-11}
    \cmidrule(lr){1-11}
   \multirow{8}{*}{\textbf{SGDET}} & STTran~\cite{cong_et_al_sttran_2021}& -  & \cellcolor{highlightColor}\textbf{20.30}  & \cellcolor{highlightColor}\textbf{31.10}  & \cellcolor{highlightColor}\textbf{45.90}  & 48.40  & 19.30  & 26.90  & 35.60  & 39.70  \\ 
  &  \quad+\textbf{\methodname(Ours)}& 70  & 19.80  (-2.46\%) & 30.20  (-2.89\%) & 45.80  (-0.22\%) & \cellcolor{highlightColor}\textbf{48.60}  (+0.41\%) & 20.80  (+7.77\%) & 29.50  (+9.67\%) & 38.70  (+8.71\%) & 42.00  (+5.79\%) \\ 
  &  \quad+\textbf{\methodname(Ours)}& 40  & 20.20  (-0.49\%) & 30.80  (-0.96\%) & 45.30  (-1.31\%) & 48.20  (-0.41\%) & 22.60  (+17.10\%) & 31.10  (+15.61\%) & 39.10  (+9.83\%) & 42.10  (+6.05\%) \\ 
  &  \quad+\textbf{\methodname(Ours)}& 10  & 20.00  (-1.48\%) & 30.10  (-3.22\%) & 45.10  (-1.74\%) & 48.50  (+0.21\%) & \cellcolor{highlightColor}\textbf{23.50}  (+21.76\%) & \cellcolor{highlightColor}\textbf{33.60}  (+24.91\%) & \cellcolor{highlightColor}\textbf{43.80}  (+23.03\%) & \cellcolor{highlightColor}\textbf{47.00}  (+18.39\%) \\ 
    \cmidrule(lr){2-11}
  &  DSGDetr~\cite{Feng_2021}& -  & \cellcolor{highlightColor}\textbf{29.80}  & \cellcolor{highlightColor}\textbf{39.00}  & \cellcolor{highlightColor}\textbf{46.40}  & 48.30  & 23.30  & 29.80  & 36.00  & 39.70  \\ 
  &  \quad+\textbf{\methodname(Ours)}& 70  & 28.60  (-4.03\%) & 38.00  (-2.56\%) & 46.40  & \cellcolor{highlightColor}\textbf{49.00}  (+1.45\%) & 25.00  (+7.30\%) & 32.00  (+7.38\%) & 39.30  (+9.17\%) & 42.50  (+7.05\%) \\ 
  &  \quad+\textbf{\methodname(Ours)}& 40  & 25.90  (-13.09\%) & 36.10  (-7.44\%) & 45.60  (-1.72\%) & 48.70  (+0.83\%) & 24.70  (+6.01\%) & 32.20  (+8.05\%) & 39.60  (+10.00\%) & 43.20  (+8.82\%) \\ 
  &  \quad+\textbf{\methodname(Ours)}& 10  & 26.50  (-11.07\%) & 36.10  (-7.44\%) & 45.20  (-2.59\%) & 48.40  (+0.21\%) & \cellcolor{highlightColor}\textbf{27.50}  (+18.03\%) & \cellcolor{highlightColor}\textbf{35.20}  (+18.12\%) & \cellcolor{highlightColor}\textbf{43.30}  (+20.28\%) & \cellcolor{highlightColor}\textbf{46.60}  (+17.38\%) \\ 
    \cmidrule(lr){2-11}
    \cmidrule(lr){1-11}
   \multirow{8}{*}{\textbf{PREDCLS}} & STTran~\cite{cong_et_al_sttran_2021}& -  & \cellcolor{highlightColor}\textbf{73.20}  & \cellcolor{highlightColor}\textbf{92.70}  & 99.20  & \cellcolor{highlightColor}\textbf{99.90}  & 45.70  & 63.40  & 80.50  & 95.60  \\ 
  &  \quad+\textbf{\methodname(Ours)}& 70  & 70.20  (-4.10\%) & 91.40  (-1.40\%) & \cellcolor{highlightColor}\textbf{99.30}  (+0.10\%) & 99.90  & 55.00  (+20.35\%) & 71.80  (+13.25\%) & 86.70  (+7.70\%) & 97.00  (+1.46\%) \\ 
  &  \quad+\textbf{\methodname(Ours)}& 40  & 67.50  (-7.79\%) & 89.70  (-3.24\%) & 99.20  & 99.90  & 54.80  (+19.91\%) & 72.10  (+13.72\%) & 86.70  (+7.70\%) & 97.20  (+1.67\%) \\ 
  &  \quad+\textbf{\methodname(Ours)}& 10  & 67.50  (-7.79\%) & 88.80  (-4.21\%) & 99.00  (-0.20\%) & 99.90  & \cellcolor{highlightColor}\textbf{65.50}  (+43.33\%) & \cellcolor{highlightColor}\textbf{82.00}  (+29.34\%) & \cellcolor{highlightColor}\textbf{93.00}  (+15.53\%) & \cellcolor{highlightColor}\textbf{99.60}  (+4.18\%) \\ 
    \cmidrule(lr){2-11}
  &  DSGDetr~\cite{Feng_2021}& -  & \cellcolor{highlightColor}\textbf{72.80}  & \cellcolor{highlightColor}\textbf{92.40}  & \cellcolor{highlightColor}\textbf{99.20}  & \cellcolor{highlightColor}\textbf{99.90}  & 45.60  & 64.40  & 80.50  & 94.70  \\ 
  &  \quad+\textbf{\methodname(Ours)}& 70  & 67.70  (-7.01\%) & 89.60  (-3.03\%) & 99.20  & 99.90  & 56.00  (+22.81\%) & 72.60  (+12.73\%) & 85.90  (+6.71\%) & 97.30  (+2.75\%) \\ 
  &  \quad+\textbf{\methodname(Ours)}& 40  & 68.00  (-6.59\%) & 90.10  (-2.49\%) & 99.20  & 99.90  & 54.50  (+19.52\%) & 71.80  (+11.49\%) & 86.40  (+7.33\%) & 97.30  (+2.75\%) \\ 
  &  \quad+\textbf{\methodname(Ours)}& 10  & 65.80  (-9.62\%) & 87.70  (-5.09\%) & 98.90  (-0.30\%) & 99.90  & \cellcolor{highlightColor}\textbf{59.40}  (+30.26\%) & \cellcolor{highlightColor}\textbf{76.20}  (+18.32\%) & \cellcolor{highlightColor}\textbf{89.80}  (+11.55\%) & \cellcolor{highlightColor}\textbf{98.10}  (+3.59\%) \\ 
    \cmidrule(lr){2-11}
    \cmidrule(lr){1-11}
    \hline
    \end{tabular}
    }
\end{table*}

\begin{table*}[!htbp]
    \centering
    \captionsetup{font=small}
    \caption{With Constraint Results for VidSGG.}
    \label{tab:sup_sgg_with_constraint}
    \renewcommand{\arraystretch}{1.2} 
    \resizebox{\textwidth}{!}{
    \begin{tabular}{l|l|l|cccc|cccc}
    \hline
        \multirow{2}{*}{Mode} & \multirow{2}{*}{Method} & \multirow{2}{*}{$\mathcal{S}$} &  \multicolumn{8}{c}{\textbf{WITH CONSTRAINT}} \\ 
        \cmidrule(lr){4-7} \cmidrule(lr){8-11} 
         & & & \textbf{R@10} & \textbf{R@20} & \textbf{R@50} & \textbf{R@100} &\textbf{mR@10} & \textbf{mR@20} & \textbf{mR@50}  & \textbf{mR@100} \\ \hline
   \multirow{8}{*}{\textbf{SGCLS}} & STTran~\cite{cong_et_al_sttran_2021}& -  & \cellcolor{highlightColor}\textbf{44.90}  & \cellcolor{highlightColor}\textbf{46.50}  & \cellcolor{highlightColor}\textbf{46.50}  & \cellcolor{highlightColor}\textbf{46.50}  & 25.00  & 27.50  & 27.60  & 27.60  \\ 
  &  \quad+\textbf{\methodname(Ours)}& 70  & 42.00  (-6.46\%) & 43.30  (-6.88\%) & 43.30  (-6.88\%) & 43.30  (-6.88\%) & 25.90  (+3.60\%) & 28.70  (+4.36\%) & 28.80  (+4.35\%) & 28.80  (+4.35\%) \\ 
  &  \quad+\textbf{\methodname(Ours)}& 40  & 42.40  (-5.57\%) & 43.70  (-6.02\%) & 43.80  (-5.81\%) & 43.80  (-5.81\%) & 27.80  (+11.20\%) & 30.60  (+11.27\%) & 30.70  (+11.23\%) & 30.70  (+11.23\%) \\ 
  &  \quad+\textbf{\methodname(Ours)}& 10  & 39.90  (-11.14\%) & 41.10  (-11.61\%) & 41.10  (-11.61\%) & 41.10  (-11.61\%) & \cellcolor{highlightColor}\textbf{32.30}  (+29.20\%) & \cellcolor{highlightColor}\textbf{36.20}  (+31.64\%) & \cellcolor{highlightColor}\textbf{36.20}  (+31.16\%) & \cellcolor{highlightColor}\textbf{36.20}  (+31.16\%) \\ 
    \cmidrule(lr){2-11}
  &  DSGDetr~\cite{Feng_2021}& -  & \cellcolor{highlightColor}\textbf{47.80}  & \cellcolor{highlightColor}\textbf{49.30}  & \cellcolor{highlightColor}\textbf{49.40}  & \cellcolor{highlightColor}\textbf{49.40}  & 25.60  & 28.10  & 28.10  & 28.10  \\ 
  &  \quad+\textbf{\methodname(Ours)}& 70  & 46.00  (-3.77\%) & 47.40  (-3.85\%) & 47.40  (-4.05\%) & 47.40  (-4.05\%) & 27.10  (+5.86\%) & 30.20  (+7.47\%) & 30.30  (+7.83\%) & 30.30  (+7.83\%) \\ 
  &  \quad+\textbf{\methodname(Ours)}& 40  & 41.20  (-13.81\%) & 42.40  (-14.00\%) & 42.40  (-14.17\%) & 42.40  (-14.17\%) & 27.90  (+8.98\%) & 30.80  (+9.61\%) & 30.80  (+9.61\%) & 30.80  (+9.61\%) \\ 
  &  \quad+\textbf{\methodname(Ours)}& 10  & 40.50  (-15.27\%) & 42.00  (-14.81\%) & 42.00  (-14.98\%) & 42.00  (-14.98\%) & \cellcolor{highlightColor}\textbf{32.20}  (+25.78\%) & \cellcolor{highlightColor}\textbf{36.00}  (+28.11\%) & \cellcolor{highlightColor}\textbf{36.00}  (+28.11\%) & \cellcolor{highlightColor}\textbf{36.00}  (+28.11\%) \\ 
    \cmidrule(lr){2-11}
    \cmidrule(lr){1-11}
   \multirow{8}{*}{\textbf{SGDET}} & STTran~\cite{cong_et_al_sttran_2021}& -  & \cellcolor{highlightColor}\textbf{19.00}  & \cellcolor{highlightColor}\textbf{29.40}  & \cellcolor{highlightColor}\textbf{32.10}  & \cellcolor{highlightColor}\textbf{32.10}  & 8.00  & 16.60  & 19.30  & 19.30  \\ 
  &  \quad+\textbf{\methodname(Ours)}& 70  & 17.90  (-5.79\%) & 27.80  (-5.44\%) & 30.60  (-4.67\%) & 30.60  (-4.67\%) & 8.20  (+2.50\%) & 17.50  (+5.42\%) & 20.60  (+6.74\%) & 20.60  (+6.74\%) \\ 
  &  \quad+\textbf{\methodname(Ours)}& 40  & 17.50  (-7.89\%) & 27.50  (-6.46\%) & 30.30  (-5.61\%) & 30.30  (-5.61\%) & 8.80  (+10.00\%) & 19.20  (+15.66\%) & 22.60  (+17.10\%) & 22.60  (+17.10\%) \\ 
  &  \quad+\textbf{\methodname(Ours)}& 10  & 16.00  (-15.79\%) & 25.60  (-12.93\%) & 28.40  (-11.53\%) & 28.40  (-11.53\%) & \cellcolor{highlightColor}\textbf{9.40}  (+17.50\%) & \cellcolor{highlightColor}\textbf{21.50}  (+29.52\%) & \cellcolor{highlightColor}\textbf{25.90}  (+34.20\%) & \cellcolor{highlightColor}\textbf{25.90}  (+34.20\%) \\ 
    \cmidrule(lr){2-11}
  &  DSGDetr~\cite{Feng_2021}& -  & \cellcolor{highlightColor}\textbf{17.10}  & \cellcolor{highlightColor}\textbf{28.80}  & \cellcolor{highlightColor}\textbf{33.90}  & \cellcolor{highlightColor}\textbf{33.90}  & 6.70  & 14.70  & 19.10  & 19.10  \\ 
  &  \quad+\textbf{\methodname(Ours)}& 70  & 16.30  (-4.68\%) & 27.50  (-4.51\%) & 32.50  (-4.13\%) & 32.60  (-3.83\%) & 7.40  (+10.45\%) & 17.60  (+19.73\%) & 23.20  (+21.47\%) & 23.20  (+21.47\%) \\ 
  &  \quad+\textbf{\methodname(Ours)}& 40  & 14.10  (-17.54\%) & 23.40  (-18.75\%) & 27.40  (-19.17\%) & 27.50  (-18.88\%) & 7.30  (+8.96\%) & 16.80  (+14.29\%) & 22.40  (+17.28\%) & 22.40  (+17.28\%) \\ 
  &  \quad+\textbf{\methodname(Ours)}& 10  & 15.40  (-9.94\%) & 25.70  (-10.76\%) & 30.10  (-11.21\%) & 30.10  (-11.21\%) & \cellcolor{highlightColor}\textbf{7.50}  (+11.94\%) & \cellcolor{highlightColor}\textbf{17.80}  (+21.09\%) & \cellcolor{highlightColor}\textbf{23.70}  (+24.08\%) & \cellcolor{highlightColor}\textbf{23.80}  (+24.61\%) \\ 
    \cmidrule(lr){2-11}
    \cmidrule(lr){1-11}
   \multirow{8}{*}{\textbf{PREDCLS}} & STTran~\cite{cong_et_al_sttran_2021}& -  & \cellcolor{highlightColor}\textbf{66.40}  & \cellcolor{highlightColor}\textbf{69.90}  & \cellcolor{highlightColor}\textbf{69.90}  & \cellcolor{highlightColor}\textbf{69.90}  & 30.50  & 34.70  & 34.80  & 34.80  \\ 
  &  \quad+\textbf{\methodname(Ours)}& 70  & 61.50  (-7.38\%) & 64.80  (-7.30\%) & 64.80  (-7.30\%) & 64.80  (-7.30\%) & 34.30  (+12.46\%) & 39.70  (+14.41\%) & 39.80  (+14.37\%) & 39.80  (+14.37\%) \\ 
  &  \quad+\textbf{\methodname(Ours)}& 40  & 57.20  (-13.86\%) & 60.20  (-13.88\%) & 60.30  (-13.73\%) & 60.30  (-13.73\%) & 37.40  (+22.62\%) & 43.60  (+25.65\%) & 43.80  (+25.86\%) & 43.80  (+25.86\%) \\ 
  &  \quad+\textbf{\methodname(Ours)}& 10  & 57.70  (-13.10\%) & 60.80  (-13.02\%) & 60.80  (-13.02\%) & 60.80  (-13.02\%) & \cellcolor{highlightColor}\textbf{44.00}  (+44.26\%) & \cellcolor{highlightColor}\textbf{52.70}  (+51.87\%) & \cellcolor{highlightColor}\textbf{52.90}  (+52.01\%) & \cellcolor{highlightColor}\textbf{52.90}  (+52.01\%) \\ 
    \cmidrule(lr){2-11}
  &  DSGDetr~\cite{Feng_2021}& -  & \cellcolor{highlightColor}\textbf{66.50}  & \cellcolor{highlightColor}\textbf{70.00}  & \cellcolor{highlightColor}\textbf{70.00}  & \cellcolor{highlightColor}\textbf{70.00}  & 31.50  & 36.10  & 36.20  & 36.20  \\ 
  &  \quad+\textbf{\methodname(Ours)}& 70  & 58.30  (-12.33\%) & 61.50  (-12.14\%) & 61.50  (-12.14\%) & 61.50  (-12.14\%) & 38.20  (+21.27\%) & 45.00  (+24.65\%) & 45.10  (+24.59\%) & 45.10  (+24.59\%) \\ 
  &  \quad+\textbf{\methodname(Ours)}& 40  & 58.00  (-12.78\%) & 61.10  (-12.71\%) & 61.10  (-12.71\%) & 61.10  (-12.71\%) & 37.30  (+18.41\%) & 43.40  (+20.22\%) & 43.50  (+20.17\%) & 43.50  (+20.17\%) \\ 
  &  \quad+\textbf{\methodname(Ours)}& 10  & 55.50  (-16.54\%) & 58.30  (-16.71\%) & 58.30  (-16.71\%) & 58.30  (-16.71\%) & \cellcolor{highlightColor}\textbf{41.00}  (+30.16\%) & \cellcolor{highlightColor}\textbf{48.10}  (+33.24\%) & \cellcolor{highlightColor}\textbf{48.20}  (+33.15\%) & \cellcolor{highlightColor}\textbf{48.20}  (+33.15\%) \\ 
    \cmidrule(lr){2-11}
    \cmidrule(lr){1-11}
    \hline
    \end{tabular}
    }
\end{table*}

\begin{table*}[!h]
    \centering
    \captionsetup{font=small}
    \caption{Semi Constraint Results for VidSGG.}
    \label{tab:sup_sgg_semi_constraint}
    \renewcommand{\arraystretch}{1.2} 
    \resizebox{\textwidth}{!}{
    \begin{tabular}{l|l|l|cccc|cccc}
    \hline
        \multirow{2}{*}{Mode} & \multirow{2}{*}{Method} & \multirow{2}{*}{$\mathcal{S}$} &  \multicolumn{8}{c}{\textbf{SEMI CONSTRAINT}} \\ 
        \cmidrule(lr){4-7} \cmidrule(lr){8-11} 
         & & & \textbf{R@10} & \textbf{R@20} & \textbf{R@50} & \textbf{R@100} &\textbf{mR@10} & \textbf{mR@20} & \textbf{mR@50}  & \textbf{mR@100} \\ \hline
   \multirow{8}{*}{\textbf{SGCLS}} & STTran~\cite{cong_et_al_sttran_2021}& -  & \cellcolor{highlightColor}\textbf{49.90}  & \cellcolor{highlightColor}\textbf{55.80}  & \cellcolor{highlightColor}\textbf{56.20}  & \cellcolor{highlightColor}\textbf{56.20}  & 29.50  & 39.90  & 40.90  & 40.90  \\ 
  &  \quad+\textbf{\methodname(Ours)}& 70  & 49.00  (-1.80\%) & 55.60  (-0.36\%) & 56.20  & 56.20  & 32.50  (+10.17\%) & 45.80  (+14.79\%) & 47.60  (+16.38\%) & 47.60  (+16.38\%) \\ 
  &  \quad+\textbf{\methodname(Ours)}& 40  & 48.60  (-2.61\%) & 54.80  (-1.79\%) & 55.20  (-1.78\%) & 55.20  (-1.78\%) & 34.30  (+16.27\%) & 48.40  (+21.30\%) & 50.00  (+22.25\%) & 50.00  (+22.25\%) \\ 
  &  \quad+\textbf{\methodname(Ours)}& 10  & 46.40  (-7.01\%) & 52.40  (-6.09\%) & 52.80  (-6.05\%) & 52.80  (-6.05\%) & \cellcolor{highlightColor}\textbf{36.20}  (+22.71\%) & \cellcolor{highlightColor}\textbf{50.50}  (+26.57\%) & \cellcolor{highlightColor}\textbf{52.20}  (+27.63\%) & \cellcolor{highlightColor}\textbf{52.20}  (+27.63\%) \\ 
    \cmidrule(lr){2-11}
  &  DSGDetr~\cite{Feng_2021}& -  & \cellcolor{highlightColor}\textbf{53.90}  & \cellcolor{highlightColor}\textbf{60.40}  & \cellcolor{highlightColor}\textbf{60.70}  & \cellcolor{highlightColor}\textbf{60.70}  & 30.10  & 40.60  & 41.60  & 41.60  \\ 
  &  \quad+\textbf{\methodname(Ours)}& 70  & 52.30  (-2.97\%) & 59.60  (-1.32\%) & 60.30  (-0.66\%) & 60.30  (-0.66\%) & 32.50  (+7.97\%) & 45.20  (+11.33\%) & 47.20  (+13.46\%) & 47.20  (+13.46\%) \\ 
  &  \quad+\textbf{\methodname(Ours)}& 40  & 50.50  (-6.31\%) & 58.50  (-3.15\%) & 59.50  (-1.98\%) & 59.50  (-1.98\%) & 33.90  (+12.62\%) & 49.00  (+20.69\%) & 51.70  (+24.28\%) & 51.70  (+24.28\%) \\ 
  &  \quad+\textbf{\methodname(Ours)}& 10  & 46.80  (-13.17\%) & 53.80  (-10.93\%) & 54.40  (-10.38\%) & 54.40  (-10.38\%) & \cellcolor{highlightColor}\textbf{36.80}  (+22.26\%) & \cellcolor{highlightColor}\textbf{52.40}  (+29.06\%) & \cellcolor{highlightColor}\textbf{54.90}  (+31.97\%) & \cellcolor{highlightColor}\textbf{54.90}  (+31.97\%) \\ 
    \cmidrule(lr){2-11}
    \cmidrule(lr){1-11}
   \multirow{8}{*}{\textbf{SGDET}} & STTran~\cite{cong_et_al_sttran_2021}& -  & \cellcolor{highlightColor}\textbf{18.60}  & \cellcolor{highlightColor}\textbf{31.00}  & \cellcolor{highlightColor}\textbf{41.20}  & 41.50  & 7.70  & 18.20  & 30.40  & 30.80  \\ 
  &  \quad+\textbf{\methodname(Ours)}& 70  & 18.00  (-3.23\%) & 30.10  (-2.90\%) & 41.10  (-0.24\%) & \cellcolor{highlightColor}\textbf{41.70}  (+0.48\%) & 7.90  (+2.60\%) & 19.00  (+4.40\%) & 33.90  (+11.51\%) & 34.60  (+12.34\%) \\ 
  &  \quad+\textbf{\methodname(Ours)}& 40  & 17.70  (-4.84\%) & 29.70  (-4.19\%) & 39.10  (-5.10\%) & 39.40  (-5.06\%) & \cellcolor{highlightColor}\textbf{8.70}  (+12.99\%) & 21.30  (+17.03\%) & 34.80  (+14.47\%) & 35.10  (+13.96\%) \\ 
  &  \quad+\textbf{\methodname(Ours)}& 10  & 16.40  (-11.83\%) & 28.20  (-9.03\%) & 37.90  (-8.01\%) & 38.20  (-7.95\%) & 8.60  (+11.69\%) & \cellcolor{highlightColor}\textbf{21.80}  (+19.78\%) & \cellcolor{highlightColor}\textbf{38.30}  (+25.99\%) & \cellcolor{highlightColor}\textbf{38.80}  (+25.97\%) \\ 
    \cmidrule(lr){2-11}
  &  DSGDetr~\cite{Feng_2021}& -  & \cellcolor{highlightColor}\textbf{16.40}  & \cellcolor{highlightColor}\textbf{28.70}  & \cellcolor{highlightColor}\textbf{40.70}  & \cellcolor{highlightColor}\textbf{41.50}  & 6.50  & 16.00  & 30.40  & 31.50  \\ 
  &  \quad+\textbf{\methodname(Ours)}& 70  & 15.80  (-3.66\%) & 27.90  (-2.79\%) & 40.20  (-1.23\%) & 41.20  (-0.72\%) & 6.90  (+6.15\%) & 17.30  (+8.12\%) & 34.20  (+12.50\%) & 35.60  (+13.02\%) \\ 
  &  \quad+\textbf{\methodname(Ours)}& 40  & 14.10  (-14.02\%) & 25.20  (-12.20\%) & 37.30  (-8.35\%) & 38.70  (-6.75\%) & 6.90  (+6.15\%) & 16.70  (+4.38\%) & 33.40  (+9.87\%) & 35.20  (+11.75\%) \\ 
  &  \quad+\textbf{\methodname(Ours)}& 10  & 15.20  (-7.32\%) & 26.80  (-6.62\%) & 37.90  (-6.88\%) & 39.00  (-6.02\%) & \cellcolor{highlightColor}\textbf{7.30}  (+12.31\%) & \cellcolor{highlightColor}\textbf{18.40}  (+15.00\%) & \cellcolor{highlightColor}\textbf{36.60}  (+20.39\%) & \cellcolor{highlightColor}\textbf{38.40}  (+21.90\%) \\ 
    \cmidrule(lr){2-11}
    \cmidrule(lr){1-11}
   \multirow{8}{*}{\textbf{PREDCLS}} & STTran~\cite{cong_et_al_sttran_2021}& -  & \cellcolor{highlightColor}\textbf{71.80}  & \cellcolor{highlightColor}\textbf{82.50}  & \cellcolor{highlightColor}\textbf{83.30}  & \cellcolor{highlightColor}\textbf{83.30}  & 36.60  & 51.80  & 53.80  & 53.80  \\ 
  &  \quad+\textbf{\methodname(Ours)}& 70  & 69.60  (-3.06\%) & 81.30  (-1.45\%) & 82.60  (-0.84\%) & 82.60  (-0.84\%) & 41.60  (+13.66\%) & 61.90  (+19.50\%) & 65.60  (+21.93\%) & 65.60  (+21.93\%) \\ 
  &  \quad+\textbf{\methodname(Ours)}& 40  & 66.40  (-7.52\%) & 77.90  (-5.58\%) & 79.30  (-4.80\%) & 79.30  (-4.80\%) & 42.10  (+15.03\%) & 62.40  (+20.46\%) & 66.20  (+23.05\%) & 66.20  (+23.05\%) \\ 
  &  \quad+\textbf{\methodname(Ours)}& 10  & 63.80  (-11.14\%) & 74.20  (-10.06\%) & 75.20  (-9.72\%) & 75.20  (-9.72\%) & \cellcolor{highlightColor}\textbf{47.70}  (+30.33\%) & \cellcolor{highlightColor}\textbf{69.70}  (+34.56\%) & \cellcolor{highlightColor}\textbf{73.40}  (+36.43\%) & \cellcolor{highlightColor}\textbf{73.40}  (+36.43\%) \\ 
    \cmidrule(lr){2-11}
  &  DSGDetr~\cite{Feng_2021}& -  & \cellcolor{highlightColor}\textbf{71.30}  & \cellcolor{highlightColor}\textbf{82.50}  & \cellcolor{highlightColor}\textbf{83.50}  & \cellcolor{highlightColor}\textbf{83.50}  & 36.50  & 52.50  & 55.20  & 55.20  \\ 
  &  \quad+\textbf{\methodname(Ours)}& 70  & 66.80  (-6.31\%) & 78.90  (-4.36\%) & 80.40  (-3.71\%) & 80.40  (-3.71\%) & 41.60  (+13.97\%) & 63.00  (+20.00\%) & 67.70  (+22.64\%) & 67.70  (+22.64\%) \\ 
  &  \quad+\textbf{\methodname(Ours)}& 40  & 66.10  (-7.29\%) & 77.30  (-6.30\%) & 78.70  (-5.75\%) & 78.70  (-5.75\%) & 42.30  (+15.89\%) & 61.70  (+17.52\%) & 65.30  (+18.30\%) & 65.30  (+18.30\%) \\ 
  &  \quad+\textbf{\methodname(Ours)}& 10  & 63.50  (-10.94\%) & 75.50  (-8.48\%) & 77.10  (-7.66\%) & 77.10  (-7.66\%) & \cellcolor{highlightColor}\textbf{43.90}  (+20.27\%) & \cellcolor{highlightColor}\textbf{65.40}  (+24.57\%) & \cellcolor{highlightColor}\textbf{69.80}  (+26.45\%) & \cellcolor{highlightColor}\textbf{69.80}  (+26.45\%) \\ 
    \cmidrule(lr){2-11}
    \cmidrule(lr){1-11}
    \hline
    \end{tabular}
    }
\end{table*}

\clearpage

\subsection{Scene Graph Anticipation}

\subsubsection{Findings}

Here, $\mathcal{S}$ - represents the amount of things included in the masked dataset. To be more precise, $\mathcal{S}=10\%$ means that only 10\% of the labels are included in training the model for the current epoch and 90\% of the labels are masked, thus voiding their contribution to the loss. So $\mathcal{S} = 70\%$ has more labels contributing to the training loss and $\mathcal{S} = 10\%$ less number of labels contributing to the training loss.

\begin{enumerate} 
    \item Table~\ref{tab:sup_sgg_no_constraint_predcls_30},
Table~\ref{tab:sup_sgg_no_constraint_predcls_50}, Table~\ref{tab:sup_sgg_no_constraint_predcls_70}, Table~\ref{tab:sup_sgg_no_constraint_predcls_90}, compare proposed method's performance across various base methods (STTran++, DSGDet++, SceneSayerODE, and SceneSayerSDE) at different {$\mathcal{F}$} values (0.3, 0.5, 0.7, and 0.9)  for Scene Graph Generation (SGA) task, under the GAGS-No Constraint setting.
        \begin{enumerate}
            \item In Table~\ref{tab:sup_sgg_no_constraint_predcls_30}, SceneSayerODE shows the most consistent gain in lower recall metrics (R@10 and mR@10) when \methodname is included. For {$\mathcal{S}$}=70, improvements with \methodname are substantial, especially for STTran++ (e.g., +13.74\% for R@10). Lower values of {$\mathcal{S}$} (e.g., {$\mathcal{S}$}=10) tend to result in less significant improvements. Metrics like mR@50 and mR@100 remain stable or show slight improvements, emphasizing \methodname's balanced handling of long-tail distributions. 
            \item In Table~\ref{tab:sup_sgg_no_constraint_predcls_50}, Table~\ref{tab:sup_sgg_no_constraint_predcls_70}, Table~\ref{tab:sup_sgg_no_constraint_predcls_90}, \methodname continues to show consistent improvements, especially for mR metrics, with substantial gains seen in SceneSayerODE and SceneSayerSDE for mR@10 and mR@20. The relative improvement in metrics is more pronounced compared to Table~\ref{tab:sup_sgg_no_constraint_predcls_30}, suggesting that \methodname is more impactful as the {$\mathcal{F}$} value increases.
        \end{enumerate}
    \item Table~\ref{tab:sup_sgg_no_constraint_sgcls_30},
Table~\ref{tab:sup_sgg_no_constraint_sgcls_50}, Table~\ref{tab:sup_sgg_no_constraint_sgcls_70}, Table~\ref{tab:sup_sgg_no_constraint_sgcls_90}, evaluates \methodname under the Partially Grounded Action Genome Scenes (PGAGS) - No Constraint setting.  
        \begin{enumerate}
            \item As {$\mathcal{F}$} increases from 0.3 to 0.9, the improvements in mR metrics, particularly for mR@10 and mR@20, become more pronounced. 
            \item \methodname improves both R metrics (favouring head classes) and mR metrics (favouring tail classes) as {$\mathcal{F}$} increases. For instance, significant gains in mR@10 and mR@20 consistently align with moderate or stable improvements in R@10 and R@20 across all models and configurations.
            \item All baseline methods (STTran++, DSGDet++, SceneSayerODE, and SceneSayerSDE) benefit from the inclusion of \textbf{\methodname(Ours)}, though the degree of improvement varies. The results also highlight \methodname's compatibility with high {$\mathcal{S}$}, with significant gains observed at high {$\mathcal{F}$} values (e.g., +49.76\% for mR@10 at F=0.7).
        \end{enumerate}
    \item Table~\ref{tab:sup_sgg_no_constraint_sgdet_30},
Table~\ref{tab:sup_sgg_no_constraint_sgdet_50}, Table~\ref{tab:sup_sgg_no_constraint_sgdet_70}, Table~\ref{tab:sup_sgg_no_constraint_sgdet_90}, present the performance evaluation of \methodname's under the Action Genome Scenes (AGS) in No Constraint graph building strategy, for Scene Graph Generation (SGA). 
        \begin{enumerate}
            \item At lower {$\mathcal{F}$} values (e.g., {$\mathcal{F}$}=0.3), the improvements in mR metrics are moderate. At higher {$\mathcal{F}$} values (e.g., {$\mathcal{F}$}=0.9), mR metrics show substantial improvement, highlighting \methodname's strong performance.
            \item For SceneSayerSDE (Table~\ref{tab:sup_sgg_no_constraint_sgdet_90}, {$\mathcal{F}$}=0.9), mR@10 increases from 19.10 to 29.30 (+53.40\%) with \methodname. 
            \item \methodname achieves balanced gains.For example in Table~\ref{tab:sup_sgg_no_constraint_sgdet_50} ({$\mathcal{F}$}=0.5), for DSGDet++, \methodname improves R@10 from 21.9 to 22.8 (+4.11\%) and mR@10 from 11.80 to 13.50 (+14.41\%).
        \end{enumerate}
    \item Table~\ref{tab:sup_sgg_with_constraint_predcls_30}, 
Table~\ref{tab:sup_sgg_with_constraint_predcls_50}, Table~\ref{tab:sup_sgg_with_constraint_predcls_70}, Table~\ref{tab:sup_sgg_with_constraint_predcls_90}, present the With Constraint evaluation results for Scene Graph Generation (SGA) for GAGS. Table~\ref{tab:sup_sgg_with_constraint_sgcls_30},
Table~\ref{tab:sup_sgg_with_constraint_sgcls_50}, Table~\ref{tab:sup_sgg_with_constraint_sgcls_70}, Table~\ref{tab:sup_sgg_with_constraint_sgcls_90}, present the With Constraint evaluation results for Scene Graph Generation (SGA) for PGAGS. Table~\ref{tab:sup_sgg_with_constraint_sgdet_30},
Table~\ref{tab:sup_sgg_with_constraint_sgdet_50}, Table~\ref{tab:sup_sgg_with_constraint_sgdet_70}, Table~\ref{tab:sup_sgg_with_constraint_sgdet_90} present the With Constraint evaluation results for Scene Graph Generation (SGA) for GAGS. 
        \begin{enumerate}
            \item \methodname consistently improves under constrained settings but with smaller gains than the No Constraint scenario. 
            \item GAGS (Tables 16-19) for fully grounded relationships and constraints result in consistent performance gains with \methodname, especially for SceneSayerSDE. 
            \item Gains in mR metrics dominate, with the highest improvements observed at F=0.9. In PGAGS, gains are moderate compared to GAGS, with mR metrics seeing smaller improvements.
        \end{enumerate}

\end{enumerate}

\clearpage

\subsubsection{Results - No Constraint Setting - Grounded Action Genome Scenes (GAGS)}

\begin{table*}[!h]
    \centering
    \captionsetup{font=small}
    \caption{\textbf{GAGS}-No Constraint-0.3 results for SGA.}
    \label{tab:sup_sgg_no_constraint_predcls_30}
    \renewcommand{\arraystretch}{1.2} 
    \resizebox{\textwidth}{!}{
\begin{tabular}{l|l|l|cccc|cccc}
    \hline
        \multirow{2}{*}{$\mathcal{F}$} & \multirow{2}{*}{Method} & \multirow{2}{*}{$\mathcal{S}$} &  \multicolumn{8}{c}{\textbf{NO CONSTRAINT}} \\ 
        \cmidrule(lr){4-7} \cmidrule(lr){8-11} 
         & & & \textbf{R@10} & \textbf{R@20} & \textbf{R@50} & \textbf{R@100} &\textbf{mR@10} & \textbf{mR@20} & \textbf{mR@50}  & \textbf{mR@100} \\ \hline
   \multirow{16}{*}{0.3} & STTran++~\cite{peddi_et_al_scene_sayer_2024}& -  & 39.30  & 55.60  & \cellcolor{highlightColor}\textbf{65.20}  & \cellcolor{highlightColor}\textbf{65.80}  & 23.10  & 35.50  & 59.20  & \cellcolor{highlightColor}\textbf{70.60}  \\ 
  &  \quad+\textbf{\methodname(Ours)}& 70  & \cellcolor{highlightColor}\textbf{44.70}  (+13.74\%) & \cellcolor{highlightColor}\textbf{57.60}  (+3.60\%) & 65.00  (-0.31\%) & 65.80  & \cellcolor{highlightColor}\textbf{24.60}  (+6.49\%) & \cellcolor{highlightColor}\textbf{37.60}  (+5.92\%) & 59.50  (+0.51\%) & 70.60  \\ 
  &  \quad+\textbf{\methodname(Ours)}& 40  & 40.20  (+2.29\%) & 55.10  (-0.90\%) & 64.80  (-0.61\%) & 65.80  & 21.50  (-6.93\%) & 33.50  (-5.63\%) & 56.20  (-5.07\%) & 69.80  (-1.13\%) \\ 
  &  \quad+\textbf{\methodname(Ours)}& 10  & 40.50  (+3.05\%) & 55.20  (-0.72\%) & 64.60  (-0.92\%) & 65.70  (-0.15\%) & 21.70  (-6.06\%) & 35.70  (+0.56\%) & \cellcolor{highlightColor}\textbf{59.90}  (+1.18\%) & 70.60  \\ 
    \cmidrule(lr){2-11}
  &  DSGDetr++~\cite{peddi_et_al_scene_sayer_2024}& -  & 43.80  & \cellcolor{highlightColor}\textbf{57.80}  & \cellcolor{highlightColor}\textbf{65.10}  & \cellcolor{highlightColor}\textbf{65.80}  & 21.20  & 34.30  & 57.40  & 70.10  \\ 
  &  \quad+\textbf{\methodname(Ours)}& 70  & 44.10  (+0.68\%) & 57.30  (-0.87\%) & 65.10  & 65.80  & 22.10  (+4.25\%) & 35.60  (+3.79\%) & 59.50  (+3.66\%) & 69.90  (-0.29\%) \\
\hline
    \end{tabular}
    }
\end{table*}

\begin{table*}[!h]
    \centering
    \captionsetup{font=small}
    \caption{\textbf{GAGS}-No Constraint-0.5 results for SGA.}
    \label{tab:sup_sgg_no_constraint_predcls_50}
    \renewcommand{\arraystretch}{1.2} 
    \resizebox{\textwidth}{!}{
    \begin{tabular}{l|l|l|cccc|cccc}
    \hline
        \multirow{2}{*}{$\mathcal{F}$} & \multirow{2}{*}{Method} & \multirow{2}{*}{$\mathcal{S}$} &  \multicolumn{8}{c}{\textbf{NO CONSTRAINT}} \\ 
        \cmidrule(lr){4-7} \cmidrule(lr){8-11} 
         & & & \textbf{R@10} & \textbf{R@20} & \textbf{R@50} & \textbf{R@100} &\textbf{mR@10} & \textbf{mR@20} & \textbf{mR@50}  & \textbf{mR@100} \\ \hline
   \multirow{16}{*}{0.5} & STTran++~\cite{peddi_et_al_scene_sayer_2024}& -  & 45.10  & 63.30  & \cellcolor{highlightColor}\textbf{73.40}  & \cellcolor{highlightColor}\textbf{74.00}  & 25.20  & 39.40  & 63.50  & 75.30  \\ 
  &  \quad+\textbf{\methodname(Ours)}& 70  & \cellcolor{highlightColor}\textbf{50.50}  (+11.97\%) & \cellcolor{highlightColor}\textbf{64.90}  (+2.53\%) & 73.30  (-0.14\%) & 74.00  & \cellcolor{highlightColor}\textbf{27.30}  (+8.33\%) & \cellcolor{highlightColor}\textbf{41.60}  (+5.58\%) & \cellcolor{highlightColor}\textbf{64.50}  (+1.57\%) & \cellcolor{highlightColor}\textbf{75.40}  (+0.13\%) \\ 
  &  \quad+\textbf{\methodname(Ours)}& 40  & 45.60  (+1.11\%) & 62.60  (-1.11\%) & 73.00  (-0.54\%) & 74.00  & 23.90  (-5.16\%) & 37.40  (-5.08\%) & 60.70  (-4.41\%) & 74.90  (-0.53\%) \\ 
  &  \quad+\textbf{\methodname(Ours)}& 10  & 45.40  (+0.67\%) & 61.90  (-2.21\%) & 72.60  (-1.09\%) & 73.90  (-0.14\%) & 23.70  (-5.95\%) & 39.20  (-0.51\%) & 64.10  (+0.94\%) & 75.00  (-0.40\%) \\ 
    \cmidrule(lr){2-11}
  &  DSGDetr++~\cite{peddi_et_al_scene_sayer_2024}& -  & 49.50  & \cellcolor{highlightColor}\textbf{65.40}  & \cellcolor{highlightColor}\textbf{73.40}  & \cellcolor{highlightColor}\textbf{74.00}  & 23.20  & 37.30  & 62.90  & 74.90  \\ 
  &  \quad+\textbf{\methodname(Ours)}& 70  & \cellcolor{highlightColor}\textbf{50.10}  (+1.21\%) & 65.00  (-0.61\%) & 73.40  & 74.00  & 24.90  (+7.33\%) & 39.70  (+6.43\%) & 64.10  (+1.91\%) & 75.00  (+0.13\%) \\
\hline
    \end{tabular}
    }
\end{table*}

\begin{table*}[!h]
    \centering
    \captionsetup{font=small}
    \caption{\textbf{GAGS}-No Constraint-0.7 results for SGA.}
    \label{tab:sup_sgg_no_constraint_predcls_70}
    \renewcommand{\arraystretch}{1.2} 
    \resizebox{\textwidth}{!}{
    \begin{tabular}{l|l|l|cccc|cccc}
    \hline
        \multirow{2}{*}{$\mathcal{F}$} & \multirow{2}{*}{Method} & \multirow{2}{*}{$\mathcal{S}$} &  \multicolumn{8}{c}{\textbf{NO CONSTRAINT}} \\ 
        \cmidrule(lr){4-7} \cmidrule(lr){8-11} 
         & & & \textbf{R@10} & \textbf{R@20} & \textbf{R@50} & \textbf{R@100} &\textbf{mR@10} & \textbf{mR@20} & \textbf{mR@50}  & \textbf{mR@100} \\ \hline
   \multirow{16}{*}{0.7} & STTran++~\cite{peddi_et_al_scene_sayer_2024}& -  & 54.70  & 74.20  & \cellcolor{highlightColor}\textbf{83.40}  & \cellcolor{highlightColor}\textbf{83.80}  & 31.20  & 47.00  & 75.40  & 86.00  \\ 
  &  \quad+\textbf{\methodname(Ours)}& 70  & \cellcolor{highlightColor}\textbf{60.40}  (+10.42\%) & \cellcolor{highlightColor}\textbf{76.10}  (+2.56\%) & 83.40  & 83.80  & \cellcolor{highlightColor}\textbf{34.40}  (+10.26\%) & \cellcolor{highlightColor}\textbf{51.50}  (+9.57\%) & 76.10  (+0.93\%) & 86.00  \\ 
  &  \quad+\textbf{\methodname(Ours)}& 40  & 56.40  (+3.11\%) & 73.60  (-0.81\%) & 83.10  (-0.36\%) & 83.80  & 30.70  (-1.60\%) & 46.50  (-1.06\%) & 73.30  (-2.79\%) & 85.90  (-0.12\%) \\ 
  &  \quad+\textbf{\methodname(Ours)}& 10  & 54.40  (-0.55\%) & 72.60  (-2.16\%) & 82.90  (-0.60\%) & 83.80  & 31.10  (-0.32\%) & 49.20  (+4.68\%) & \cellcolor{highlightColor}\textbf{76.50}  (+1.46\%) & \cellcolor{highlightColor}\textbf{86.20}  (+0.23\%) \\ 
    \cmidrule(lr){2-11}
  &  DSGDetr++~\cite{peddi_et_al_scene_sayer_2024}& -  & 59.60  & \cellcolor{highlightColor}\textbf{76.10}  & \cellcolor{highlightColor}\textbf{83.40}  & \cellcolor{highlightColor}\textbf{83.80}  & 28.60  & 46.10  & 73.80  & 85.80  \\ 
  &  \quad+\textbf{\methodname(Ours)}& 70  & \cellcolor{highlightColor}\textbf{60.20}  (+1.01\%) & 75.80  (-0.39\%) & 83.40  & 83.80  & 32.40  (+13.29\%) & 49.70  (+7.81\%) & 76.00  (+2.98\%) & 86.00  (+0.23\%) \\
\hline
    \end{tabular}
    }
\end{table*}

\begin{table*}[!h]
    \centering
    \captionsetup{font=small}
    \caption{\textbf{GAGS}-No Constraint-0.9 results for SGA.}
    \label{tab:sup_sgg_no_constraint_predcls_90}
    \renewcommand{\arraystretch}{1.2} 
    \resizebox{\textwidth}{!}{
    \begin{tabular}{l|l|l|cccc|cccc}
    \hline
        \multirow{2}{*}{$\mathcal{F}$} & \multirow{2}{*}{Method} & \multirow{2}{*}{$\mathcal{S}$} &  \multicolumn{8}{c}{\textbf{NO CONSTRAINT}} \\ 
        \cmidrule(lr){4-7} \cmidrule(lr){8-11} 
         & & & \textbf{R@10} & \textbf{R@20} & \textbf{R@50} & \textbf{R@100} &\textbf{mR@10} & \textbf{mR@20} & \textbf{mR@50}  & \textbf{mR@100} \\ \hline
   \multirow{16}{*}{0.9} & STTran++~\cite{peddi_et_al_scene_sayer_2024}& -  & 68.70  & 86.80  & \cellcolor{highlightColor}\textbf{93.50}  & \cellcolor{highlightColor}\textbf{93.80}  & 42.50  & 60.80  & 84.80  & \cellcolor{highlightColor}\textbf{94.90}  \\ 
  &  \quad+\textbf{\methodname(Ours)}& 70  & \cellcolor{highlightColor}\textbf{74.30}  (+8.15\%) & \cellcolor{highlightColor}\textbf{88.00}  (+1.38\%) & 93.50  & 93.80  & \cellcolor{highlightColor}\textbf{47.50}  (+11.76\%) & \cellcolor{highlightColor}\textbf{64.70}  (+6.41\%) & 89.00  (+4.95\%) & 94.90  \\ 
  &  \quad+\textbf{\methodname(Ours)}& 40  & 70.50  (+2.62\%) & 86.20  (-0.69\%) & 93.50  & 93.80  & 43.00  (+1.18\%) & 61.00  (+0.33\%) & 88.50  (+4.36\%) & 94.70  (-0.21\%) \\ 
  &  \quad+\textbf{\methodname(Ours)}& 10  & 67.50  (-1.75\%) & 85.20  (-1.84\%) & 93.40  (-0.11\%) & 93.80  & 46.30  (+8.94\%) & 64.40  (+5.92\%) & \cellcolor{highlightColor}\textbf{91.40}  (+7.78\%) & 94.90  \\ 
    \cmidrule(lr){2-11}
  &  DSGDetr++~\cite{peddi_et_al_scene_sayer_2024}& -  & 73.60  & \cellcolor{highlightColor}\textbf{88.30}  & \cellcolor{highlightColor}\textbf{93.50}  & \cellcolor{highlightColor}\textbf{93.80}  & 39.00  & 58.80  & 83.20  & \cellcolor{highlightColor}\textbf{94.90}  \\ 
  &  \quad+\textbf{\methodname(Ours)}& 70  & \cellcolor{highlightColor}\textbf{74.10}  (+0.68\%) & 87.90  (-0.45\%) & 93.50  & 93.80  & 46.30  (+18.72\%) & 63.50  (+7.99\%) & 88.90  (+6.85\%) & 94.90  \\
\hline
    \end{tabular}
    }
\end{table*}

\clearpage

\subsubsection{Results - No Constraint Setting - Partially Grounded Action Genome Scenes (PGAGS)}

\begin{table*}[!h]
    \centering
    \captionsetup{font=small}
    \caption{\textbf{PGAGS}-No Constraint-0.3 results for SGA.}
    \label{tab:sup_sgg_no_constraint_sgcls_30}
    \renewcommand{\arraystretch}{1.2} 
    \resizebox{\textwidth}{!}{
    \begin{tabular}{l|l|l|cccc|cccc}
    \hline
        \multirow{2}{*}{$\mathcal{F}$} & \multirow{2}{*}{Method} & \multirow{2}{*}{$\mathcal{S}$} &  \multicolumn{8}{c}{\textbf{NO CONSTRAINT}} \\ 
        \cmidrule(lr){4-7} \cmidrule(lr){8-11} 
         & & & \textbf{R@10} & \textbf{R@20} & \textbf{R@50} & \textbf{R@100} &\textbf{mR@10} & \textbf{mR@20} & \textbf{mR@50}  & \textbf{mR@100} \\ \hline
   \multirow{16}{*}{0.3} & STTran++~\cite{peddi_et_al_scene_sayer_2024}& -  & 31.00  & 42.30  & 46.60  & 46.80  & 18.50  & 28.70  & 48.80  & 52.50  \\ 
  &  \quad+\textbf{\methodname(Ours)}& 70  & 33.20  (+7.10\%) & 41.30  (-2.36\%) & 46.60  & \cellcolor{highlightColor}\textbf{47.00}  (+0.43\%) & 16.60  (-10.27\%) & 27.20  (-5.23\%) & 49.30  (+1.02\%) & \cellcolor{highlightColor}\textbf{52.80}  (+0.57\%) \\ 
  &  \quad+\textbf{\methodname(Ours)}& 40  & \cellcolor{highlightColor}\textbf{34.00}  (+9.68\%) & \cellcolor{highlightColor}\textbf{42.50}  (+0.47\%) & \cellcolor{highlightColor}\textbf{46.70}  (+0.21\%) & 47.00  (+0.43\%) & 19.70  (+6.49\%) & 29.60  (+3.14\%) & \cellcolor{highlightColor}\textbf{49.60}  (+1.64\%) & 52.80  (+0.57\%) \\ 
  &  \quad+\textbf{\methodname(Ours)}& 10  & 29.40  (-5.16\%) & 40.20  (-4.96\%) & 45.80  (-1.72\%) & 46.20  (-1.28\%) & \cellcolor{highlightColor}\textbf{21.20}  (+14.59\%) & \cellcolor{highlightColor}\textbf{31.50}  (+9.76\%) & 49.20  (+0.82\%) & 52.40  (-0.19\%) \\ 
    \cmidrule(lr){2-11}
  &  DSGDetr++~\cite{peddi_et_al_scene_sayer_2024}& -  & \cellcolor{highlightColor}\textbf{35.60}  & \cellcolor{highlightColor}\textbf{44.00}  & \cellcolor{highlightColor}\textbf{48.20}  & \cellcolor{highlightColor}\textbf{48.50}  & 17.90  & 28.60  & 49.20  & \cellcolor{highlightColor}\textbf{53.80}  \\ 
  &  \quad+\textbf{\methodname(Ours)}& 70  & 34.40  (-3.37\%) & 43.50  (-1.14\%) & 48.20  & 48.50  & \cellcolor{highlightColor}\textbf{21.10}  (+17.88\%) & 30.50  (+6.64\%) & 49.70  (+1.02\%) & 53.80  \\
\hline
    \end{tabular}
    }
\end{table*}

\begin{table*}[!h]
    \centering
    \captionsetup{font=small}
    \caption{\textbf{PGAGS}-No Constraint-0.5 results for SGA.}
    \label{tab:sup_sgg_no_constraint_sgcls_50}
    \renewcommand{\arraystretch}{1.2} 
    \resizebox{\textwidth}{!}{
    \begin{tabular}{l|l|l|cccc|cccc}
    \hline
        \multirow{2}{*}{$\mathcal{F}$} & \multirow{2}{*}{Method} & \multirow{2}{*}{$\mathcal{S}$} &  \multicolumn{8}{c}{\textbf{NO CONSTRAINT}} \\ 
        \cmidrule(lr){4-7} \cmidrule(lr){8-11} 
         & & & \textbf{R@10} & \textbf{R@20} & \textbf{R@50} & \textbf{R@100} &\textbf{mR@10} & \textbf{mR@20} & \textbf{mR@50}  & \textbf{mR@100} \\ \hline
   \multirow{16}{*}{0.5} & STTran++~\cite{peddi_et_al_scene_sayer_2024}& -  & 34.30  & 46.70  & \cellcolor{highlightColor}\textbf{51.50}  & \cellcolor{highlightColor}\textbf{51.60}  & 20.90  & 32.50  & 50.10  & \cellcolor{highlightColor}\textbf{53.20}  \\ 
  &  \quad+\textbf{\methodname(Ours)}& 70  & 36.60  (+6.71\%) & 45.70  (-2.14\%) & 51.10  (-0.78\%) & 51.50  (-0.19\%) & 18.70  (-10.53\%) & 30.70  (-5.54\%) & 49.40  (-1.40\%) & 52.70  (-0.94\%) \\ 
  &  \quad+\textbf{\methodname(Ours)}& 40  & \cellcolor{highlightColor}\textbf{37.30}  (+8.75\%) & \cellcolor{highlightColor}\textbf{46.90}  (+0.43\%) & 51.10  (-0.78\%) & 51.30  (-0.58\%) & 21.90  (+4.78\%) & 33.20  (+2.15\%) & 50.00  (-0.20\%) & 52.30  (-1.69\%) \\ 
  &  \quad+\textbf{\methodname(Ours)}& 10  & 32.50  (-5.25\%) & 44.10  (-5.57\%) & 50.40  (-2.14\%) & 50.80  (-1.55\%) & \cellcolor{highlightColor}\textbf{24.00}  (+14.83\%) & \cellcolor{highlightColor}\textbf{34.90}  (+7.38\%) & \cellcolor{highlightColor}\textbf{50.50}  (+0.80\%) & 52.70  (-0.94\%) \\ 
    \cmidrule(lr){2-11}
  &  DSGDetr++~\cite{peddi_et_al_scene_sayer_2024}& -  & \cellcolor{highlightColor}\textbf{39.60}  & \cellcolor{highlightColor}\textbf{49.60}  & \cellcolor{highlightColor}\textbf{54.30}  & \cellcolor{highlightColor}\textbf{54.50}  & 19.90  & 32.30  & 50.60  & 55.80  \\ 
  &  \quad+\textbf{\methodname(Ours)}& 70  & 37.40  (-5.56\%) & 48.30  (-2.62\%) & 53.70  (-1.10\%) & 54.00  (-0.92\%) & \cellcolor{highlightColor}\textbf{23.10}  (+16.08\%) & 33.60  (+4.02\%) & 50.50  (-0.20\%) & 55.30  (-0.90\%) \\
\hline
    \end{tabular}
    }
\end{table*}

\begin{table*}[!h]
    \centering
    \captionsetup{font=small}
    \caption{\textbf{PGAGS}-No Constraint-0.7 results for SGA.}
    \label{tab:sup_sgg_no_constraint_sgcls_70}
    \renewcommand{\arraystretch}{1.2} 
    \resizebox{\textwidth}{!}{
    \begin{tabular}{l|l|l|cccc|cccc}
    \hline
        \multirow{2}{*}{$\mathcal{F}$} & \multirow{2}{*}{Method} & \multirow{2}{*}{$\mathcal{S}$} &  \multicolumn{8}{c}{\textbf{NO CONSTRAINT}} \\ 
        \cmidrule(lr){4-7} \cmidrule(lr){8-11} 
         & & & \textbf{R@10} & \textbf{R@20} & \textbf{R@50} & \textbf{R@100} &\textbf{mR@10} & \textbf{mR@20} & \textbf{mR@50}  & \textbf{mR@100} \\ \hline
   \multirow{16}{*}{0.7} & STTran++~\cite{peddi_et_al_scene_sayer_2024}& -  & 41.40  & 54.90  & \cellcolor{highlightColor}\textbf{59.60}  & \cellcolor{highlightColor}\textbf{59.70}  & 25.30  & 38.30  & 56.10  & 58.40  \\ 
  &  \quad+\textbf{\methodname(Ours)}& 70  & 44.80  (+8.21\%) & 54.40  (-0.91\%) & 59.30  (-0.50\%) & 59.50  (-0.34\%) & 24.40  (-3.56\%) & 38.70  (+1.04\%) & 56.40  (+0.53\%) & \cellcolor{highlightColor}\textbf{59.90}  (+2.57\%) \\ 
  &  \quad+\textbf{\methodname(Ours)}& 40  & \cellcolor{highlightColor}\textbf{45.50}  (+9.90\%) & \cellcolor{highlightColor}\textbf{55.50}  (+1.09\%) & 59.60  & 59.70  & 28.00  (+10.67\%) & 41.70  (+8.88\%) & 57.10  (+1.78\%) & 59.30  (+1.54\%) \\ 
  &  \quad+\textbf{\methodname(Ours)}& 10  & 39.60  (-4.35\%) & 53.00  (-3.46\%) & 59.10  (-0.84\%) & 59.50  (-0.34\%) & \cellcolor{highlightColor}\textbf{29.80}  (+17.79\%) & \cellcolor{highlightColor}\textbf{43.90}  (+14.62\%) & \cellcolor{highlightColor}\textbf{57.60}  (+2.67\%) & 59.40  (+1.71\%) \\ 
    \cmidrule(lr){2-11}
  &  DSGDetr++~\cite{peddi_et_al_scene_sayer_2024}& -  & \cellcolor{highlightColor}\textbf{47.70}  & \cellcolor{highlightColor}\textbf{58.10}  & \cellcolor{highlightColor}\textbf{62.40}  & \cellcolor{highlightColor}\textbf{62.60}  & 24.80  & 39.50  & 57.30  & 61.10  \\ 
  &  \quad+\textbf{\methodname(Ours)}& 70  & 45.40  (-4.82\%) & 57.30  (-1.38\%) & 62.10  (-0.48\%) & 62.20  (-0.64\%) & \cellcolor{highlightColor}\textbf{28.70}  (+15.73\%) & 40.90  (+3.54\%) & 57.60  (+0.52\%) & 60.80  (-0.49\%) \\
\hline
    \end{tabular}
    }
\end{table*}

\begin{table*}[!h]
    \centering
    \captionsetup{font=small}
    \caption{\textbf{PGAGS}-No Constraint-0.9 results for SGA.}
    \label{tab:sup_sgg_no_constraint_sgcls_90}
    \renewcommand{\arraystretch}{1.2} 
    \resizebox{\textwidth}{!}{
    \begin{tabular}{l|l|l|cccc|cccc}
    \hline
        \multirow{2}{*}{$\mathcal{F}$} & \multirow{2}{*}{Method} & \multirow{2}{*}{$\mathcal{S}$} &  \multicolumn{8}{c}{\textbf{NO CONSTRAINT}} \\ 
        \cmidrule(lr){4-7} \cmidrule(lr){8-11} 
         & & & \textbf{R@10} & \textbf{R@20} & \textbf{R@50} & \textbf{R@100} &\textbf{mR@10} & \textbf{mR@20} & \textbf{mR@50}  & \textbf{mR@100} \\ \hline
   \multirow{16}{*}{0.9} & STTran++~\cite{peddi_et_al_scene_sayer_2024}& -  & 50.00  & 61.60  & 64.40  & 64.50  & 32.40  & 46.90  & 60.60  & 62.60  \\ 
  &  \quad+\textbf{\methodname(Ours)}& 70  & \cellcolor{highlightColor}\textbf{54.00}  (+8.00\%) & \cellcolor{highlightColor}\textbf{62.30}  (+1.14\%) & \cellcolor{highlightColor}\textbf{65.00}  (+0.93\%) & 65.00  (+0.78\%) & 35.10  (+8.33\%) & 48.70  (+3.84\%) & 61.40  (+1.32\%) & 63.10  (+0.80\%) \\ 
  &  \quad+\textbf{\methodname(Ours)}& 40  & 53.30  (+6.60\%) & 62.20  (+0.97\%) & 65.00  (+0.93\%) & \cellcolor{highlightColor}\textbf{65.10}  (+0.93\%) & 36.80  (+13.58\%) & 49.60  (+5.76\%) & 62.10  (+2.48\%) & 63.40  (+1.28\%) \\ 
  &  \quad+\textbf{\methodname(Ours)}& 10  & 48.20  (-3.60\%) & 60.50  (-1.79\%) & 64.90  (+0.78\%) & 65.10  (+0.93\%) & \cellcolor{highlightColor}\textbf{40.90}  (+26.23\%) & \cellcolor{highlightColor}\textbf{52.10}  (+11.09\%) & \cellcolor{highlightColor}\textbf{63.80}  (+5.28\%) & \cellcolor{highlightColor}\textbf{64.60}  (+3.19\%) \\ 
    \cmidrule(lr){2-11}
  &  DSGDetr++~\cite{peddi_et_al_scene_sayer_2024}& -  & \cellcolor{highlightColor}\textbf{57.10}  & \cellcolor{highlightColor}\textbf{66.40}  & \cellcolor{highlightColor}\textbf{69.40}  & \cellcolor{highlightColor}\textbf{69.50}  & 33.00  & 49.00  & 64.20  & 68.20  \\ 
  &  \quad+\textbf{\methodname(Ours)}& 70  & 54.90  (-3.85\%) & 65.40  (-1.51\%) & 68.80  (-0.86\%) & 68.90  (-0.86\%) & 35.90  (+8.79\%) & 49.40  (+0.82\%) & 63.90  (-0.47\%) & 67.50  (-1.03\%) \\
\hline
    \end{tabular}
    }
\end{table*}

\clearpage

\subsubsection{Results - No Constraint Setting - Action Genome Scenes (AGS)}

\begin{table*}[!h]
    \centering
    \captionsetup{font=small}
    \caption{\textbf{AGS}-No Constraint-0.3 results for SGA.}
    \label{tab:sup_sgg_no_constraint_sgdet_30}
    \renewcommand{\arraystretch}{1.2} 
    \resizebox{\textwidth}{!}{
    \begin{tabular}{l|l|l|cccc|cccc}
    \hline
        \multirow{2}{*}{$\mathcal{F}$} & \multirow{2}{*}{Method} & \multirow{2}{*}{$\mathcal{S}$} &  \multicolumn{8}{c}{\textbf{NO CONSTRAINT}} \\ 
        \cmidrule(lr){4-7} \cmidrule(lr){8-11} 
         & & & \textbf{R@10} & \textbf{R@20} & \textbf{R@50} & \textbf{R@100} &\textbf{mR@10} & \textbf{mR@20} & \textbf{mR@50}  & \textbf{mR@100} \\ \hline
   \multirow{16}{*}{0.3} & STTran++~\cite{peddi_et_al_scene_sayer_2024}& -  & \cellcolor{highlightColor}\textbf{22.90}  & \cellcolor{highlightColor}\textbf{36.00}  & \cellcolor{highlightColor}\textbf{51.30}  & \cellcolor{highlightColor}\textbf{55.20}  & \cellcolor{highlightColor}\textbf{13.10}  & 20.20  & 36.20  & 49.90  \\ 
  &  \quad+\textbf{\methodname(Ours)}& 70  & 22.70  (-0.87\%) & 35.80  (-0.56\%) & 50.50  (-1.56\%) & 55.20  & 12.20  (-6.87\%) & 19.00  (-5.94\%) & 35.70  (-1.38\%) & 50.30  (+0.80\%) \\ 
  &  \quad+\textbf{\methodname(Ours)}& 40  & 22.10  (-3.49\%) & 35.50  (-1.39\%) & 50.50  (-1.56\%) & 55.00  (-0.36\%) & 12.80  (-2.29\%) & \cellcolor{highlightColor}\textbf{21.10}  (+4.46\%) & \cellcolor{highlightColor}\textbf{37.50}  (+3.59\%) & \cellcolor{highlightColor}\textbf{51.10}  (+2.40\%) \\ 
  &  \quad+\textbf{\methodname(Ours)}& 10  & 19.50  (-14.85\%) & 30.40  (-15.56\%) & 48.50  (-5.46\%) & 54.90  (-0.54\%) & 11.10  (-15.27\%) & 18.50  (-8.42\%) & 36.50  (+0.83\%) & 51.10  (+2.40\%) \\ 
    \cmidrule(lr){2-11}
  &  DSGDetr++~\cite{peddi_et_al_scene_sayer_2024}& -  & 20.80  & 33.10  & \cellcolor{highlightColor}\textbf{49.20}  & 53.90  & 11.50  & 17.80  & 34.10  & 49.10  \\ 
  &  \quad+\textbf{\methodname(Ours)}& 70  & \cellcolor{highlightColor}\textbf{21.30}  (+2.40\%) & \cellcolor{highlightColor}\textbf{33.60}  (+1.51\%) & 49.00  (-0.41\%) & \cellcolor{highlightColor}\textbf{54.30}  (+0.74\%) & 12.40  (+7.83\%) & 19.30  (+8.43\%) & 34.90  (+2.35\%) & 48.90  (-0.41\%) \\
\hline
    \end{tabular}
    }
\end{table*}

\begin{table*}[!h]
    \centering
    \captionsetup{font=small}
    \caption{\textbf{AGS}-No Constraint-0.5 results for SGA.}
    \label{tab:sup_sgg_no_constraint_sgdet_50}
    \renewcommand{\arraystretch}{1.2} 
    \resizebox{\textwidth}{!}{
    \begin{tabular}{l|l|l|cccc|cccc}
    \hline
        \multirow{2}{*}{$\mathcal{F}$} & \multirow{2}{*}{Method} & \multirow{2}{*}{$\mathcal{S}$} &  \multicolumn{8}{c}{\textbf{NO CONSTRAINT}} \\ 
        \cmidrule(lr){4-7} \cmidrule(lr){8-11} 
         & & & \textbf{R@10} & \textbf{R@20} & \textbf{R@50} & \textbf{R@100} &\textbf{mR@10} & \textbf{mR@20} & \textbf{mR@50}  & \textbf{mR@100} \\ \hline
   \multirow{16}{*}{0.5} & STTran++~\cite{peddi_et_al_scene_sayer_2024}& -  & \cellcolor{highlightColor}\textbf{24.60}  & \cellcolor{highlightColor}\textbf{38.40}  & \cellcolor{highlightColor}\textbf{54.50}  & \cellcolor{highlightColor}\textbf{58.60}  & \cellcolor{highlightColor}\textbf{13.90}  & \cellcolor{highlightColor}\textbf{21.30}  & 38.50  & 52.50  \\ 
  &  \quad+\textbf{\methodname(Ours)}& 70  & 24.00  (-2.44\%) & 38.10  (-0.78\%) & 54.00  (-0.92\%) & 58.60  & 12.60  (-9.35\%) & 19.90  (-6.57\%) & 37.90  (-1.56\%) & 53.40  (+1.71\%) \\ 
  &  \quad+\textbf{\methodname(Ours)}& 40  & 23.10  (-6.10\%) & 37.60  (-2.08\%) & 53.60  (-1.65\%) & 58.50  (-0.17\%) & 12.90  (-7.19\%) & 21.30  & 39.00  (+1.30\%) & 53.90  (+2.67\%) \\ 
  &  \quad+\textbf{\methodname(Ours)}& 10  & 21.00  (-14.63\%) & 33.10  (-13.80\%) & 51.90  (-4.77\%) & 58.40  (-0.34\%) & 12.10  (-12.95\%) & 20.00  (-6.10\%) & \cellcolor{highlightColor}\textbf{39.60}  (+2.86\%) & \cellcolor{highlightColor}\textbf{54.30}  (+3.43\%) \\ 
    \cmidrule(lr){2-11}
  &  DSGDetr++~\cite{peddi_et_al_scene_sayer_2024}& -  & 21.90  & 34.70  & 52.10  & 57.30  & 11.80  & 18.20  & 36.10  & 52.70  \\ 
  &  \quad+\textbf{\methodname(Ours)}& 70  & \cellcolor{highlightColor}\textbf{22.80}  (+4.11\%) & \cellcolor{highlightColor}\textbf{36.10}  (+4.03\%) & \cellcolor{highlightColor}\textbf{52.90}  (+1.54\%) & \cellcolor{highlightColor}\textbf{58.30}  (+1.75\%) & 13.10  (+11.02\%) & 20.60  (+13.19\%) & 37.80  (+4.71\%) & 55.60  (+5.50\%) \\
\hline
    \end{tabular}
    }
\end{table*}

\begin{table*}[!h]
    \centering
    \captionsetup{font=small}
    \caption{\textbf{AGS}-No Constraint-0.7 results for SGA.}
    \label{tab:sup_sgg_no_constraint_sgdet_70}
    \renewcommand{\arraystretch}{1.2} 
    \resizebox{\textwidth}{!}{
    \begin{tabular}{l|l|l|cccc|cccc}
    \hline
        \multirow{2}{*}{$\mathcal{F}$} & \multirow{2}{*}{Method} & \multirow{2}{*}{$\mathcal{S}$} &  \multicolumn{8}{c}{\textbf{NO CONSTRAINT}} \\ 
        \cmidrule(lr){4-7} \cmidrule(lr){8-11} 
         & & & \textbf{R@10} & \textbf{R@20} & \textbf{R@50} & \textbf{R@100} &\textbf{mR@10} & \textbf{mR@20} & \textbf{mR@50}  & \textbf{mR@100} \\ \hline
   \multirow{16}{*}{0.7} & STTran++~\cite{peddi_et_al_scene_sayer_2024}& -  & \cellcolor{highlightColor}\textbf{27.20}  & 42.10  & \cellcolor{highlightColor}\textbf{59.70}  & \cellcolor{highlightColor}\textbf{63.90}  & \cellcolor{highlightColor}\textbf{15.70}  & 23.70  & 41.90  & 57.50  \\ 
  &  \quad+\textbf{\methodname(Ours)}& 70  & 26.80  (-1.47\%) & \cellcolor{highlightColor}\textbf{43.00}  (+2.14\%) & 59.70  & 63.90  & 14.90  (-5.10\%) & 23.90  (+0.84\%) & 42.60  (+1.67\%) & 58.30  (+1.39\%) \\ 
  &  \quad+\textbf{\methodname(Ours)}& 40  & 25.80  (-5.15\%) & 41.80  (-0.71\%) & 59.30  (-0.67\%) & 63.80  (-0.16\%) & 14.90  (-5.10\%) & \cellcolor{highlightColor}\textbf{24.70}  (+4.22\%) & \cellcolor{highlightColor}\textbf{44.00}  (+5.01\%) & 58.90  (+2.43\%) \\ 
  &  \quad+\textbf{\methodname(Ours)}& 10  & 23.60  (-13.24\%) & 36.50  (-13.30\%) & 57.00  (-4.52\%) & 63.50  (-0.63\%) & 14.00  (-10.83\%) & 23.20  (-2.11\%) & 43.70  (+4.30\%) & \cellcolor{highlightColor}\textbf{59.60}  (+3.65\%) \\ 
    \cmidrule(lr){2-11}
  &  DSGDetr++~\cite{peddi_et_al_scene_sayer_2024}& -  & \cellcolor{highlightColor}\textbf{24.90}  & 38.20  & 56.90  & 62.00  & 13.20  & 20.00  & 38.80  & 56.50  \\ 
  &  \quad+\textbf{\methodname(Ours)}& 70  & 24.80  (-0.40\%) & \cellcolor{highlightColor}\textbf{40.00}  (+4.71\%) & \cellcolor{highlightColor}\textbf{57.80}  (+1.58\%) & \cellcolor{highlightColor}\textbf{63.20}  (+1.94\%) & 14.40  (+9.09\%) & 23.20  (+16.00\%) & 41.30  (+6.44\%) & 56.90  (+0.71\%) \\
\hline
    \end{tabular}
    }
\end{table*}

\begin{table*}[!h]
    \centering
    \captionsetup{font=small}
    \caption{\textbf{AGS}-No Constraint-0.9 results for SGA.}
    \label{tab:sup_sgg_no_constraint_sgdet_90}
    \renewcommand{\arraystretch}{1.2} 
    \resizebox{\textwidth}{!}{
    \begin{tabular}{l|l|l|cccc|cccc}
    \hline
        \multirow{2}{*}{$\mathcal{F}$} & \multirow{2}{*}{Method} & \multirow{2}{*}{$\mathcal{S}$} &  \multicolumn{8}{c}{\textbf{NO CONSTRAINT}} \\ 
        \cmidrule(lr){4-7} \cmidrule(lr){8-11} 
         & & & \textbf{R@10} & \textbf{R@20} & \textbf{R@50} & \textbf{R@100} &\textbf{mR@10} & \textbf{mR@20} & \textbf{mR@50}  & \textbf{mR@100} \\ \hline
   \multirow{16}{*}{0.9} & STTran++~\cite{peddi_et_al_scene_sayer_2024}& -  & \cellcolor{highlightColor}\textbf{30.20}  & 45.80  & 64.40  & \cellcolor{highlightColor}\textbf{68.40}  & 18.20  & 27.50  & 47.20  & 62.80  \\ 
  &  \quad+\textbf{\methodname(Ours)}& 70  & 30.10  (-0.33\%) & \cellcolor{highlightColor}\textbf{47.80}  (+4.37\%) & \cellcolor{highlightColor}\textbf{64.50}  (+0.16\%) & 68.40  & 19.50  (+7.14\%) & 29.60  (+7.64\%) & 50.00  (+5.93\%) & 64.00  (+1.91\%) \\ 
  &  \quad+\textbf{\methodname(Ours)}& 40  & 29.30  (-2.98\%) & 47.10  (+2.84\%) & 64.50  (+0.16\%) & 68.30  (-0.15\%) & \cellcolor{highlightColor}\textbf{20.50}  (+12.64\%) & \cellcolor{highlightColor}\textbf{31.80}  (+15.64\%) & \cellcolor{highlightColor}\textbf{51.90}  (+9.96\%) & 63.80  (+1.59\%) \\ 
  &  \quad+\textbf{\methodname(Ours)}& 10  & 26.30  (-12.91\%) & 40.70  (-11.14\%) & 60.90  (-5.43\%) & 68.00  (-0.58\%) & 18.80  (+3.30\%) & 30.40  (+10.55\%) & 51.40  (+8.90\%) & \cellcolor{highlightColor}\textbf{64.80}  (+3.18\%) \\ 
    \cmidrule(lr){2-11}
  &  DSGDetr++~\cite{peddi_et_al_scene_sayer_2024}& -  & \cellcolor{highlightColor}\textbf{28.70}  & 43.40  & 62.70  & 67.10  & 15.50  & 23.80  & 44.90  & 61.70  \\ 
  &  \quad+\textbf{\methodname(Ours)}& 70  & 27.90  (-2.79\%) & \cellcolor{highlightColor}\textbf{45.10}  (+3.92\%) & \cellcolor{highlightColor}\textbf{63.90}  (+1.91\%) & \cellcolor{highlightColor}\textbf{68.40}  (+1.94\%) & 18.50  (+19.35\%) & 28.70  (+20.59\%) & 48.50  (+8.02\%) & 62.00  (+0.49\%) \\
\hline
    \end{tabular}
    }
\end{table*}

\clearpage

\subsubsection{Results - With Constraint Setting - Grounded Action Genome Scenes (GAGS)}

\begin{table*}[!h]
    \centering
    \captionsetup{font=small}
    \caption{\textbf{GAGS}-With Constraint-0.3 results for SGA.}
    \label{tab:sup_sgg_with_constraint_predcls_30}
    \renewcommand{\arraystretch}{1.2} 
    \resizebox{\textwidth}{!}{
    \begin{tabular}{l|l|l|cccc|cccc}
    \hline
        \multirow{2}{*}{$\mathcal{F}$} & \multirow{2}{*}{Method} & \multirow{2}{*}{$\mathcal{S}$} &  \multicolumn{8}{c}{\textbf{WITH CONSTRAINT}} \\ 
        \cmidrule(lr){4-7} \cmidrule(lr){8-11} 
         & & & \textbf{R@10} & \textbf{R@20} & \textbf{R@50} & \textbf{R@100} &\textbf{mR@10} & \textbf{mR@20} & \textbf{mR@50}  & \textbf{mR@100} \\ \hline
   \multirow{16}{*}{0.3} & STTran++~\cite{peddi_et_al_scene_sayer_2024}& -  & 26.50  & 29.50  & 29.60  & 29.60  & 15.90  & 18.50  & 18.60  & 18.60  \\ 
  &  \quad+\textbf{\methodname(Ours)}& 70  & \cellcolor{highlightColor}\textbf{36.60}  (+38.11\%) & \cellcolor{highlightColor}\textbf{39.70}  (+34.58\%) & \cellcolor{highlightColor}\textbf{39.80}  (+34.46\%) & \cellcolor{highlightColor}\textbf{39.80}  (+34.46\%) & \cellcolor{highlightColor}\textbf{18.70}  (+17.61\%) & \cellcolor{highlightColor}\textbf{21.40}  (+15.68\%) & \cellcolor{highlightColor}\textbf{21.40}  (+15.05\%) & \cellcolor{highlightColor}\textbf{21.40}  (+15.05\%) \\ 
  &  \quad+\textbf{\methodname(Ours)}& 40  & 32.60  (+23.02\%) & 35.60  (+20.68\%) & 35.60  (+20.27\%) & 35.60  (+20.27\%) & 17.00  (+6.92\%) & 19.40  (+4.86\%) & 19.40  (+4.30\%) & 19.40  (+4.30\%) \\ 
  &  \quad+\textbf{\methodname(Ours)}& 10  & 33.30  (+25.66\%) & 35.80  (+21.36\%) & 35.80  (+20.95\%) & 35.80  (+20.95\%) & 17.80  (+11.95\%) & 20.40  (+10.27\%) & 20.40  (+9.68\%) & 20.40  (+9.68\%) \\ 
    \cmidrule(lr){2-11}
  &  DSGDetr++~\cite{peddi_et_al_scene_sayer_2024}& -  & 36.20  & 39.30  & 39.30  & 39.30  & 15.10  & 17.50  & 17.50  & 17.50  \\ 
  &  \quad+\textbf{\methodname(Ours)}& 70  & \cellcolor{highlightColor}\textbf{37.00}  (+2.21\%) & \cellcolor{highlightColor}\textbf{40.20}  (+2.29\%) & \cellcolor{highlightColor}\textbf{40.30}  (+2.54\%) & \cellcolor{highlightColor}\textbf{40.30}  (+2.54\%) & 17.20  (+13.91\%) & 19.90  (+13.71\%) & 20.00  (+14.29\%) & 20.00  (+14.29\%) \\
\hline
    \end{tabular}
    }
\end{table*}

\begin{table*}[!h]
    \centering
    \captionsetup{font=small}
    \caption{\textbf{GAGS}-With Constraint-0.5 results for SGA.}
    \label{tab:sup_sgg_with_constraint_predcls_50}
    \renewcommand{\arraystretch}{1.2} 
    \resizebox{\textwidth}{!}{
    \begin{tabular}{l|l|l|cccc|cccc}
    \hline
        \multirow{2}{*}{$\mathcal{F}$} & \multirow{2}{*}{Method} & \multirow{2}{*}{$\mathcal{S}$} &  \multicolumn{8}{c}{\textbf{WITH CONSTRAINT}} \\ 
        \cmidrule(lr){4-7} \cmidrule(lr){8-11} 
         & & & \textbf{R@10} & \textbf{R@20} & \textbf{R@50} & \textbf{R@100} &\textbf{mR@10} & \textbf{mR@20} & \textbf{mR@50}  & \textbf{mR@100} \\ \hline
   \multirow{16}{*}{0.5} & STTran++~\cite{peddi_et_al_scene_sayer_2024}& -  & 30.90  & 34.20  & 34.20  & 34.20  & 17.80  & 20.90  & 21.00  & 21.00  \\ 
  &  \quad+\textbf{\methodname(Ours)}& 70  & \cellcolor{highlightColor}\textbf{41.70}  (+34.95\%) & \cellcolor{highlightColor}\textbf{45.00}  (+31.58\%) & \cellcolor{highlightColor}\textbf{45.10}  (+31.87\%) & \cellcolor{highlightColor}\textbf{45.10}  (+31.87\%) & \cellcolor{highlightColor}\textbf{21.30}  (+19.66\%) & \cellcolor{highlightColor}\textbf{24.80}  (+18.66\%) & \cellcolor{highlightColor}\textbf{24.80}  (+18.10\%) & \cellcolor{highlightColor}\textbf{24.80}  (+18.10\%) \\ 
  &  \quad+\textbf{\methodname(Ours)}& 40  & 36.70  (+18.77\%) & 40.00  (+16.96\%) & 40.00  (+16.96\%) & 40.00  (+16.96\%) & 19.10  (+7.30\%) & 22.10  (+5.74\%) & 22.10  (+5.24\%) & 22.10  (+5.24\%) \\ 
  &  \quad+\textbf{\methodname(Ours)}& 10  & 37.50  (+21.36\%) & 40.20  (+17.54\%) & 40.20  (+17.54\%) & 40.20  (+17.54\%) & 19.90  (+11.80\%) & 22.70  (+8.61\%) & 22.80  (+8.57\%) & 22.80  (+8.57\%) \\ 
    \cmidrule(lr){2-11}
  &  DSGDetr++~\cite{peddi_et_al_scene_sayer_2024}& -  & 41.20  & 44.60  & 44.70  & 44.70  & 17.10  & 20.00  & 20.00  & 20.00  \\ 
  &  \quad+\textbf{\methodname(Ours)}& 70  & \cellcolor{highlightColor}\textbf{41.90}  (+1.70\%) & \cellcolor{highlightColor}\textbf{45.40}  (+1.79\%) & \cellcolor{highlightColor}\textbf{45.40}  (+1.57\%) & \cellcolor{highlightColor}\textbf{45.40}  (+1.57\%) & 19.90  (+16.37\%) & 23.10  (+15.50\%) & 23.10  (+15.50\%) & 23.10  (+15.50\%) \\
\hline
    \end{tabular}
    }
\end{table*}

\begin{table*}[!h]
    \centering
    \captionsetup{font=small}
    \caption{\textbf{GAGS}-With Constraint-0.7 results for SGA.}
    \label{tab:sup_sgg_with_constraint_predcls_70}
    \renewcommand{\arraystretch}{1.2} 
    \resizebox{\textwidth}{!}{
    \begin{tabular}{l|l|l|cccc|cccc}
    \hline
        \multirow{2}{*}{$\mathcal{F}$} & \multirow{2}{*}{Method} & \multirow{2}{*}{$\mathcal{S}$} &  \multicolumn{8}{c}{\textbf{WITH CONSTRAINT}} \\ 
        \cmidrule(lr){4-7} \cmidrule(lr){8-11} 
         & & & \textbf{R@10} & \textbf{R@20} & \textbf{R@50} & \textbf{R@100} &\textbf{mR@10} & \textbf{mR@20} & \textbf{mR@50}  & \textbf{mR@100} \\ \hline
   \multirow{16}{*}{0.7} & STTran++~\cite{peddi_et_al_scene_sayer_2024}& -  & 37.30  & 40.50  & 40.50  & 40.50  & 21.90  & 25.00  & 25.00  & 25.00  \\ 
  &  \quad+\textbf{\methodname(Ours)}& 70  & \cellcolor{highlightColor}\textbf{49.50}  (+32.71\%) & \cellcolor{highlightColor}\textbf{52.70}  (+30.12\%) & \cellcolor{highlightColor}\textbf{52.70}  (+30.12\%) & \cellcolor{highlightColor}\textbf{52.70}  (+30.12\%) & \cellcolor{highlightColor}\textbf{26.20}  (+19.63\%) & \cellcolor{highlightColor}\textbf{30.10}  (+20.40\%) & \cellcolor{highlightColor}\textbf{30.10}  (+20.40\%) & \cellcolor{highlightColor}\textbf{30.10}  (+20.40\%) \\ 
  &  \quad+\textbf{\methodname(Ours)}& 40  & 44.80  (+20.11\%) & 47.70  (+17.78\%) & 47.70  (+17.78\%) & 47.70  (+17.78\%) & 24.00  (+9.59\%) & 26.90  (+7.60\%) & 26.90  (+7.60\%) & 26.90  (+7.60\%) \\ 
  &  \quad+\textbf{\methodname(Ours)}& 10  & 43.70  (+17.16\%) & 46.20  (+14.07\%) & 46.20  (+14.07\%) & 46.20  (+14.07\%) & 25.80  (+17.81\%) & 29.10  (+16.40\%) & 29.10  (+16.40\%) & 29.10  (+16.40\%) \\ 
    \cmidrule(lr){2-11}
  &  DSGDetr++~\cite{peddi_et_al_scene_sayer_2024}& -  & 48.50  & 51.80  & 51.80  & 51.80  & 20.80  & 23.80  & 23.80  & 23.80  \\ 
  &  \quad+\textbf{\methodname(Ours)}& 70  & \cellcolor{highlightColor}\textbf{49.80}  (+2.68\%) & \cellcolor{highlightColor}\textbf{52.90}  (+2.12\%) & \cellcolor{highlightColor}\textbf{53.00}  (+2.32\%) & \cellcolor{highlightColor}\textbf{53.00}  (+2.32\%) & 25.30  (+21.63\%) & 28.70  (+20.59\%) & 28.70  (+20.59\%) & 28.70  (+20.59\%) \\
\hline
    \end{tabular}
    }
\end{table*}

\begin{table*}[!h]
    \centering
    \captionsetup{font=small}
    \caption{\textbf{GAGS}-With Constraint-0.9 results for SGA.}
    \label{tab:sup_sgg_with_constraint_predcls_90}
    \renewcommand{\arraystretch}{1.2} 
    \resizebox{\textwidth}{!}{
    \begin{tabular}{l|l|l|cccc|cccc}
    \hline
        \multirow{2}{*}{$\mathcal{F}$} & \multirow{2}{*}{Method} & \multirow{2}{*}{$\mathcal{S}$} &  \multicolumn{8}{c}{\textbf{WITH CONSTRAINT}} \\ 
        \cmidrule(lr){4-7} \cmidrule(lr){8-11} 
         & & & \textbf{R@10} & \textbf{R@20} & \textbf{R@50} & \textbf{R@100} &\textbf{mR@10} & \textbf{mR@20} & \textbf{mR@50}  & \textbf{mR@100} \\ \hline
   \multirow{16}{*}{0.9} & STTran++~\cite{peddi_et_al_scene_sayer_2024}& -  & 45.50  & 47.80  & 47.80  & 47.80  & 28.80  & 31.30  & 31.30  & 31.30  \\ 
  &  \quad+\textbf{\methodname(Ours)}& 70  & \cellcolor{highlightColor}\textbf{59.00}  (+29.67\%) & \cellcolor{highlightColor}\textbf{61.30}  (+28.24\%) & \cellcolor{highlightColor}\textbf{61.30}  (+28.24\%) & \cellcolor{highlightColor}\textbf{61.30}  (+28.24\%) & 34.20  (+18.75\%) & 37.20  (+18.85\%) & 37.30  (+19.17\%) & 37.30  (+19.17\%) \\ 
  &  \quad+\textbf{\methodname(Ours)}& 40  & 53.90  (+18.46\%) & 56.00  (+17.15\%) & 56.00  (+17.15\%) & 56.00  (+17.15\%) & 31.90  (+10.76\%) & 34.40  (+9.90\%) & 34.40  (+9.90\%) & 34.40  (+9.90\%) \\ 
  &  \quad+\textbf{\methodname(Ours)}& 10  & 50.90  (+11.87\%) & 52.70  (+10.25\%) & 52.70  (+10.25\%) & 52.70  (+10.25\%) & \cellcolor{highlightColor}\textbf{36.70}  (+27.43\%) & \cellcolor{highlightColor}\textbf{40.30}  (+28.75\%) & \cellcolor{highlightColor}\textbf{40.30}  (+28.75\%) & \cellcolor{highlightColor}\textbf{40.30}  (+28.75\%) \\ 
    \cmidrule(lr){2-11}
  &  DSGDetr++~\cite{peddi_et_al_scene_sayer_2024}& -  & 57.70  & 60.00  & 60.00  & 60.00  & 27.00  & 29.40  & 29.40  & 29.40  \\ 
  &  \quad+\textbf{\methodname(Ours)}& 70  & \cellcolor{highlightColor}\textbf{59.20}  (+2.60\%) & \cellcolor{highlightColor}\textbf{61.40}  (+2.33\%) & \cellcolor{highlightColor}\textbf{61.40}  (+2.33\%) & \cellcolor{highlightColor}\textbf{61.40}  (+2.33\%) & 34.20  (+26.67\%) & 36.90  (+25.51\%) & 36.90  (+25.51\%) & 36.90  (+25.51\%) \\
\hline
    \end{tabular}
    }
\end{table*}

\clearpage

\subsubsection{Results - With Constraint Setting - Partially Grounded Action Genome Scenes (PGAGS)}

\begin{table*}[!h]
    \centering
    \captionsetup{font=small}
    \caption{\textbf{PGAGS}-With Constraint-0.3 results for SGA.}
    \label{tab:sup_sgg_with_constraint_sgcls_30}
    \renewcommand{\arraystretch}{1.2} 
    \resizebox{\textwidth}{!}{
    \begin{tabular}{l|l|l|cccc|cccc}
    \hline
        \multirow{2}{*}{$\mathcal{F}$} & \multirow{2}{*}{Method} & \multirow{2}{*}{$\mathcal{S}$} &  \multicolumn{8}{c}{\textbf{WITH CONSTRAINT}} \\ 
        \cmidrule(lr){4-7} \cmidrule(lr){8-11} 
         & & & \textbf{R@10} & \textbf{R@20} & \textbf{R@50} & \textbf{R@100} &\textbf{mR@10} & \textbf{mR@20} & \textbf{mR@50}  & \textbf{mR@100} \\ \hline
   \multirow{16}{*}{0.3} & STTran++~\cite{peddi_et_al_scene_sayer_2024}& -  & 20.30  & 21.40  & 21.40  & 21.40  & 12.60  & 13.60  & 13.60  & 13.60  \\ 
  &  \quad+\textbf{\methodname(Ours)}& 70  & 26.70  (+31.53\%) & 27.70  (+29.44\%) & 27.70  (+29.44\%) & 27.70  (+29.44\%) & 12.80  (+1.59\%) & 13.60  & 13.60  & 13.60  \\ 
  &  \quad+\textbf{\methodname(Ours)}& 40  & \cellcolor{highlightColor}\textbf{27.60}  (+35.96\%) & \cellcolor{highlightColor}\textbf{28.70}  (+34.11\%) & \cellcolor{highlightColor}\textbf{28.70}  (+34.11\%) & \cellcolor{highlightColor}\textbf{28.70}  (+34.11\%) & 15.10  (+19.84\%) & 16.30  (+19.85\%) & 16.30  (+19.85\%) & 16.30  (+19.85\%) \\ 
  &  \quad+\textbf{\methodname(Ours)}& 10  & 21.80  (+7.39\%) & 22.40  (+4.67\%) & 22.40  (+4.67\%) & 22.40  (+4.67\%) & \cellcolor{highlightColor}\textbf{17.60}  (+39.68\%) & \cellcolor{highlightColor}\textbf{18.90}  (+38.97\%) & \cellcolor{highlightColor}\textbf{18.90}  (+38.97\%) & \cellcolor{highlightColor}\textbf{18.90}  (+38.97\%) \\ 
    \cmidrule(lr){2-11}
  &  DSGDetr++~\cite{peddi_et_al_scene_sayer_2024}& -  & \cellcolor{highlightColor}\textbf{28.10}  & \cellcolor{highlightColor}\textbf{29.30}  & \cellcolor{highlightColor}\textbf{29.30}  & \cellcolor{highlightColor}\textbf{29.30}  & 13.00  & 14.00  & 14.00  & 14.00  \\ 
  &  \quad+\textbf{\methodname(Ours)}& 70  & 26.50  (-5.69\%) & 27.60  (-5.80\%) & 27.60  (-5.80\%) & 27.60  (-5.80\%) & 14.70  (+13.08\%) & 15.80  (+12.86\%) & 15.80  (+12.86\%) & 15.80  (+12.86\%) \\
\hline
    \end{tabular}
    }
\end{table*}

\begin{table*}[!h]
    \centering
    \captionsetup{font=small}
    \caption{\textbf{PGAGS}-With Constraint-0.5 results for SGA.}
    \label{tab:sup_sgg_with_constraint_sgcls_50}
    \renewcommand{\arraystretch}{1.2} 
    \resizebox{\textwidth}{!}{
    \begin{tabular}{l|l|l|cccc|cccc}
    \hline
        \multirow{2}{*}{$\mathcal{F}$} & \multirow{2}{*}{Method} & \multirow{2}{*}{$\mathcal{S}$} &  \multicolumn{8}{c}{\textbf{WITH CONSTRAINT}} \\ 
        \cmidrule(lr){4-7} \cmidrule(lr){8-11} 
         & & & \textbf{R@10} & \textbf{R@20} & \textbf{R@50} & \textbf{R@100} &\textbf{mR@10} & \textbf{mR@20} & \textbf{mR@50}  & \textbf{mR@100} \\ \hline
   \multirow{16}{*}{0.5} & STTran++~\cite{peddi_et_al_scene_sayer_2024}& -  & 22.50  & 23.80  & 23.80  & 23.80  & 14.30  & 15.80  & 15.80  & 15.80  \\ 
  &  \quad+\textbf{\methodname(Ours)}& 70  & 29.50  (+31.11\%) & 30.60  (+28.57\%) & 30.60  (+28.57\%) & 30.60  (+28.57\%) & 14.50  (+1.40\%) & 15.70  (-0.63\%) & 15.70  (-0.63\%) & 15.70  (-0.63\%) \\ 
  &  \quad+\textbf{\methodname(Ours)}& 40  & \cellcolor{highlightColor}\textbf{30.20}  (+34.22\%) & \cellcolor{highlightColor}\textbf{31.40}  (+31.93\%) & \cellcolor{highlightColor}\textbf{31.40}  (+31.93\%) & \cellcolor{highlightColor}\textbf{31.40}  (+31.93\%) & 16.90  (+18.18\%) & 18.50  (+17.09\%) & 18.50  (+17.09\%) & 18.50  (+17.09\%) \\ 
  &  \quad+\textbf{\methodname(Ours)}& 10  & 24.10  (+7.11\%) & 25.00  (+5.04\%) & 25.00  (+5.04\%) & 25.00  (+5.04\%) & \cellcolor{highlightColor}\textbf{20.20}  (+41.26\%) & \cellcolor{highlightColor}\textbf{22.00}  (+39.24\%) & \cellcolor{highlightColor}\textbf{22.00}  (+39.24\%) & \cellcolor{highlightColor}\textbf{22.00}  (+39.24\%) \\ 
    \cmidrule(lr){2-11}
  &  DSGDetr++~\cite{peddi_et_al_scene_sayer_2024}& -  & \cellcolor{highlightColor}\textbf{31.70}  & \cellcolor{highlightColor}\textbf{33.00}  & \cellcolor{highlightColor}\textbf{33.00}  & \cellcolor{highlightColor}\textbf{33.00}  & 15.00  & 16.30  & 16.30  & 16.30  \\ 
  &  \quad+\textbf{\methodname(Ours)}& 70  & 28.70  (-9.46\%) & 29.90  (-9.39\%) & 29.90  (-9.39\%) & 29.90  (-9.39\%) & 16.80  (+12.00\%) & 18.30  (+12.27\%) & 18.30  (+12.27\%) & 18.30  (+12.27\%) \\
\hline
    \end{tabular}
    }
\end{table*}

\begin{table*}[!h]
    \centering
    \captionsetup{font=small}
    \caption{\textbf{PGAGS}-With Constraint-0.7 results for SGA.}
    \label{tab:sup_sgg_with_constraint_sgcls_70}
    \renewcommand{\arraystretch}{1.2} 
    \resizebox{\textwidth}{!}{
    \begin{tabular}{l|l|l|cccc|cccc}
    \hline
        \multirow{2}{*}{$\mathcal{F}$} & \multirow{2}{*}{Method} & \multirow{2}{*}{$\mathcal{S}$} &  \multicolumn{8}{c}{\textbf{WITH CONSTRAINT}} \\ 
        \cmidrule(lr){4-7} \cmidrule(lr){8-11} 
         & & & \textbf{R@10} & \textbf{R@20} & \textbf{R@50} & \textbf{R@100} &\textbf{mR@10} & \textbf{mR@20} & \textbf{mR@50}  & \textbf{mR@100} \\ \hline
   \multirow{16}{*}{0.7} & STTran++~\cite{peddi_et_al_scene_sayer_2024}& -  & 27.10  & 28.20  & 28.20  & 28.20  & 17.20  & 18.60  & 18.60  & 18.60  \\ 
  &  \quad+\textbf{\methodname(Ours)}& 70  & \cellcolor{highlightColor}\textbf{36.10}  (+33.21\%) & \cellcolor{highlightColor}\textbf{37.10}  (+31.56\%) & \cellcolor{highlightColor}\textbf{37.10}  (+31.56\%) & \cellcolor{highlightColor}\textbf{37.10}  (+31.56\%) & 18.70  (+8.72\%) & 19.90  (+6.99\%) & 19.90  (+6.99\%) & 19.90  (+6.99\%) \\ 
  &  \quad+\textbf{\methodname(Ours)}& 40  & 36.00  (+32.84\%) & 37.00  (+31.21\%) & 37.00  (+31.21\%) & 37.00  (+31.21\%) & 21.00  (+22.09\%) & 22.70  (+22.04\%) & 22.70  (+22.04\%) & 22.70  (+22.04\%) \\ 
  &  \quad+\textbf{\methodname(Ours)}& 10  & 29.60  (+9.23\%) & 30.20  (+7.09\%) & 30.20  (+7.09\%) & 30.20  (+7.09\%) & \cellcolor{highlightColor}\textbf{25.60}  (+48.84\%) & \cellcolor{highlightColor}\textbf{27.40}  (+47.31\%) & \cellcolor{highlightColor}\textbf{27.40}  (+47.31\%) & \cellcolor{highlightColor}\textbf{27.40}  (+47.31\%) \\ 
    \cmidrule(lr){2-11}
  &  DSGDetr++~\cite{peddi_et_al_scene_sayer_2024}& -  & \cellcolor{highlightColor}\textbf{37.30}  & \cellcolor{highlightColor}\textbf{38.50}  & \cellcolor{highlightColor}\textbf{38.50}  & \cellcolor{highlightColor}\textbf{38.50}  & 18.10  & 19.40  & 19.40  & 19.40  \\ 
  &  \quad+\textbf{\methodname(Ours)}& 70  & 34.30  (-8.04\%) & 35.40  (-8.05\%) & 35.40  (-8.05\%) & 35.40  (-8.05\%) & 20.70  (+14.36\%) & 22.10  (+13.92\%) & 22.10  (+13.92\%) & 22.10  (+13.92\%) \\
\hline
    \end{tabular}
    }
\end{table*}

\begin{table*}[!h]
    \centering
    \captionsetup{font=small}
    \caption{\textbf{PGAGS}-With Constraint-0.9 results for SGA.}
    \label{tab:sup_sgg_with_constraint_sgcls_90}
    \renewcommand{\arraystretch}{1.2} 
    \resizebox{\textwidth}{!}{
    \begin{tabular}{l|l|l|cccc|cccc}
    \hline
        \multirow{2}{*}{$\mathcal{F}$} & \multirow{2}{*}{Method} & \multirow{2}{*}{$\mathcal{S}$} &  \multicolumn{8}{c}{\textbf{WITH CONSTRAINT}} \\ 
        \cmidrule(lr){4-7} \cmidrule(lr){8-11} 
         & & & \textbf{R@10} & \textbf{R@20} & \textbf{R@50} & \textbf{R@100} &\textbf{mR@10} & \textbf{mR@20} & \textbf{mR@50}  & \textbf{mR@100} \\ \hline
   \multirow{16}{*}{0.9} & STTran++~\cite{peddi_et_al_scene_sayer_2024}& -  & 31.00  & 31.80  & 31.80  & 31.80  & 21.20  & 22.50  & 22.50  & 22.50  \\ 
  &  \quad+\textbf{\methodname(Ours)}& 70  & \cellcolor{highlightColor}\textbf{42.00}  (+35.48\%) & \cellcolor{highlightColor}\textbf{42.80}  (+34.59\%) & \cellcolor{highlightColor}\textbf{42.80}  (+34.59\%) & \cellcolor{highlightColor}\textbf{42.80}  (+34.59\%) & 26.00  (+22.64\%) & 27.20  (+20.89\%) & 27.20  (+20.89\%) & 27.20  (+20.89\%) \\ 
  &  \quad+\textbf{\methodname(Ours)}& 40  & 39.90  (+28.71\%) & 40.60  (+27.67\%) & 40.60  (+27.67\%) & 40.60  (+27.67\%) & 26.60  (+25.47\%) & 28.10  (+24.89\%) & 28.10  (+24.89\%) & 28.10  (+24.89\%) \\ 
  &  \quad+\textbf{\methodname(Ours)}& 10  & 32.70  (+5.48\%) & 33.20  (+4.40\%) & 33.20  (+4.40\%) & 33.20  (+4.40\%) & \cellcolor{highlightColor}\textbf{32.60}  (+53.77\%) & \cellcolor{highlightColor}\textbf{34.20}  (+52.00\%) & \cellcolor{highlightColor}\textbf{34.20}  (+52.00\%) & \cellcolor{highlightColor}\textbf{34.20}  (+52.00\%) \\ 
    \cmidrule(lr){2-11}
  &  DSGDetr++~\cite{peddi_et_al_scene_sayer_2024}& -  & \cellcolor{highlightColor}\textbf{43.10}  & \cellcolor{highlightColor}\textbf{44.00}  & \cellcolor{highlightColor}\textbf{44.00}  & \cellcolor{highlightColor}\textbf{44.00}  & 22.20  & 23.40  & 23.40  & 23.40  \\ 
  &  \quad+\textbf{\methodname(Ours)}& 70  & 40.00  (-7.19\%) & 40.80  (-7.27\%) & 40.80  (-7.27\%) & 40.80  (-7.27\%) & 24.90  (+12.16\%) & 26.20  (+11.97\%) & 26.20  (+11.97\%) & 26.20  (+11.97\%) \\
\hline
    \end{tabular}
    }
\end{table*}

\clearpage

\subsubsection{Results - With Constraint Setting - Action Genome Scenes (AGS)}

\begin{table*}[!h]
    \centering
    \captionsetup{font=small}
    \caption{\textbf{AGS}-With Constraint-0.3 results for SGA.}
    \label{tab:sup_sgg_with_constraint_sgdet_30}
    \renewcommand{\arraystretch}{1.2} 
    \resizebox{\textwidth}{!}{
    \begin{tabular}{l|l|l|cccc|cccc}
    \hline
        \multirow{2}{*}{$\mathcal{F}$} & \multirow{2}{*}{Method} & \multirow{2}{*}{$\mathcal{S}$} &  \multicolumn{8}{c}{\textbf{WITH CONSTRAINT}} \\ 
        \cmidrule(lr){4-7} \cmidrule(lr){8-11} 
         & & & \textbf{R@10} & \textbf{R@20} & \textbf{R@50} & \textbf{R@100} &\textbf{mR@10} & \textbf{mR@20} & \textbf{mR@50}  & \textbf{mR@100} \\ \hline
   \multirow{16}{*}{0.3} & STTran++~\cite{peddi_et_al_scene_sayer_2024}& -  & 19.40  & \cellcolor{highlightColor}\textbf{31.10}  & \cellcolor{highlightColor}\textbf{33.50}  & \cellcolor{highlightColor}\textbf{33.50}  & 7.60  & 15.30  & 16.90  & 16.90  \\ 
  &  \quad+\textbf{\methodname(Ours)}& 70  & \cellcolor{highlightColor}\textbf{19.70}  (+1.55\%) & 30.20  (-2.89\%) & 32.40  (-3.28\%) & 32.40  (-3.28\%) & 7.80  (+2.63\%) & 15.10  (-1.31\%) & 16.80  (-0.59\%) & 16.80  (-0.59\%) \\ 
  &  \quad+\textbf{\methodname(Ours)}& 40  & 19.20  (-1.03\%) & 30.30  (-2.57\%) & 32.60  (-2.69\%) & 32.60  (-2.69\%) & 8.40  (+10.53\%) & 16.80  (+9.80\%) & 18.60  (+10.06\%) & 18.60  (+10.06\%) \\ 
  &  \quad+\textbf{\methodname(Ours)}& 10  & 18.30  (-5.67\%) & 27.40  (-11.90\%) & 29.30  (-12.54\%) & 29.30  (-12.54\%) & \cellcolor{highlightColor}\textbf{8.90}  (+17.11\%) & \cellcolor{highlightColor}\textbf{17.90}  (+16.99\%) & \cellcolor{highlightColor}\textbf{19.70}  (+16.57\%) & \cellcolor{highlightColor}\textbf{19.70}  (+16.57\%) \\ 
    \cmidrule(lr){2-11}
  &  DSGDetr++~\cite{peddi_et_al_scene_sayer_2024}& -  & \cellcolor{highlightColor}\textbf{18.80}  & \cellcolor{highlightColor}\textbf{28.30}  & \cellcolor{highlightColor}\textbf{29.90}  & \cellcolor{highlightColor}\textbf{29.90}  & 7.10  & 12.80  & 13.80  & 13.80  \\ 
  &  \quad+\textbf{\methodname(Ours)}& 70  & 18.60  (-1.06\%) & 27.40  (-3.18\%) & 29.10  (-2.68\%) & 29.10  (-2.68\%) & 8.20  (+15.49\%) & 15.20  (+18.75\%) & 16.60  (+20.29\%) & 16.60  (+20.29\%) \\
\hline
    \end{tabular}
    }
\end{table*}

\begin{table*}[!h]
    \centering
    \captionsetup{font=small}
    \caption{\textbf{AGS}-With Constraint-0.5 results for SGA.}
    \label{tab:sup_sgg_with_constraint_sgdet_50}
    \renewcommand{\arraystretch}{1.2} 
    \resizebox{\textwidth}{!}{
    \begin{tabular}{l|l|l|cccc|cccc}
    \hline
        \multirow{2}{*}{$\mathcal{F}$} & \multirow{2}{*}{Method} & \multirow{2}{*}{$\mathcal{S}$} &  \multicolumn{8}{c}{\textbf{WITH CONSTRAINT}} \\ 
        \cmidrule(lr){4-7} \cmidrule(lr){8-11} 
         & & & \textbf{R@10} & \textbf{R@20} & \textbf{R@50} & \textbf{R@100} &\textbf{mR@10} & \textbf{mR@20} & \textbf{mR@50}  & \textbf{mR@100} \\ \hline
   \multirow{16}{*}{0.5} & STTran++~\cite{peddi_et_al_scene_sayer_2024}& -  & 20.40  & \cellcolor{highlightColor}\textbf{33.30}  & \cellcolor{highlightColor}\textbf{35.90}  & \cellcolor{highlightColor}\textbf{35.90}  & 7.90  & 16.40  & 18.40  & 18.40  \\ 
  &  \quad+\textbf{\methodname(Ours)}& 70  & \cellcolor{highlightColor}\textbf{21.10}  (+3.43\%) & 32.70  (-1.80\%) & 34.90  (-2.79\%) & 34.90  (-2.79\%) & 8.30  (+5.06\%) & 16.50  (+0.61\%) & 18.30  (-0.54\%) & 18.30  (-0.54\%) \\ 
  &  \quad+\textbf{\methodname(Ours)}& 40  & 20.60  (+0.98\%) & 32.50  (-2.40\%) & 35.00  (-2.51\%) & 35.00  (-2.51\%) & 8.90  (+12.66\%) & 17.80  (+8.54\%) & 19.80  (+7.61\%) & 19.80  (+7.61\%) \\ 
  &  \quad+\textbf{\methodname(Ours)}& 10  & 19.80  (-2.94\%) & 29.70  (-10.81\%) & 31.80  (-11.42\%) & 31.80  (-11.42\%) & \cellcolor{highlightColor}\textbf{9.30}  (+17.72\%) & \cellcolor{highlightColor}\textbf{18.70}  (+14.02\%) & \cellcolor{highlightColor}\textbf{20.90}  (+13.59\%) & \cellcolor{highlightColor}\textbf{20.90}  (+13.59\%) \\ 
    \cmidrule(lr){2-11}
  &  DSGDetr++~\cite{peddi_et_al_scene_sayer_2024}& -  & 19.80  & \cellcolor{highlightColor}\textbf{30.00}  & \cellcolor{highlightColor}\textbf{31.90}  & \cellcolor{highlightColor}\textbf{31.90}  & 7.40  & 13.40  & 14.60  & 14.60  \\ 
  &  \quad+\textbf{\methodname(Ours)}& 70  & \cellcolor{highlightColor}\textbf{20.10}  (+1.52\%) & 29.60  (-1.33\%) & 31.50  (-1.25\%) & 31.50  (-1.25\%) & 8.70  (+17.57\%) & 16.40  (+22.39\%) & 17.90  (+22.60\%) & 17.90  (+22.60\%) \\
\hline
    \end{tabular}
    }
\end{table*}

\begin{table*}[!h]
    \centering
    \captionsetup{font=small}
    \caption{\textbf{AGS}-With Constraint-0.7 results for SGA.}
    \label{tab:sup_sgg_with_constraint_sgdet_70}
    \renewcommand{\arraystretch}{1.2} 
    \resizebox{\textwidth}{!}{
    \begin{tabular}{l|l|l|cccc|cccc}
    \hline
        \multirow{2}{*}{$\mathcal{F}$} & \multirow{2}{*}{Method} & \multirow{2}{*}{$\mathcal{S}$} &  \multicolumn{8}{c}{\textbf{WITH CONSTRAINT}} \\ 
        \cmidrule(lr){4-7} \cmidrule(lr){8-11} 
         & & & \textbf{R@10} & \textbf{R@20} & \textbf{R@50} & \textbf{R@100} &\textbf{mR@10} & \textbf{mR@20} & \textbf{mR@50}  & \textbf{mR@100} \\ \hline
   \multirow{16}{*}{0.7} & STTran++~\cite{peddi_et_al_scene_sayer_2024}& -  & 22.50  & 36.40  & \cellcolor{highlightColor}\textbf{39.20}  & \cellcolor{highlightColor}\textbf{39.20}  & 9.10  & 18.20  & 20.20  & 20.20  \\ 
  &  \quad+\textbf{\methodname(Ours)}& 70  & \cellcolor{highlightColor}\textbf{23.50}  (+4.44\%) & \cellcolor{highlightColor}\textbf{36.50}  (+0.27\%) & 39.10  (-0.26\%) & 39.10  (-0.26\%) & 9.90  (+8.79\%) & 19.30  (+6.04\%) & 21.50  (+6.44\%) & 21.50  (+6.44\%) \\ 
  &  \quad+\textbf{\methodname(Ours)}& 40  & 23.00  (+2.22\%) & 36.00  (-1.10\%) & 38.70  (-1.28\%) & 38.70  (-1.28\%) & 10.40  (+14.29\%) & 20.50  (+12.64\%) & 22.80  (+12.87\%) & 22.80  (+12.87\%) \\ 
  &  \quad+\textbf{\methodname(Ours)}& 10  & 21.90  (-2.67\%) & 32.70  (-10.16\%) & 34.90  (-10.97\%) & 34.90  (-10.97\%) & \cellcolor{highlightColor}\textbf{10.90}  (+19.78\%) & \cellcolor{highlightColor}\textbf{21.90}  (+20.33\%) & \cellcolor{highlightColor}\textbf{24.10}  (+19.31\%) & \cellcolor{highlightColor}\textbf{24.10}  (+19.31\%) \\ 
    \cmidrule(lr){2-11}
  &  DSGDetr++~\cite{peddi_et_al_scene_sayer_2024}& -  & 22.20  & \cellcolor{highlightColor}\textbf{33.20}  & \cellcolor{highlightColor}\textbf{35.10}  & \cellcolor{highlightColor}\textbf{35.10}  & 8.40  & 14.80  & 16.00  & 16.00  \\ 
  &  \quad+\textbf{\methodname(Ours)}& 70  & \cellcolor{highlightColor}\textbf{22.40}  (+0.90\%) & 32.60  (-1.81\%) & 34.50  (-1.71\%) & 34.50  (-1.71\%) & 10.10  (+20.24\%) & 18.40  (+24.32\%) & 20.10  (+25.63\%) & 20.10  (+25.63\%) \\
\hline
    \end{tabular}
    }
\end{table*}

\begin{table*}[!h]
    \centering
    \captionsetup{font=small}
    \caption{\textbf{AGS}-With Constraint-0.9 results for SGA.}
    \label{tab:sup_sgg_with_constraint_sgdet_90}
    \renewcommand{\arraystretch}{1.2} 
    \resizebox{\textwidth}{!}{
    \begin{tabular}{l|l|l|cccc|cccc}
    \hline
        \multirow{2}{*}{$\mathcal{F}$} & \multirow{2}{*}{Method} & \multirow{2}{*}{$\mathcal{S}$} &  \multicolumn{8}{c}{\textbf{WITH CONSTRAINT}} \\ 
        \cmidrule(lr){4-7} \cmidrule(lr){8-11} 
         & & & \textbf{R@10} & \textbf{R@20} & \textbf{R@50} & \textbf{R@100} &\textbf{mR@10} & \textbf{mR@20} & \textbf{mR@50}  & \textbf{mR@100} \\ \hline
   \multirow{16}{*}{0.9} & STTran++~\cite{peddi_et_al_scene_sayer_2024}& -  & 24.60  & 39.80  & 42.90  & 42.90  & 9.80  & 20.90  & 23.50  & 23.50  \\ 
  &  \quad+\textbf{\methodname(Ours)}& 70  & \cellcolor{highlightColor}\textbf{25.80}  (+4.88\%) & \cellcolor{highlightColor}\textbf{40.50}  (+1.76\%) & \cellcolor{highlightColor}\textbf{43.50}  (+1.40\%) & \cellcolor{highlightColor}\textbf{43.50}  (+1.40\%) & 11.10  (+13.27\%) & 22.70  (+8.61\%) & 25.40  (+8.09\%) & 25.40  (+8.09\%) \\ 
  &  \quad+\textbf{\methodname(Ours)}& 40  & 24.70  (+0.41\%) & 39.50  (-0.75\%) & 42.50  (-0.93\%) & 42.50  (-0.93\%) & 12.30  (+25.51\%) & 25.50  (+22.01\%) & 28.20  (+20.00\%) & 28.20  (+20.00\%) \\ 
  &  \quad+\textbf{\methodname(Ours)}& 10  & 23.60  (-4.07\%) & 35.60  (-10.55\%) & 37.90  (-11.66\%) & 37.90  (-11.66\%) & \cellcolor{highlightColor}\textbf{12.60}  (+28.57\%) & \cellcolor{highlightColor}\textbf{27.10}  (+29.67\%) & \cellcolor{highlightColor}\textbf{29.90}  (+27.23\%) & \cellcolor{highlightColor}\textbf{29.90}  (+27.23\%) \\ 
    \cmidrule(lr){2-11}
  &  DSGDetr++~\cite{peddi_et_al_scene_sayer_2024}& -  & \cellcolor{highlightColor}\textbf{24.80}  & \cellcolor{highlightColor}\textbf{37.50}  & \cellcolor{highlightColor}\textbf{39.70}  & \cellcolor{highlightColor}\textbf{39.70}  & 9.50  & 17.70  & 19.20  & 19.20  \\ 
  &  \quad+\textbf{\methodname(Ours)}& 70  & 24.70  (-0.40\%) & 37.10  (-1.07\%) & 39.50  (-0.50\%) & 39.50  (-0.50\%) & 11.90  (+25.26\%) & 22.50  (+27.12\%) & 24.90  (+29.69\%) & 24.90  (+29.69\%) \\
\hline
    \end{tabular}
    }
\end{table*}

\clearpage

\subsection{Robust Video Scene Graph Generation}

\subsubsection{Findings}
Table~\ref{tab:sup_sgg_corruptions_3}, Table~\ref{tab:sup_sgg_corruptions_5} present the Robustness Evaluation Results for SGCLS and PREDCLS for Scene Graph Generation (SGG) under various corruption scenarios. These experiments assess how well models, with and without \methodname, handle different levels of data corruption. The settings include 15 corruption types and three graph-building strategies (With Cosntraint, No Constraint, Semi Constraint). Results highlight the impact of \methodname in improving robustness across these scenarios. \methodname performs best against Fog, Brightness, Saturate and moderate gains under Defocus blur, Gaussian Blur. \methodname shows an average of 25\% gains for With Constraint mR@50, 10\% gains for No Constraint mR@50.

\subsubsection{Results}

\begin{table*}[!h]
    \centering
    \captionsetup{font=small}
    \caption{Robustness Evaluation Results for SGG.}
    \label{tab:sup_sgg_corruptions_3}
    \renewcommand{\arraystretch}{1.2} 
    \resizebox{\textwidth}{!}{
    \begin{tabular}{l|l|l|l|ccc|cccccc|ccc}
    \hline
      \multirow{2}{*}{Severity} & \multirow{2}{*}{Mode} & \multirow{2}{*}{Corruption} & \multirow{2}{*}{Method} & \multicolumn{3}{c}{\textbf{With Constraint}} & \multicolumn{6}{c}{\textbf{No Constraint}} & \multicolumn{3}{c}{\textbf{Semi Constraint}} \\ 
        \cmidrule(lr){5-7} \cmidrule(lr){8-13} \cmidrule(lr){14-16} 
  & & & & \textbf{mR@10} & \textbf{mR@20} & \textbf{mR@50} & \textbf{R@10} & \textbf{R@20} & \textbf{R@50}  & \textbf{mR@10} & \textbf{mR@20} & \textbf{mR@50}  & \textbf{mR@10} & \textbf{mR@20} & \textbf{mR@50}  \\ \hline
   \multirow{32}{*}{3} &      \multirow{32}{*}{sgcls} & \multirow{2}{*}{Gaussian Noise} &         DSGDetr~\cite{Feng_2021} & 9.6 & 10.3 & 10.3 & 20.9 & 25.4 & 26.8 & 15.7 & 19.4 & 23.4 & 11.4 & 15.3 & 15.7  \\ 
    &    & &         \quad+\textbf{\methodname(Ours)} & \cellcolor{highlightColor} \textbf{13.7} (+42.7\%) & \cellcolor{highlightColor} \textbf{14.9} (+44.7\%) & \cellcolor{highlightColor} \textbf{15.0} (+45.6\%) & \cellcolor{highlightColor} \textbf{21.0} (+0.5\%) & \cellcolor{highlightColor} \textbf{27.3} (+7.5\%) & \cellcolor{highlightColor} \textbf{30.1} (+12.3\%) & \cellcolor{highlightColor} \textbf{20.8} (+32.5\%) & \cellcolor{highlightColor} \textbf{25.4} (+30.9\%) & \cellcolor{highlightColor} \textbf{29.1} (+24.4\%) & \cellcolor{highlightColor} \textbf{15.6} (+36.8\%) & \cellcolor{highlightColor} \textbf{21.5} (+40.5\%) & \cellcolor{highlightColor} \textbf{22.2} (+41.4\%)  \\ 
 \cmidrule(lr){4-16}  
     &    &\multirow{2}{*}{Shot Noise} &         DSGDetr~\cite{Feng_2021} & 10.0 & 10.8 & 10.8 & 21.9 & 26.6 & 28.1 & 16.6 & 20.4 & 26.5 & 12.1 & 16.3 & 16.6  \\ 
    &    & &         \quad+\textbf{\methodname(Ours)} & \cellcolor{highlightColor} \textbf{15.0} (+50.0\%) & \cellcolor{highlightColor} \textbf{16.5} (+52.8\%) & \cellcolor{highlightColor} \textbf{16.5} (+52.8\%) & \cellcolor{highlightColor} \textbf{22.6} (+3.2\%) & \cellcolor{highlightColor} \textbf{29.4} (+10.5\%) & \cellcolor{highlightColor} \textbf{32.6} (+16.0\%) & \cellcolor{highlightColor} \textbf{22.4} (+34.9\%) & \cellcolor{highlightColor} \textbf{27.3} (+33.8\%) & \cellcolor{highlightColor} \textbf{31.0} (+17.0\%) & \cellcolor{highlightColor} \textbf{17.2} (+42.1\%) & \cellcolor{highlightColor} \textbf{23.3} (+42.9\%) & \cellcolor{highlightColor} \textbf{24.2} (+45.8\%)  \\ 
 \cmidrule(lr){4-16}  
     &    &\multirow{2}{*}{Impulse Noise} &         DSGDetr~\cite{Feng_2021} & 8.7 & 9.4 & 9.5 & 19.2 & 23.4 & 24.7 & 14.4 & 17.7 & 23.4 & 10.4 & 14.2 & 14.6  \\ 
    &    & &         \quad+\textbf{\methodname(Ours)} & \cellcolor{highlightColor} \textbf{12.4} (+42.5\%) & \cellcolor{highlightColor} \textbf{13.7} (+45.7\%) & \cellcolor{highlightColor} \textbf{13.7} (+44.2\%) & \cellcolor{highlightColor} \textbf{19.4} (+1.0\%) & \cellcolor{highlightColor} \textbf{25.3} (+8.1\%) & \cellcolor{highlightColor} \textbf{27.9} (+13.0\%) & \cellcolor{highlightColor} \textbf{18.5} (+28.5\%) & \cellcolor{highlightColor} \textbf{22.8} (+28.8\%) & \cellcolor{highlightColor} \textbf{26.2} (+12.0\%) & \cellcolor{highlightColor} \textbf{13.9} (+33.7\%) & \cellcolor{highlightColor} \textbf{19.2} (+35.2\%) & \cellcolor{highlightColor} \textbf{20.1} (+37.7\%)  \\ 
 \cmidrule(lr){4-16}  
     &    &\multirow{2}{*}{Speckle Noise} &         DSGDetr~\cite{Feng_2021} & 12.6 & 13.7 & 13.7 & 26.2 & 32.2 & 34.2 & 20.3 & 25.0 & 32.2 & 15.0 & 20.7 & 21.2  \\ 
    &    & &         \quad+\textbf{\methodname(Ours)} & \cellcolor{highlightColor} \textbf{17.6} (+39.7\%) & \cellcolor{highlightColor} \textbf{19.3} (+40.9\%) & \cellcolor{highlightColor} \textbf{19.3} (+40.9\%) & \cellcolor{highlightColor} \textbf{26.4} (+0.8\%) & \cellcolor{highlightColor} \textbf{34.5} (+7.1\%) & \cellcolor{highlightColor} \textbf{38.4} (+12.3\%) & \cellcolor{highlightColor} \textbf{27.1} (+33.5\%) & \cellcolor{highlightColor} \textbf{32.6} (+30.4\%) & \cellcolor{highlightColor} \textbf{37.3} (+15.8\%) & \cellcolor{highlightColor} \textbf{20.4} (+36.0\%) & \cellcolor{highlightColor} \textbf{27.8} (+34.3\%) & \cellcolor{highlightColor} \textbf{28.9} (+36.3\%)  \\ 
 \cmidrule(lr){4-16}  
     &    &\multirow{2}{*}{Gaussian Blur} &         DSGDetr~\cite{Feng_2021} & 21.1 & 22.8 & 22.8 & \cellcolor{highlightColor} \textbf{43.7} & \cellcolor{highlightColor} \textbf{53.3} & 56.5 & 33.4 & 41.0 & 52.3 & 24.6 & 33.2 & 34.0  \\ 
    &    & &         \quad+\textbf{\methodname(Ours)} & \cellcolor{highlightColor} \textbf{26.1} (+23.7\%) & \cellcolor{highlightColor} \textbf{28.7} (+25.9\%) & \cellcolor{highlightColor} \textbf{28.8} (+26.3\%) & 39.4 (-9.8\%) & 52.0 (-2.4\%) & \cellcolor{highlightColor} \textbf{57.8} (+2.3\%) & \cellcolor{highlightColor} \textbf{39.5} (+18.3\%) & \cellcolor{highlightColor} \textbf{48.8} (+19.0\%) & \cellcolor{highlightColor} \textbf{58.2} (+11.3\%) & \cellcolor{highlightColor} \textbf{29.4} (+19.5\%) & \cellcolor{highlightColor} \textbf{40.8} (+22.9\%) & \cellcolor{highlightColor} \textbf{42.5} (+25.0\%)  \\ 
 \cmidrule(lr){4-16}  
     &    &\multirow{2}{*}{Defocus Blur} &         DSGDetr~\cite{Feng_2021} & 20.9 & 22.6 & 22.6 & \cellcolor{highlightColor} \textbf{43.2} & \cellcolor{highlightColor} \textbf{52.8} & 55.9 & 32.7 & 40.4 & 51.7 & 24.3 & 32.7 & 33.5  \\ 
    &    & &         \quad+\textbf{\methodname(Ours)} & \cellcolor{highlightColor} \textbf{25.6} (+22.5\%) & \cellcolor{highlightColor} \textbf{28.1} (+24.3\%) & \cellcolor{highlightColor} \textbf{28.1} (+24.3\%) & 39.0 (-9.7\%) & 51.4 (-2.7\%) & \cellcolor{highlightColor} \textbf{57.3} (+2.5\%) & \cellcolor{highlightColor} \textbf{38.6} (+18.0\%) & \cellcolor{highlightColor} \textbf{47.7} (+18.1\%) & \cellcolor{highlightColor} \textbf{57.0} (+10.3\%) & \cellcolor{highlightColor} \textbf{28.7} (+18.1\%) & \cellcolor{highlightColor} \textbf{40.0} (+22.3\%) & \cellcolor{highlightColor} \textbf{41.7} (+24.5\%)  \\ 
 \cmidrule(lr){4-16}  
     &    &\multirow{2}{*}{Fog} &         DSGDetr~\cite{Feng_2021} & 22.6 & 24.9 & 24.9 & \cellcolor{highlightColor} \textbf{47.8} & \cellcolor{highlightColor} \textbf{58.4} & 62.0 & 35.5 & 43.4 & 54.6 & 26.6 & 36.1 & 37.2  \\ 
    &    & &         \quad+\textbf{\methodname(Ours)} & \cellcolor{highlightColor} \textbf{28.3} (+25.2\%) & \cellcolor{highlightColor} \textbf{31.8} (+27.7\%) & \cellcolor{highlightColor} \textbf{31.9} (+28.1\%) & 42.6 (-10.9\%) & 56.0 (-4.1\%) & \cellcolor{highlightColor} \textbf{62.1} (+0.2\%) & \cellcolor{highlightColor} \textbf{43.8} (+23.4\%) & \cellcolor{highlightColor} \textbf{53.0} (+22.1\%) & \cellcolor{highlightColor} \textbf{61.7} (+13.0\%) & \cellcolor{highlightColor} \textbf{31.8} (+19.5\%) & \cellcolor{highlightColor} \textbf{45.7} (+26.6\%) & \cellcolor{highlightColor} \textbf{48.2} (+29.6\%)  \\ 
 \cmidrule(lr){4-16}  
     &    &\multirow{2}{*}{Frost} &         DSGDetr~\cite{Feng_2021} & 16.7 & 18.5 & 18.5 & \cellcolor{highlightColor} \textbf{34.5} & \cellcolor{highlightColor} \textbf{42.3} & 45.1 & 26.8 & 33.0 & 40.1 & 19.6 & 26.8 & 27.7  \\ 
    &    & &         \quad+\textbf{\methodname(Ours)} & \cellcolor{highlightColor} \textbf{22.4} (+34.1\%) & \cellcolor{highlightColor} \textbf{25.0} (+35.1\%) & \cellcolor{highlightColor} \textbf{25.1} (+35.7\%) & 31.9 (-7.5\%) & 42.0 (-0.7\%) & \cellcolor{highlightColor} \textbf{47.1} (+4.4\%) & \cellcolor{highlightColor} \textbf{34.9} (+30.2\%) & \cellcolor{highlightColor} \textbf{42.3} (+28.2\%) & \cellcolor{highlightColor} \textbf{48.2} (+20.2\%) & \cellcolor{highlightColor} \textbf{25.9} (+32.1\%) & \cellcolor{highlightColor} \textbf{36.5} (+36.2\%) & \cellcolor{highlightColor} \textbf{38.4} (+38.6\%)  \\ 
 \cmidrule(lr){4-16}  
     &    &\multirow{2}{*}{Spatter} &         DSGDetr~\cite{Feng_2021} & 18.6 & 20.3 & 20.3 & \cellcolor{highlightColor} \textbf{41.4} & \cellcolor{highlightColor} \textbf{51.0} & 54.3 & 29.0 & 36.5 & 48.0 & 21.7 & 29.3 & 30.1  \\ 
    &    & &         \quad+\textbf{\methodname(Ours)} & \cellcolor{highlightColor} \textbf{24.7} (+32.8\%) & \cellcolor{highlightColor} \textbf{27.4} (+35.0\%) & \cellcolor{highlightColor} \textbf{27.5} (+35.5\%) & 38.6 (-6.8\%) & 50.6 (-0.8\%) & \cellcolor{highlightColor} \textbf{56.1} (+3.3\%) & \cellcolor{highlightColor} \textbf{37.4} (+29.0\%) & \cellcolor{highlightColor} \textbf{46.1} (+26.3\%) & \cellcolor{highlightColor} \textbf{55.8} (+16.2\%) & \cellcolor{highlightColor} \textbf{27.9} (+28.6\%) & \cellcolor{highlightColor} \textbf{39.3} (+34.1\%) & \cellcolor{highlightColor} \textbf{41.4} (+37.5\%)  \\ 
 \cmidrule(lr){4-16}  
     &    &\multirow{2}{*}{Contrast} &         DSGDetr~\cite{Feng_2021} & 20.0 & 21.8 & 21.8 & \cellcolor{highlightColor} \textbf{42.4} & \cellcolor{highlightColor} \textbf{52.2} & \cellcolor{highlightColor} \textbf{55.5} & 31.3 & 38.6 & 48.8 & 23.6 & 31.9 & 32.9  \\ 
    &    & &         \quad+\textbf{\methodname(Ours)} & \cellcolor{highlightColor} \textbf{24.9} (+24.5\%) & \cellcolor{highlightColor} \textbf{27.7} (+27.1\%) & \cellcolor{highlightColor} \textbf{27.8} (+27.5\%) & 36.9 (-13.0\%) & 49.0 (-6.1\%) & 54.6 (-1.6\%) & \cellcolor{highlightColor} \textbf{37.9} (+21.1\%) & \cellcolor{highlightColor} \textbf{46.0} (+19.2\%) & \cellcolor{highlightColor} \textbf{54.3} (+11.3\%) & \cellcolor{highlightColor} \textbf{27.5} (+16.5\%) & \cellcolor{highlightColor} \textbf{38.6} (+21.0\%) & \cellcolor{highlightColor} \textbf{40.6} (+23.4\%)  \\ 
 \cmidrule(lr){4-16}  
     &    &\multirow{2}{*}{Brightness} &         DSGDetr~\cite{Feng_2021} & 23.6 & 25.7 & 25.7 & \cellcolor{highlightColor} \textbf{50.8} & \cellcolor{highlightColor} \textbf{61.9} & 65.5 & 36.8 & 45.4 & 57.5 & 27.6 & 37.5 & 38.6  \\ 
    &    & &         \quad+\textbf{\methodname(Ours)} & \cellcolor{highlightColor} \textbf{29.8} (+26.3\%) & \cellcolor{highlightColor} \textbf{33.2} (+29.2\%) & \cellcolor{highlightColor} \textbf{33.2} (+29.2\%) & 45.5 (-10.4\%) & 59.9 (-3.2\%) & \cellcolor{highlightColor} \textbf{66.1} (+0.9\%) & \cellcolor{highlightColor} \textbf{45.0} (+22.3\%) & \cellcolor{highlightColor} \textbf{55.4} (+22.0\%) & \cellcolor{highlightColor} \textbf{65.3} (+13.6\%) & \cellcolor{highlightColor} \textbf{33.7} (+22.1\%) & \cellcolor{highlightColor} \textbf{48.1} (+28.3\%) & \cellcolor{highlightColor} \textbf{50.6} (+31.1\%)  \\ 
 \cmidrule(lr){4-16}  
     &    &\multirow{2}{*}{Pixelate} &         DSGDetr~\cite{Feng_2021} & 21.6 & 23.3 & 23.3 & \cellcolor{highlightColor} \textbf{48.1} & \cellcolor{highlightColor} \textbf{59.7} & 63.8 & 33.5 & 42.6 & 56.9 & 25.4 & 33.7 & 34.6  \\ 
    &    & &         \quad+\textbf{\methodname(Ours)} & \cellcolor{highlightColor} \textbf{27.7} (+28.2\%) & \cellcolor{highlightColor} \textbf{30.5} (+30.9\%) & \cellcolor{highlightColor} \textbf{30.5} (+30.9\%) & 43.8 (-8.9\%) & 57.7 (-3.4\%) & \cellcolor{highlightColor} \textbf{64.1} (+0.5\%) & \cellcolor{highlightColor} \textbf{42.5} (+26.9\%) & \cellcolor{highlightColor} \textbf{52.8} (+23.9\%) & \cellcolor{highlightColor} \textbf{63.1} (+10.9\%) & \cellcolor{highlightColor} \textbf{31.4} (+23.6\%) & \cellcolor{highlightColor} \textbf{44.5} (+32.0\%) & \cellcolor{highlightColor} \textbf{46.7} (+35.0\%)  \\ 
 \cmidrule(lr){4-16}  
     &    &\multirow{2}{*}{Compression} &         DSGDetr~\cite{Feng_2021} & 19.9 & 21.5 & 21.6 & \cellcolor{highlightColor} \textbf{45.0} & \cellcolor{highlightColor} \textbf{55.7} & 59.5 & 31.3 & 40.1 & 52.4 & 23.4 & 31.2 & 32.1  \\ 
    &    & &         \quad+\textbf{\methodname(Ours)} & \cellcolor{highlightColor} \textbf{26.6} (+33.7\%) & \cellcolor{highlightColor} \textbf{29.3} (+36.3\%) & \cellcolor{highlightColor} \textbf{29.4} (+36.1\%) & 41.9 (-6.9\%) & 55.1 (-1.1\%) & \cellcolor{highlightColor} \textbf{61.5} (+3.4\%) & \cellcolor{highlightColor} \textbf{40.8} (+30.4\%) & \cellcolor{highlightColor} \textbf{50.5} (+25.9\%) & \cellcolor{highlightColor} \textbf{60.2} (+14.9\%) & \cellcolor{highlightColor} \textbf{29.8} (+27.4\%) & \cellcolor{highlightColor} \textbf{42.1} (+34.9\%) & \cellcolor{highlightColor} \textbf{44.0} (+37.1\%)  \\ 
 \cmidrule(lr){4-16}  
     &    &\multirow{2}{*}{Sun Glare} &         DSGDetr~\cite{Feng_2021} & 12.1 & 13.2 & 13.2 & \cellcolor{highlightColor} \textbf{26.3} & 32.5 & 34.7 & 19.3 & 24.4 & 30.2 & 14.2 & 19.2 & 19.6  \\ 
    &    & &         \quad+\textbf{\methodname(Ours)} & \cellcolor{highlightColor} \textbf{17.3} (+43.0\%) & \cellcolor{highlightColor} \textbf{19.4} (+47.0\%) & \cellcolor{highlightColor} \textbf{19.4} (+47.0\%) & 25.8 (-1.9\%) & \cellcolor{highlightColor} \textbf{34.3} (+5.5\%) & \cellcolor{highlightColor} \textbf{38.5} (+11.0\%) & \cellcolor{highlightColor} \textbf{26.6} (+37.8\%) & \cellcolor{highlightColor} \textbf{32.2} (+32.0\%) & \cellcolor{highlightColor} \textbf{37.3} (+23.5\%) & \cellcolor{highlightColor} \textbf{19.4} (+36.6\%) & \cellcolor{highlightColor} \textbf{27.6} (+43.7\%) & \cellcolor{highlightColor} \textbf{29.0} (+48.0\%)  \\ 
 \cmidrule(lr){4-16}  
     &    &\multirow{2}{*}{Dust} &         DSGDetr~\cite{Feng_2021} & 13.2 & 14.5 & 14.6 & \cellcolor{highlightColor} \textbf{28.5} & \cellcolor{highlightColor} \textbf{35.4} & \cellcolor{highlightColor} \textbf{37.7} & 21.5 & 26.8 & 34.2 & 15.4 & 20.8 & 21.4  \\ 
    &    & &         \quad+\textbf{\methodname(Ours)} & \cellcolor{highlightColor} \textbf{16.6} (+25.8\%) & \cellcolor{highlightColor} \textbf{18.6} (+28.3\%) & \cellcolor{highlightColor} \textbf{18.6} (+27.4\%) & 25.1 (-11.9\%) & 32.9 (-7.1\%) & 36.8 (-2.4\%) & \cellcolor{highlightColor} \textbf{25.9} (+20.5\%) & \cellcolor{highlightColor} \textbf{31.0} (+15.7\%) & \cellcolor{highlightColor} \textbf{37.0} (+8.2\%) & \cellcolor{highlightColor} \textbf{19.1} (+24.0\%) & \cellcolor{highlightColor} \textbf{26.3} (+26.4\%) & \cellcolor{highlightColor} \textbf{27.4} (+28.0\%)  \\ 
 \cmidrule(lr){4-16}  
     &    &\multirow{2}{*}{Saturate} &         DSGDetr~\cite{Feng_2021} & 25.9 & 28.4 & 28.4 & \cellcolor{highlightColor} \textbf{54.6} & \cellcolor{highlightColor} \textbf{66.2} & 69.9 & 40.7 & 49.3 & 62.4 & 30.6 & 41.7 & 42.9  \\ 
    &    & &         \quad+\textbf{\methodname(Ours)} & \cellcolor{highlightColor} \textbf{31.5} (+21.6\%) & \cellcolor{highlightColor} \textbf{35.1} (+23.6\%) & \cellcolor{highlightColor} \textbf{35.2} (+23.9\%) & 48.8 (-10.6\%) & 63.8 (-3.6\%) & \cellcolor{highlightColor} \textbf{70.4} (+0.7\%) & \cellcolor{highlightColor} \textbf{47.4} (+16.5\%) & \cellcolor{highlightColor} \textbf{58.4} (+18.5\%) & \cellcolor{highlightColor} \textbf{68.8} (+10.3\%) & \cellcolor{highlightColor} \textbf{35.7} (+16.7\%) & \cellcolor{highlightColor} \textbf{51.0} (+22.3\%) & \cellcolor{highlightColor} \textbf{53.5} (+24.7\%)  \\ 
          \hline 
    \end{tabular}
    }
\end{table*}

\begin{table*}[!h]
    \centering
    \captionsetup{font=small}
    \caption{Robustness Evaluation Results for SGG.}
    \label{tab:sup_sgg_corruptions_5}
    \renewcommand{\arraystretch}{1.2} 
    \resizebox{\textwidth}{!}{
    \begin{tabular}{l|l|l|l|ccc|cccccc|ccc}
    \hline
      \multirow{2}{*}{Severity} & \multirow{2}{*}{Mode} & \multirow{2}{*}{Corruption} & \multirow{2}{*}{Method} & \multicolumn{3}{c}{\textbf{With Constraint}} & \multicolumn{6}{c}{\textbf{No Constraint}} & \multicolumn{3}{c}{\textbf{Semi Constraint}} \\ 
        \cmidrule(lr){5-7} \cmidrule(lr){8-13} \cmidrule(lr){14-16} 
  & & & & \textbf{mR@10} & \textbf{mR@20} & \textbf{mR@50} & \textbf{R@10} & \textbf{R@20} & \textbf{R@50}  & \textbf{mR@10} & \textbf{mR@20} & \textbf{mR@50}  & \textbf{mR@10} & \textbf{mR@20} & \textbf{mR@50}  \\ \hline
   \multirow{32}{*}{5} &      \multirow{32}{*}{predcls} & \multirow{2}{*}{Gaussian Noise} &         STTran~\cite{cong_et_al_sttran_2021} & 20.0 & 22.3 & 22.4 & \cellcolor{highlightColor} \textbf{64.2} & \cellcolor{highlightColor} \textbf{87.6} & 99.0 & 31.4 & 52.5 & 79.7 & 26.0 & 36.6 & 38.5  \\ 
    &    & &         \quad+\textbf{\methodname(Ours)} & \cellcolor{highlightColor} \textbf{37.6} (+88.0\%) & \cellcolor{highlightColor} \textbf{43.8} (+96.4\%) & \cellcolor{highlightColor} \textbf{43.9} (+96.0\%) & 62.5 (-2.6\%) & 84.6 (-3.4\%) & \cellcolor{highlightColor} \textbf{99.0} (0.0\%) & \cellcolor{highlightColor} \textbf{57.5} (+83.1\%) & \cellcolor{highlightColor} \textbf{77.7} (+48.0\%) & \cellcolor{highlightColor} \textbf{92.7} (+16.3\%) & \cellcolor{highlightColor} \textbf{42.2} (+62.3\%) & \cellcolor{highlightColor} \textbf{60.0} (+63.9\%) & \cellcolor{highlightColor} \textbf{62.9} (+63.4\%)  \\ 
 \cmidrule(lr){4-16}  
     &    &\multirow{2}{*}{Shot Noise} &         STTran~\cite{cong_et_al_sttran_2021} & 20.3 & 22.8 & 22.9 & \cellcolor{highlightColor} \textbf{64.6} & \cellcolor{highlightColor} \textbf{88.0} & 99.0 & 32.1 & 53.4 & 79.9 & 26.4 & 37.1 & 39.0  \\ 
    &    & &         \quad+\textbf{\methodname(Ours)} & \cellcolor{highlightColor} \textbf{37.5} (+84.7\%) & \cellcolor{highlightColor} \textbf{43.7} (+91.7\%) & \cellcolor{highlightColor} \textbf{43.8} (+91.3\%) & 62.4 (-3.4\%) & 84.6 (-3.9\%) & \cellcolor{highlightColor} \textbf{99.0} (0.0\%) & \cellcolor{highlightColor} \textbf{57.1} (+77.9\%) & \cellcolor{highlightColor} \textbf{77.9} (+45.9\%) & \cellcolor{highlightColor} \textbf{92.8} (+16.1\%) & \cellcolor{highlightColor} \textbf{42.2} (+59.8\%) & \cellcolor{highlightColor} \textbf{60.5} (+63.1\%) & \cellcolor{highlightColor} \textbf{63.6} (+63.1\%)  \\ 
 \cmidrule(lr){4-16}  
     &    &\multirow{2}{*}{Impulse Noise} &         STTran~\cite{cong_et_al_sttran_2021} & 20.3 & 22.8 & 22.9 & \cellcolor{highlightColor} \textbf{64.6} & \cellcolor{highlightColor} \textbf{88.0} & 99.0 & 31.9 & 52.9 & 79.8 & 26.3 & 37.2 & 39.1  \\ 
    &    & &         \quad+\textbf{\methodname(Ours)} & \cellcolor{highlightColor} \textbf{37.7} (+85.7\%) & \cellcolor{highlightColor} \textbf{43.8} (+92.1\%) & \cellcolor{highlightColor} \textbf{43.9} (+91.7\%) & 62.4 (-3.4\%) & 84.7 (-3.8\%) & \cellcolor{highlightColor} \textbf{99.0} (0.0\%) & \cellcolor{highlightColor} \textbf{57.4} (+79.9\%) & \cellcolor{highlightColor} \textbf{77.7} (+46.9\%) & \cellcolor{highlightColor} \textbf{92.6} (+16.0\%) & \cellcolor{highlightColor} \textbf{42.0} (+59.7\%) & \cellcolor{highlightColor} \textbf{60.1} (+61.6\%) & \cellcolor{highlightColor} \textbf{63.1} (+61.4\%)  \\ 
 \cmidrule(lr){4-16}  
     &    &\multirow{2}{*}{Speckle Noise} &         STTran~\cite{cong_et_al_sttran_2021} & 23.1 & 26.1 & 26.2 & \cellcolor{highlightColor} \textbf{66.8} & \cellcolor{highlightColor} \textbf{89.6} & \cellcolor{highlightColor} \textbf{99.1} & 37.1 & 57.8 & 80.3 & 30.2 & 43.1 & 45.3  \\ 
    &    & &         \quad+\textbf{\methodname(Ours)} & \cellcolor{highlightColor} \textbf{38.8} (+68.0\%) & \cellcolor{highlightColor} \textbf{45.5} (+74.3\%) & \cellcolor{highlightColor} \textbf{45.5} (+73.7\%) & 60.2 (-9.9\%) & 82.9 (-7.5\%) & 98.4 (-0.7\%) & \cellcolor{highlightColor} \textbf{59.4} (+60.1\%) & \cellcolor{highlightColor} \textbf{77.8} (+34.6\%) & \cellcolor{highlightColor} \textbf{92.5} (+15.2\%) & \cellcolor{highlightColor} \textbf{43.6} (+44.4\%) & \cellcolor{highlightColor} \textbf{60.7} (+40.8\%) & \cellcolor{highlightColor} \textbf{63.1} (+39.3\%)  \\ 
 \cmidrule(lr){4-16}  
     &    &\multirow{2}{*}{Gaussian Blur} &         STTran~\cite{cong_et_al_sttran_2021} & 25.3 & 28.5 & 28.6 & \cellcolor{highlightColor} \textbf{67.9} & \cellcolor{highlightColor} \textbf{89.2} & 99.0 & 38.2 & 57.6 & 79.9 & 30.6 & 43.1 & 44.9  \\ 
    &    & &         \quad+\textbf{\methodname(Ours)} & \cellcolor{highlightColor} \textbf{38.7} (+53.0\%) & \cellcolor{highlightColor} \textbf{45.7} (+60.4\%) & \cellcolor{highlightColor} \textbf{45.9} (+60.5\%) & 65.0 (-4.3\%) & 85.7 (-3.9\%) & \cellcolor{highlightColor} \textbf{99.0} (0.0\%) & \cellcolor{highlightColor} \textbf{59.5} (+55.8\%) & \cellcolor{highlightColor} \textbf{78.1} (+35.6\%) & \cellcolor{highlightColor} \textbf{92.5} (+15.8\%) & \cellcolor{highlightColor} \textbf{43.0} (+40.5\%) & \cellcolor{highlightColor} \textbf{62.0} (+43.9\%) & \cellcolor{highlightColor} \textbf{64.5} (+43.7\%)  \\ 
 \cmidrule(lr){4-16}  
     &    &\multirow{2}{*}{Defocus Blur} &         STTran~\cite{cong_et_al_sttran_2021} & 25.8 & 29.2 & 29.3 & \cellcolor{highlightColor} \textbf{68.4} & \cellcolor{highlightColor} \textbf{89.6} & \cellcolor{highlightColor} \textbf{99.1} & 39.0 & 58.1 & 80.2 & 31.4 & 44.1 & 46.0  \\ 
    &    & &         \quad+\textbf{\methodname(Ours)} & \cellcolor{highlightColor} \textbf{38.9} (+50.8\%) & \cellcolor{highlightColor} \textbf{46.0} (+57.5\%) & \cellcolor{highlightColor} \textbf{46.2} (+57.7\%) & 65.2 (-4.7\%) & 86.0 (-4.0\%) & 99.0 (-0.1\%) & \cellcolor{highlightColor} \textbf{60.0} (+53.8\%) & \cellcolor{highlightColor} \textbf{78.5} (+35.1\%) & \cellcolor{highlightColor} \textbf{92.8} (+15.7\%) & \cellcolor{highlightColor} \textbf{43.6} (+38.9\%) & \cellcolor{highlightColor} \textbf{63.2} (+43.3\%) & \cellcolor{highlightColor} \textbf{65.9} (+43.3\%)  \\ 
 \cmidrule(lr){4-16}  
     &    &\multirow{2}{*}{Fog} &         STTran~\cite{cong_et_al_sttran_2021} & 26.5 & 30.2 & 30.3 & \cellcolor{highlightColor} \textbf{70.2} & \cellcolor{highlightColor} \textbf{91.1} & \cellcolor{highlightColor} \textbf{99.1} & 41.6 & 61.0 & 80.5 & 33.2 & 46.8 & 48.7  \\ 
    &    & &         \quad+\textbf{\methodname(Ours)} & \cellcolor{highlightColor} \textbf{42.6} (+60.8\%) & \cellcolor{highlightColor} \textbf{50.9} (+68.5\%) & \cellcolor{highlightColor} \textbf{51.1} (+68.6\%) & 64.8 (-7.7\%) & 86.3 (-5.3\%) & 98.8 (-0.3\%) & \cellcolor{highlightColor} \textbf{63.8} (+53.4\%) & \cellcolor{highlightColor} \textbf{80.2} (+31.5\%) & \cellcolor{highlightColor} \textbf{92.7} (+15.2\%) & \cellcolor{highlightColor} \textbf{46.2} (+39.2\%) & \cellcolor{highlightColor} \textbf{65.5} (+40.0\%) & \cellcolor{highlightColor} \textbf{68.2} (+40.0\%)  \\ 
 \cmidrule(lr){4-16}  
     &    &\multirow{2}{*}{Frost} &         STTran~\cite{cong_et_al_sttran_2021} & 25.6 & 29.2 & 29.2 & \cellcolor{highlightColor} \textbf{69.4} & \cellcolor{highlightColor} \textbf{90.7} & \cellcolor{highlightColor} \textbf{99.1} & 41.0 & 60.9 & 80.5 & 32.7 & 46.1 & 48.0  \\ 
    &    & &         \quad+\textbf{\methodname(Ours)} & \cellcolor{highlightColor} \textbf{41.0} (+60.2\%) & \cellcolor{highlightColor} \textbf{49.0} (+67.8\%) & \cellcolor{highlightColor} \textbf{49.2} (+68.5\%) & 62.2 (-10.4\%) & 84.3 (-7.1\%) & 98.5 (-0.6\%) & \cellcolor{highlightColor} \textbf{62.5} (+52.4\%) & \cellcolor{highlightColor} \textbf{78.6} (+29.1\%) & \cellcolor{highlightColor} \textbf{92.7} (+15.2\%) & \cellcolor{highlightColor} \textbf{45.1} (+37.9\%) & \cellcolor{highlightColor} \textbf{62.9} (+36.4\%) & \cellcolor{highlightColor} \textbf{65.1} (+35.6\%)  \\ 
 \cmidrule(lr){4-16}  
     &    &\multirow{2}{*}{Spatter} &         STTran~\cite{cong_et_al_sttran_2021} & 25.7 & 29.2 & 29.3 & \cellcolor{highlightColor} \textbf{69.5} & \cellcolor{highlightColor} \textbf{91.1} & \cellcolor{highlightColor} \textbf{99.2} & 40.0 & 60.0 & 80.3 & 32.0 & 45.0 & 47.0  \\ 
    &    & &         \quad+\textbf{\methodname(Ours)} & \cellcolor{highlightColor} \textbf{41.3} (+60.7\%) & \cellcolor{highlightColor} \textbf{48.8} (+67.1\%) & \cellcolor{highlightColor} \textbf{48.9} (+66.9\%) & 58.8 (-15.4\%) & 82.3 (-9.7\%) & 98.0 (-1.2\%) & \cellcolor{highlightColor} \textbf{62.0} (+55.0\%) & \cellcolor{highlightColor} \textbf{78.6} (+31.0\%) & \cellcolor{highlightColor} \textbf{92.5} (+15.2\%) & \cellcolor{highlightColor} \textbf{44.6} (+39.4\%) & \cellcolor{highlightColor} \textbf{62.0} (+37.8\%) & \cellcolor{highlightColor} \textbf{64.3} (+36.8\%)  \\ 
 \cmidrule(lr){4-16}  
     &    &\multirow{2}{*}{Contrast} &         STTran~\cite{cong_et_al_sttran_2021} & 20.5 & 22.9 & 23.0 & \cellcolor{highlightColor} \textbf{64.9} & \cellcolor{highlightColor} \textbf{87.9} & 99.0 & 32.4 & 53.0 & 79.6 & 26.5 & 36.9 & 38.5  \\ 
    &    & &         \quad+\textbf{\methodname(Ours)} & \cellcolor{highlightColor} \textbf{37.7} (+83.9\%) & \cellcolor{highlightColor} \textbf{44.2} (+93.0\%) & \cellcolor{highlightColor} \textbf{44.3} (+92.6\%) & 63.3 (-2.5\%) & 85.0 (-3.3\%) & \cellcolor{highlightColor} \textbf{99.0} (0.0\%) & \cellcolor{highlightColor} \textbf{56.4} (+74.1\%) & \cellcolor{highlightColor} \textbf{76.0} (+43.4\%) & \cellcolor{highlightColor} \textbf{92.0} (+15.6\%) & \cellcolor{highlightColor} \textbf{41.2} (+55.5\%) & \cellcolor{highlightColor} \textbf{57.6} (+56.1\%) & \cellcolor{highlightColor} \textbf{59.6} (+54.8\%)  \\ 
 \cmidrule(lr){4-16}  
     &    &\multirow{2}{*}{Brightness} &         STTran~\cite{cong_et_al_sttran_2021} & 28.2 & 32.0 & 32.1 & \cellcolor{highlightColor} \textbf{71.3} & \cellcolor{highlightColor} \textbf{91.6} & \cellcolor{highlightColor} \textbf{99.2} & 42.8 & 62.0 & 80.4 & 34.5 & 49.0 & 51.2  \\ 
    &    & &         \quad+\textbf{\methodname(Ours)} & \cellcolor{highlightColor} \textbf{42.3} (+50.0\%) & \cellcolor{highlightColor} \textbf{50.4} (+57.5\%) & \cellcolor{highlightColor} \textbf{50.5} (+57.3\%) & 65.9 (-7.6\%) & 87.2 (-4.8\%) & 98.9 (-0.3\%) & \cellcolor{highlightColor} \textbf{64.0} (+49.5\%) & \cellcolor{highlightColor} \textbf{80.8} (+30.3\%) & \cellcolor{highlightColor} \textbf{92.8} (+15.4\%) & \cellcolor{highlightColor} \textbf{46.9} (+35.9\%) & \cellcolor{highlightColor} \textbf{67.8} (+38.4\%) & \cellcolor{highlightColor} \textbf{71.0} (+38.7\%)  \\ 
 \cmidrule(lr){4-16}  
     &    &\multirow{2}{*}{Pixelate} &         STTran~\cite{cong_et_al_sttran_2021} & 24.9 & 27.9 & 27.9 & \cellcolor{highlightColor} \textbf{67.3} & \cellcolor{highlightColor} \textbf{89.4} & \cellcolor{highlightColor} \textbf{99.1} & 37.5 & 57.5 & 80.0 & 30.5 & 42.9 & 44.9  \\ 
    &    & &         \quad+\textbf{\methodname(Ours)} & \cellcolor{highlightColor} \textbf{38.4} (+54.2\%) & \cellcolor{highlightColor} \textbf{45.5} (+63.1\%) & \cellcolor{highlightColor} \textbf{45.7} (+63.8\%) & 63.3 (-5.9\%) & 84.9 (-5.0\%) & 98.9 (-0.2\%) & \cellcolor{highlightColor} \textbf{59.6} (+58.9\%) & \cellcolor{highlightColor} \textbf{78.3} (+36.2\%) & \cellcolor{highlightColor} \textbf{92.5} (+15.6\%) & \cellcolor{highlightColor} \textbf{43.6} (+43.0\%) & \cellcolor{highlightColor} \textbf{61.9} (+44.3\%) & \cellcolor{highlightColor} \textbf{64.3} (+43.2\%)  \\ 
 \cmidrule(lr){4-16}  
     &    &\multirow{2}{*}{Compression} &         STTran~\cite{cong_et_al_sttran_2021} & 23.0 & 25.8 & 25.8 & \cellcolor{highlightColor} \textbf{66.0} & \cellcolor{highlightColor} \textbf{88.0} & 99.0 & 35.0 & 54.6 & 79.8 & 27.8 & 38.9 & 40.6  \\ 
    &    & &         \quad+\textbf{\methodname(Ours)} & \cellcolor{highlightColor} \textbf{36.4} (+58.3\%) & \cellcolor{highlightColor} \textbf{42.2} (+63.6\%) & \cellcolor{highlightColor} \textbf{42.3} (+64.0\%) & 63.8 (-3.3\%) & 85.1 (-3.3\%) & \cellcolor{highlightColor} \textbf{99.0} (0.0\%) & \cellcolor{highlightColor} \textbf{54.9} (+56.9\%) & \cellcolor{highlightColor} \textbf{75.8} (+38.8\%) & \cellcolor{highlightColor} \textbf{92.4} (+15.8\%) & \cellcolor{highlightColor} \textbf{41.0} (+47.5\%) & \cellcolor{highlightColor} \textbf{58.6} (+50.6\%) & \cellcolor{highlightColor} \textbf{61.7} (+52.0\%)  \\ 
 \cmidrule(lr){4-16}  
     &    &\multirow{2}{*}{Sun Glare} &         STTran~\cite{cong_et_al_sttran_2021} & 22.5 & 25.1 & 25.2 & \cellcolor{highlightColor} \textbf{66.7} & \cellcolor{highlightColor} \textbf{89.9} & \cellcolor{highlightColor} \textbf{99.1} & 36.7 & 56.7 & 80.0 & 28.9 & 40.5 & 42.3  \\ 
    &    & &         \quad+\textbf{\methodname(Ours)} & \cellcolor{highlightColor} \textbf{40.2} (+78.7\%) & \cellcolor{highlightColor} \textbf{47.5} (+89.2\%) & \cellcolor{highlightColor} \textbf{47.7} (+89.3\%) & 57.9 (-13.2\%) & 81.7 (-9.1\%) & 98.0 (-1.1\%) & \cellcolor{highlightColor} \textbf{60.3} (+64.3\%) & \cellcolor{highlightColor} \textbf{77.5} (+36.7\%) & \cellcolor{highlightColor} \textbf{92.7} (+15.9\%) & \cellcolor{highlightColor} \textbf{43.3} (+49.8\%) & \cellcolor{highlightColor} \textbf{59.3} (+46.4\%) & \cellcolor{highlightColor} \textbf{61.0} (+44.2\%)  \\ 
          \hline 
    \end{tabular}
    }
\end{table*}

\clearpage

\subsection{Robust Scene Graph Anticipation}

\subsubsection{Findings}
Table~\ref{tab:sup_sga_corruptions_vertical}, present the Robustness Evaluation Results for SGCLS for methods STTran+, DSGDetr+, STTran++, DSGDetr++ for Scene Graph Anticipation under various corruption scenarios. The results measure mR@10, mR@20, and mR@50, focusing on the impact of \methodname across different noise types. For Gaussian Noise, STTran++, mR@10 improves from 5.9 to 9.4 (+59.3\%); for DSGDet++, mR@20 improves from 5.7 to 8.8 (+44.4\%). However, for STTran+ \methodname underperforms for all metrics. The same can be observed for Dust, Spatter, Frost and Impulse noises; for all other corruptions, \methodname outperforms existing methods with the highest increments seen for STTran++ with an average of 40\% higher metrics.  

\subsubsection{Results}

\begin{table}[!h]
    \centering
    \captionsetup{font=small}
    \caption{Robustness Evaluation Results for SGA.}
    \label{tab:sup_sga_corruptions_vertical}
    \renewcommand{\arraystretch}{1.2} 
    \resizebox{0.55\linewidth}{!}{
\begin{tabular}{l|l|l|l|cccccc|cccccc}
    \hline
      \multirow{2}{*}{$\mathcal{F}$} & \multirow{2}{*}{Mode} & \multirow{2}{*}{Corruption} & \multirow{2}{*}{Method} & \multicolumn{6}{c}{\textbf{With Constraint}} & \multicolumn{6}{c}{\textbf{No Constraint}}  \\ 
        \cmidrule(lr){5-10} \cmidrule(lr){11-16} 
  & & & & \textbf{R@10} & \textbf{R@20} & \textbf{R@50} & \textbf{mR@10} & \textbf{mR@20} & \textbf{mR@50} & \textbf{R@10} & \textbf{R@20} & \textbf{R@50} & \textbf{mR@10} & \textbf{mR@20} & \textbf{mR@50}  \\ \hline
   \multirow{128}{*}{0.5} &      \multirow{128}{*}{sgcls} & \multirow{8}{*}{Gaussian Noise} &         STTran+~\cite{peddi_et_al_scene_sayer_2024} & \cellcolor{highlightColor} \textbf{13.2} & \cellcolor{highlightColor} \textbf{13.5} & \cellcolor{highlightColor} \textbf{13.5} & \cellcolor{highlightColor} \textbf{7.9} & \cellcolor{highlightColor} \textbf{8.4} & \cellcolor{highlightColor} \textbf{8.4} & \cellcolor{highlightColor} \textbf{22.1} & \cellcolor{highlightColor} \textbf{27.2} & \cellcolor{highlightColor} \textbf{28.7} & \cellcolor{highlightColor} \textbf{13.1} & \cellcolor{highlightColor} \textbf{18.9} & \cellcolor{highlightColor} \textbf{25.7}  \\ 
    &    & &         \quad+\textbf{\methodname(Ours)} & 7.9 (-40.2\%) & 8.1 (-40.0\%) & 8.1 (-40.0\%) & 5.1 (-35.4\%) & 5.4 (-35.7\%) & 5.4 (-35.7\%) & 12.0 (-45.7\%) & 18.1 (-33.5\%) & 22.8 (-20.6\%) & 7.0 (-46.6\%) & 11.7 (-38.1\%) & 19.8 (-23.0\%)  \\ 
    &    & &         DSGDetr+~\cite{peddi_et_al_scene_sayer_2024} & 10.1 & 10.4 & 10.4 & 5.5 & 5.8 & 5.8 & \cellcolor{highlightColor} \textbf{15.2} & \cellcolor{highlightColor} \textbf{21.7} & \cellcolor{highlightColor} \textbf{24.4} & 8.2 & 13.6 & \cellcolor{highlightColor} \textbf{22.2}  \\
\hline 
    \end{tabular}
    }
\end{table}

\clearpage

\newpage














\end{document}